\documentclass[10pt,twocolumn,letterpaper]{article}

\usepackage{iccv}
\usepackage{times}
\usepackage{epsfig}
\usepackage{graphicx}
\usepackage{amsmath}
\usepackage{amssymb}
\usepackage{booktabs}
\usepackage[accsupp]{axessibility}
\usepackage{caption}

\usepackage[noend]{algpseudocode}
\usepackage{algorithmicx,algorithm}

\usepackage{threeparttable}
\usepackage{multirow,booktabs}

\usepackage{float}
\usepackage{placeins}
\usepackage{dblfloatfix}

\usepackage{cuted}

\usepackage[export]{adjustbox}

\usepackage[pagebackref=true,breaklinks=true,letterpaper=true,colorlinks,bookmarks=false]{hyperref}

\usepackage[capitalize]{cleveref}
\crefname{section}{Sec.}{Secs.}
\Crefname{section}{Section}{Sections}
\Crefname{table}{Table}{Tables}
\crefname{table}{Tab.}{Tabs.}

\iccvfinalcopy % *** Uncomment this line for the final submission

\def\iccvPaperID{12682} % *** Enter the ICCV Paper ID here

\ificcvfinal\fi

\begin{document}

%%%%%%%%% TITLE
\title{StyleDomain: Efficient and Lightweight Parameterizations of StyleGAN for One-shot and Few-shot Domain Adaptation}

\author{
Aibek Alanov\textsuperscript{\rm 2,1,}\thanks{Equal contribution}\:, Vadim Titov\textsuperscript{\rm 2,}\footnotemark[1]\:, Maksim Nakhodnov\textsuperscript{\rm 2,3}\footnotemark[1]\:, Dmitry Vetrov\textsuperscript{\rm 1,2}\\
	\small \textsuperscript{\rm 1}HSE University \quad \textsuperscript{\rm 2}AIRI \quad \textsuperscript{\rm 3}Lomonosov Moscow State University\\
	%Moscow, Russia\\
	{\small \tt aalanov@hse.ru, titov@2a2i.org,  nakhodnov@2a2i.org, dvetrov@hse.ru} 
}

\maketitle
\begin{strip}
\centering
  \vspace{-1.5cm}
  \includegraphics[width=\textwidth]{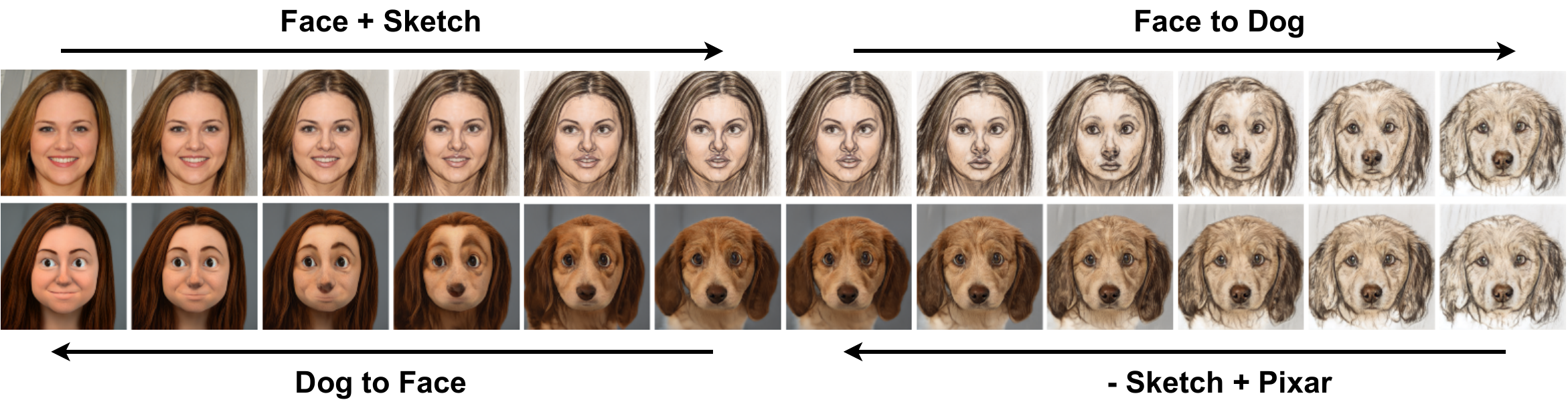}
  \captionof{figure}{Smooth cross-domain image morphing. Morphing in the space of generator weights could be successfully combined with morphing using StyleDomain directions, i.e. directions in the StyleSpace that can adapt the generator to new domains (see \Cref{sec:styledirections}). }
  \label{fig:teaser}
\end{strip}
\ificcvfinal\thispagestyle{empty}\fi

%%%%%%%%% ABSTRACT
\begin{abstract}
   Domain adaptation of GANs is a problem of fine-tuning GAN models pretrained on a large dataset (e.g. StyleGAN) to a specific domain with few samples (e.g. painting faces, sketches, etc.). While there are many methods that tackle this problem in different ways, there are still many important questions that remain unanswered. In this paper, we provide a systematic and in-depth analysis of the domain adaptation problem of GANs, focusing on the StyleGAN model. We perform a detailed exploration of the most important parts of StyleGAN that are responsible for adapting the generator to a new domain depending on the similarity between the source and target domains. As a result of this study, we propose new efficient and lightweight parameterizations of StyleGAN for domain adaptation. Particularly, we show that there exist directions in StyleSpace (StyleDomain directions) that are sufficient for adapting to similar domains. For dissimilar domains, we propose Affine+ and AffineLight+ parameterizations that allows us to outperform existing baselines in few-shot adaptation while having significantly less training parameters. Finally, we examine StyleDomain directions and discover their many surprising properties that we apply for domain mixing and cross-domain image morphing. Source code can be found at \href{https://github.com/AIRI-Institute/StyleDomain}{https://github.com/AIRI-Institute/StyleDomain}.
\end{abstract}

%%%%%%%%% BODY TEXT

\section{Introduction}
\label{sec:intro}
Recent years GANs \cite{goodfellow2014generative, karras2019style, karras2020analyzing, karras2021alias, brock2018large} have shown impressive results in image synthesis and offered many ways to control the generated data. In particular, the state-of-the-art StyleGAN models \cite{karras2019style, karras2020analyzing, karras2021alias}
have many practical applications such as image enhancement \cite{yang2021gan, luo2021time, chan2021glean, wang2021towards}, image editing \cite{jahanian2019steerability, shen2020interpreting, harkonen2020ganspace, tewari2020stylerig, abdal2019image2stylegan, wu2021stylespace, patashnik2021styleclip}, image-to-image translation \cite{pinkney2020resolution, huang2021unsupervised, song2021agilegan, gal2022stylegan} thanks to their high-quality image generation and their latent representation that has rich semantics and disentangled controls for localized meaningful image manipulations. However, it comes at a price, as the training of StyleGAN requires a large, high-quality dataset that significantly limits its applicability because many real-world domains are represented by few images. The standard approach to deal with this problem is transfer learning, i.e. fine-tuning the model pretrained on the source domain $A$ to the target domain $B$.

There are many domain adaptation methods for StyleGAN \cite{karras2020training, tran2021data, zhao2020differentiable, zhao2020image, liu2020towards, yang2021data, ojha2021few, pinkney2020resolution, alanov2022hyperdomainnet, gal2022stylegan, zhu2021mind, wu2021stylealign} that tackle this problem in different ways depending on the number of available images (e.g., one-shot/few-shot) from the target domain $B$ and the similarity between the source $A$ and target $B$ domains (e.g., faces → sketches, artistic portraits, or faces → dogs, cats). Most of these works implicitly assume that StyleGAN can be adapted to a new domain only if we fine-tune almost all its weights, even for similar domains. However, this common wisdom is poorly investigated and verified and there is a lack of analysis of which parts of StyleGAN are important depending on different data regimes and the similarity between domains. 

%They investigate their properties such as which parts of StyleGAN are changed mostly during domain adaptation and the transferability of latent rich semantics between aligned models. However, this empirical study can be further improved as many important questions remain untouched.  

%Some approaches consider the case when we adapt the StyleGAN to a far domain (e.g. faces $\rightarrow$ cars) and they typically fine-tune all its weights. Other methods deal with the target domains that are similar to the source one. For this case there are works that significantly reduce the parameter space for fine-tuning StyleGAN. However, all these papers focus on a specific aspect of the domain adaptation problem of StyleGAN and do not explore it in a more systematic way except the StyleAlign paper.

In this work, we aim to provide a systematic and comprehensive analysis of this question. Our investigation of the properties of the aligned StyleGAN models consists of several parts. First, in \Cref{sec:importance}, we identify what parts of the StyleGAN are sufficient for its adaptation depending on the similarity between the source $A$ and target $B$ domains. We discover that fine-tuning the whole synthesis convolutional network is not always necessary. In the case of similar $A$ and $B$ domains, the affine layers are sufficient for the adaptation. For more distant domains, we should optimize more parameters, however not the whole network. It suggests investigating new more efficient and lightweight parameterizations of StyleGAN to utilize them for domain adaptation.

%\vspace{-0.2cm}

In the second part of our analysis, we propose two new parameterizations of StyleGAN. For similar domains, we consider the latent space that is formed by the output of affine layers, i.e. StyleSpace \cite{wu2021stylespace}. We show that we can directly optimize directions in this space that can adapt to similar target domains with the same quality as fine-tuning all weights of StyleGAN (we call such directions as \emph{StyleDomain} directions). Further, we explore that we can zero out 80\% of StyleDomain direction coordinates without a quality degradation that gives even more lightweight parameterization (\textit{StyleSpaceSparse}). For more distant domains, we propose a new parameterization \textit{Affine$+$} that consists of affine layers and only one convolutional block from the synthesis network. It reduces the number of trainable parameters by 6 times and achieves the same quality. Then, we further improve Affine+ parameterization by utilizing low-rank decomposition for weights of affine layers and obtain \textit{AffineLight$+$} parameterization. It allows us to optimize by two orders less parameters compared to training the whole StyleGAN. These parameterizations show the state-of-the-art performance for few-shot adaptation for dissimilar domains outperforming more complicated and expensive baselines.

%1) we identify what parts of the StyleGAN model are responsible (sufficient) for adapting the generator to a new domain depending on the distance between source and target domains; 2) we explore the StyleSpace of StyleGAN and first show that there are directions in this space that can adapt the whole generator to new domains; 3) we examine these StyleSpace directions that correspond to different domains and discover their surprising properties: a) we can combine these directions to obtain new mixed domain, b) they can be successfully applied to other aligned StyleGAN model; our findings contributes to a better understanding of the observations from the StyleAlign paper (why aligned models are semantically aligned). 

Additionally, in \Cref{sec:styledirections}, we inspect StyleDomain directions and discover their surprising properties. The first one is mixability, i.e. we can sum up these directions to obtain a new mixed domain (e.g., see \Cref{fig:style_domain_mixing} as a mix of the Joker style, Pixar style and the style from the image). The second impressive property is transferability, i.e. the same StyleDomain directions can be applied to StyleGAN models that were fine-tuned to other domains (e.g., see \Cref{fig:style_domain_transfer} where we apply directions found for faces to dogs, cats and churches). We apply these findings to standard computer vision tasks such as image-to-image translation and cross-domain morphing. 
%In \Cref{sec:practical} we show that image-to-image translation can be solved by significantly less trainable parameters while achieving the same performance. In addition, we consider complex cross-domain image morphing between various domains that can be easily implemented based on our findings.

%To summarize, we conduct a comprehensive empirical study on the StyleGAN domain adaptation problem and reveal its many interesting aspects. We first show that StyleSpace has rich semantics not only for image editing but also for domain adaptation. We leverage our findings for a variety of computer vision applications. 

%Our analysis and insights deliver many practical improvements and applications:
%1) we show that the training space for domain adaptation can be significantly reduced; 2) we can easily mix different domains by combining corresponding StyleSpace directions; 3) we can transfer face-specific styles to non-face domains (cats/dogs, churches, etc.); 4) we can easily combine semantic editing and domain adaptation (improve?); 5) we can easily solve I2I and transfer learning tasks; 6) we can easily morph between different domains. 

\begin{figure*}[!h]
  \centering
  \includegraphics[width=\textwidth]{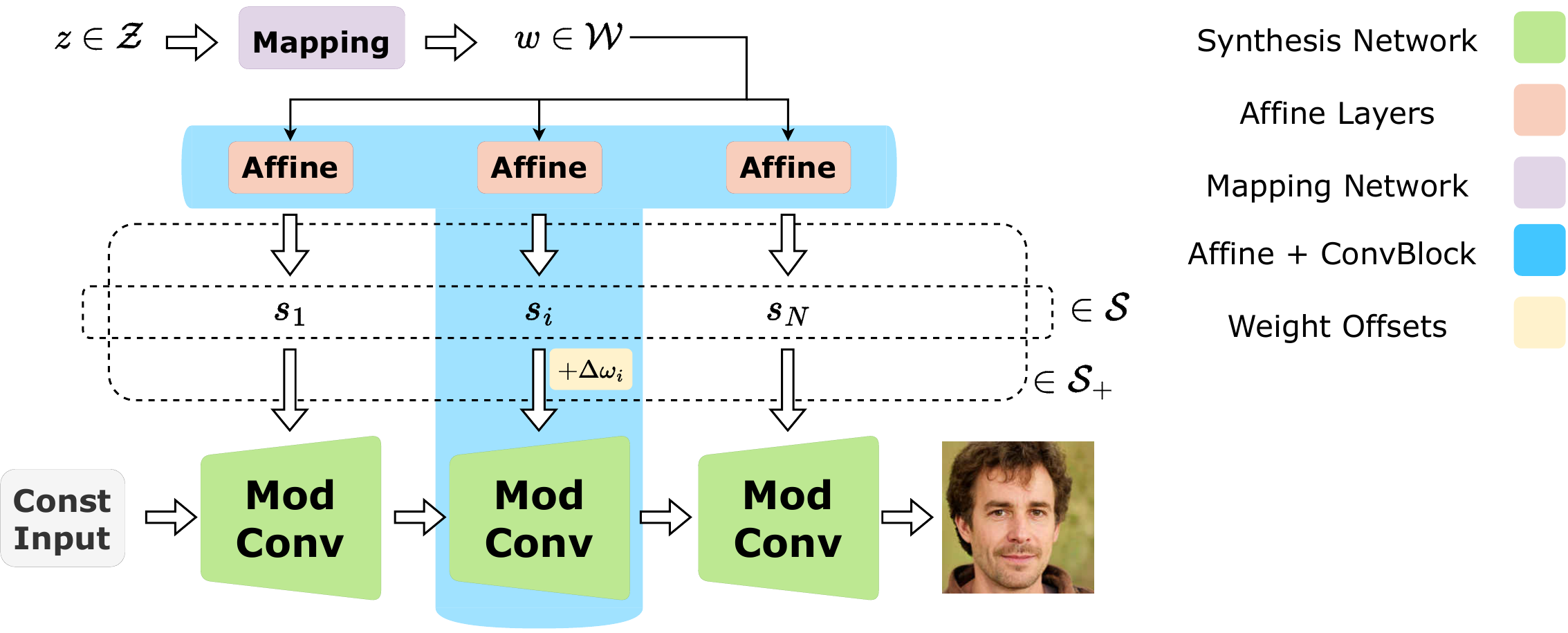}
  \caption{StyleGAN2 architecture. We introduce new latent space $S+$ for the for domain adaptation that combines StyleSpace and weight offsets for one block from the synthesis network.}
  \label{fig:diagram}
  \vspace{-0.5cm}
\end{figure*}

\section{Related Work}
\label{sec:related}

\textbf{Latent Spaces of StyleGAN.} 
After recent remarkable success of GANs \cite{goodfellow2014generative, karras2019style, karras2020analyzing, karras2021alias, brock2018large} in image synthesis, many works appeared that explore their latent representation for controllable image manipulation. In particular, the latent space of StyleGAN \cite{karras2019style, karras2020analyzing, karras2021alias} has attracted considerable attention. It consists of three levels: (i) the first latent space, $\mathcal{Z}$, is raw random noise (typically Gaussian); (ii) the intermediate latent spaces $\mathcal{W}, \mathcal{W}+$ \cite{abdal2019image2stylegan} are formed by the output of the mapping network; (iii) the last level is StyleSpace, $\mathcal{S}$, \cite{wu2021stylespace} that is spanned by the channel-wise style parameters after affine layers. It has been shown that these latent spaces have rich semantics \cite{karras2019style, karras2020analyzing}, and especially the StyleSpace that demonstrates the most disentangled and localized semantical directions \cite{wu2021stylespace}. In recent years many works have proposed to utilize such appealing properties of the StyleGAN latent spaces for image editing tasks \cite{jahanian2019steerability, shen2020interpreting, harkonen2020ganspace, tewari2020stylerig, abdal2019image2stylegan, wu2021stylespace, patashnik2021styleclip}. To apply these methods for real images it is necessary to inverse them into one of the latent space of StyleGAN which is another task that draws significant attention \cite{abdal2019image2stylegan, karras2020analyzing, guan2020collaborative, pidhorskyi2020adversarial, richardson2021encoding, tov2021designing, zhu2016generative, zhu2020domain}. We should note that all mentioned methods for image manipulation by controlling StyleGAN latent space allow only in-domain editing. In this paper, we show that in StyleSpace there exist such directions that can change the domain of images.

\textbf{Domain Adaptation of StyleGAN.} Recent years the problem of fine-tuning StyleGAN has generated a great deal of interest as it allows training the state-of-the-art generative model for a domain with few samples. There have appeared many works that tackle this problem in different ways depending on how similar the target domain is to the source one. Roughly, these methods can be divided into two groups. The first one deals with the case when the target and source domains are dissimilar (e.g. faces $\rightarrow$ cats, churches, etc.). In contrast, the second group considers the setting of similar domains (e.g. faces $\rightarrow$ stylized faces, painting faces, sketches, etc.). Methods from the first group typically require hundreds or thousands of samples from a new domain to adapt faithfully and they leverage data augmentations \cite{karras2020training, tran2021data, zhao2020differentiable, zhao2020image}, or freeze some layers of the discriminator to prevent overfitting \cite{mo2020freeze}, or train the discriminator with auxiliary losses to match the data more accurately \cite{liu2020towards, yang2021data}. 

In the setting of the second group it is sufficient to have only several samples (up to dozens) from the target domain for the successful adaptation and such approaches utilize another techniques. In particular, they introduce additional regularization terms \cite{tseng2021regularizing, li2020few}, preserve pairwise distances between instances in the source domain via cross-domain consistency loss \cite{ojha2021few}, mix weights of the fine-tuned and the base generators \cite{pinkney2020resolution}, use an auxiliary small network to enhance the training \cite{wang2020minegan}. More advanced methods \cite{gal2022stylegan, zhu2021mind, alanov2022hyperdomainnet} utilize the pretrained CLIP model \cite{radford2021learning} as the vision-language supervision for the text-based adaptation \cite{gal2022stylegan, alanov2022hyperdomainnet} or the one-shot image-based adaptation \cite{chefer2022image, zhu2021mind, chong2022jojogan, zhang2022generalized, zhang2022towards, alanov2022hyperdomainnet}. 

Recent work \cite{zhao2022few} analyzes the performance of methods from the second group for dissimilar domains in the low data regime. It was shown that those methods have a significant quality degradation in the few-shot regime. However, it can be partially mitigated by constraining model parameters.
 There are also approaches that introduce more lightweight parameterizations for domain adaptation \cite{noguchi2019image, robb2020few, chefer2022image, alanov2022hyperdomainnet}, however they work only for similar domains. In our paper, we propose efficient and highly lightweight parameterizations for both similar and dissimilar domains and show that they can achieve results on par with the state-of-the-art methods that optimize all weights of StyleGAN. 

There are few works that analyze the domain adaptation process of StyleGAN thoroughly. The paper \cite{wu2021stylealign} is the first attempt to perform such an in-depth study. In particular, they explore which parts of StyleGAN are mostly changed during fine-tuning and the transferability of the latent semantics after domain adaptation. However, this work does not analyze which parts of StyleGAN are sufficient for adapting depending on the similarity between source and target domains. In our paper, we provide a more systematic and comprehensive analysis that completes and improves the results from \cite{wu2021stylealign}.

%\section{Importance of Each Part of the StyleGAN for the Domain Adaptation}

\section{Importance of Each Part of the StyleGAN}
\label{sec:importance}
In this section, our goal is to analyze what parts of StyleGAN are important for domain adaptation. Similarly to previous works, we specifically focus on the state-of-the-art GAN architecture, StyleGAN2 \cite{karras2020analyzing}. For the source domain, we consider FFHQ \cite{karras2019style} as it is the large high-quality dataset that is suitable for training StyleGAN2 from scratch. For the target one, we test a wide range of different domains that we will describe further.

\textbf{StyleGAN2 structure and its main components.} We provide a diagram description of the StyleGAN2 architecture in \Cref{fig:diagram}. It consists of three parts:
\begin{itemize}
    \item \emph{mapping network} $f_M$ that transforms the input noise $z \in \mathcal{Z}$ (typically Gaussian) to the intermediate latent vector $w \in \mathcal{W}$;
    \item \emph{affine layers} $f_1^A, \dots, f_N^A$, each of them takes as input the vector $w$ and maps it to corresponding style vector $s_1 = f_1^A(w), \dots, s_N = f_N^A(w)$. The concatenation of these vectors form the vector from the StyleSpace: $s = (s_1, \dots, s_N) \in \mathcal{S}$;
    \item \emph{synthesis network} that is a composition of modulated convolutions. The weights of each convolution are modulated by the input style vector $s_i$ and applied to the input feature maps. The synthesis network also has tRGB convolutional layers that transform feature maps to RGB images and they also are modulated by corresponding style vectors. 
\end{itemize}
Accordingly, the StyleGAN2 $G_{\theta}$ generates from the input noise $z$ the output image $I = G_{\theta}(s(z))$ where $\theta$ are weights of the synthesis network, $s(z) = f^A(f_M(z))$, $f^A = \{f_1^A, \dots, f_N^A\}$ .

We will analyze these three components of StyleGAN2 and their impact on the domain adaptation process. It is common wisdom that the most important part for adaptation is the synthesis network, while the mapping network and affine layers are mostly responsible for the semantic manipulations within the source domain \cite{wu2021stylealign}. We aim to verify whether such a conception is correct.

In our experiments, we additionally consider the impact of the combination of affine layers and one convolutional block from the synthesis network on domain adaptation. It is a way to probe the intermediate case between affine layers and the synthesis network.

\textbf{Method to analyze the impact of each component.} The paper \cite{wu2021stylealign} has proposed to analyze the impact of each component by resetting its weights in the fine-tuned generator to their pretrained values and assessing the quality of the generated images. In our work, we propose another approach: to directly fine-tune only one of these components to explore which ones are sufficient for domain adaptation.

Let us describe our method in more detail. The optimization process for the problem of domain adaptation is the following
\begin{gather}
    \mathcal{L}_D\left(\left\{G_{\theta}(s(z_i)) \right\}_{i=1}^K\right) \rightarrow \min\limits,
\end{gather}
where $\mathcal{L}_D$ is domain adaptation loss that depends on the domain $D$ (we discuss it further) and the samples from the generator $\left\{G_{\theta}(s(z_i)) \right\}_{i=1}^K$, $z_1, \dots, z_K \in \mathcal{Z}$ are sampled noises.

Typically the generator $G_{\theta}$ is optimized with respect to all components, i.e.
\begin{gather}
    \mathcal{L}_D\left(\left\{G_{\theta}(s(z_i))\right\}_{i=1}^K\right) \rightarrow \min\limits_{\theta, f^A, f_M}.
\end{gather}

We propose to investigate settings when we optimize with respect to only one these components. We denote each parameter space as: $\{\theta\}$ -- \emph{SyntConv} parameterization, $\{f^A\}$ -- \emph{Affine} parameterization, $\{f_M\}$ -- \emph{Mapping} parameterization. The case we fine-tune all components of the StyleGAN2 we call \emph{Full} parameterization.

\begin{table*}[!t]
\centering
\caption{Quality and Diversity metrics \cite{alanov2022hyperdomainnet} for text-based and one-shot image-based domain adaptations with different parameterizations. Affine, StyleSpace and StyleSpaceSparse parameterizations achieve results comparable with the Full one. }
	\label{table:sim_dom_style}
  \begin{tabular}{ llllllllllll }
    \toprule
    & & \multicolumn{2}{c}{\textbf{Botero}} & \multicolumn{2}{c}{\textbf{Sketch}} & \multicolumn{2}{c}{\textbf{Disney (image)}} & \multicolumn{2}{c}{\textbf{Titan Erwin (image)}} \\
\cmidrule(lr){3-4} \cmidrule(lr){5-6} \cmidrule(lr){7-8} \cmidrule(lr){9-10}
Parameter Space & Size & Quality & Diversity & Quality & Diversity & Quality & Diversity & Quality & Diversity \\
\midrule\midrule
Full & 30.3M & $0.312$ & $0.228$ & $0.208$ & $0.296$ & $0.713$ & $0.247$ & $0.760$ & $0.194$ \\
SyntConv & 23.6M & $0.311$ & $0.224$ & $0.191$ & $0.292$ & $0.711$ & $0.259$ & $0.741$ & $0.217$ \\
Affine & 4.6M & $0.298$ & $0.221$ & $0.194$ & $0.296$ & $0.565$ & $0.359$ & $0.650$ & $0.314$ \\
Mapping & 2.1M & $0.226$ & $0.115$ & $0.182$ & $0.143$ & $0.717$ & $0.080$ & $0.645$ & $0.102$ \\
\midrule
\textbf{StyleSpace} & 6.0K & $0.309$ & $0.23$ & $0.193$ & $0.306$ & $0.627$ & $0.308$ & $0.672$ & $0.296$ \\
\textbf{StyleSpaceSparse} & \textbf{1.2K} & $0.322$ & $0.213$ & $0.201$ & $0.269$ & $0.617$ & $0.304$ & $0.659$ & $0.303$ \\
    \midrule
    \bottomrule
  \end{tabular}
\end{table*}

\begin{figure}[!b]
  \centering
  \includegraphics[width=0.5\textwidth]{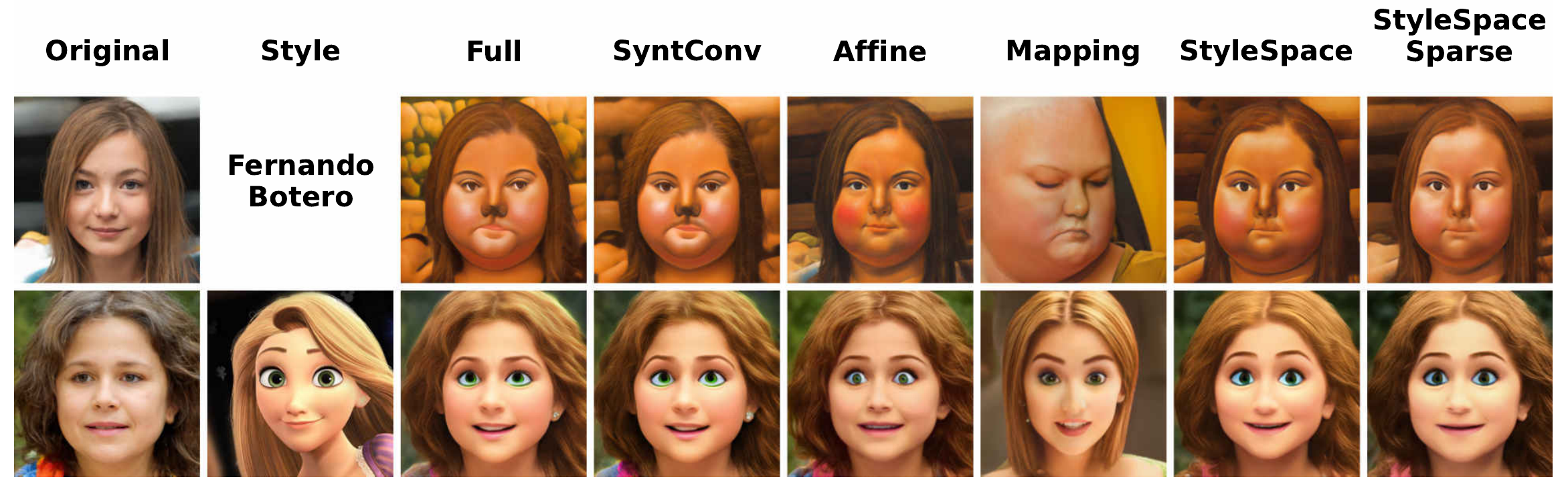}
  \caption{Text-based and image-based adaptation for different parameterizations. Affine, StyleSpace and StyleSpaceSparse parameterizations yield performance comparable with Full one. This style image is called "Disney".}
  \label{fig:sim_dom_parts}
\end{figure}

\textbf{Domain adaptation settings.} In our study, we consider two settings: one-shot and few-shot. For each data regime, we use different domains depending on their similarity to the source domain of realistic faces from FFHQ:
\begin{itemize}
    \item one-shot domains: for this setting, we consider only similar domains. It is the case when the target domain has the same geometry of faces and it preserves the identity of the person. It alters only the style of the image. In this regime, we consider not only one-shot image-based adaptation with the reference stylized face but additionally text-based adaptation with the text description of the target style (e.g., "Photo in the style of anime (pixar, sketch, etc.)”). See examples of such domains in \Cref{fig:sim_dom_parts}. 
    \item few-shot domains: for this regime, we examine more distant domains that have a face-like form but change the face geometry and identity in a stronger manner. As examples, we consider AFHQ dogs faces and cats faces \cite{choi2020stargan}. 
\end{itemize}

Depending on the data regime, we use different domain loss function $\mathcal{L}_D$. For one-shot domain adaptation, we apply the optimization loss from StyleGAN-NADA \cite{gal2022stylegan} in the case of text-based adaptation and another loss from DiFa \cite{zhang2022towards} for one-shot image-based adaptation. For few-shot domain adaptation, we utilize the fine-tuning procedure from StyleGAN-ADA \cite{karras2020training}. For more details about domain adaptation loss functions see \Cref{app:losses}.

To obtain quantitative comparisons in the case of a one-shot setting, we use Quality and Diversity metrics that were proposed in the HyperDomainNet paper \cite{alanov2022hyperdomainnet}. For few-shot adaptation, we compute FID metric \cite{heusel2017gans} using the standard protocol from \cite{karras2020training}.

\textbf{Analysis for one-shot domains.} For the analysis, we choose different text-based and one-shot image-based domains (see \Cref{app:sim_dom} for the full list and more details).
In experiments, we consider the four parameterizations (Full, SyntConv, Affine, Mapping) we discussed above.

We provide qualitative results in \Cref{fig:sim_dom_parts} with quantitative ones in \Cref{table:sim_dom_style}. More results see in \Cref{app:sim_dom}.

We observe that all three parameterizations, Full, SyntConv and Affine, show comparable performance in terms of both visual quality and objective metrics. The fact that the synthesis network is sufficient for similar domains was clear from the previous work \cite{wu2021stylealign}. However, our finding that the affine part is also sufficient is a new and surprising result. It means that we can change the domain of generated images without fine-tuning the synthesis network but just passing the modified style vector that comes from the affine part. Also, we observe that the mapping network shows poor visual quality and low diversity in the generated images when considering Diversity metric. It indicates that for successful adaptation it is important to update the style vector from $\mathcal{S}$ rather than the intermediate latent vector from $\mathcal{W}$ space.

\begin{table}[!b]
\centering
\caption{FID scores for domain adaptation with different parameterizations. We observe a significant gap between Affine and Full parametrizations that, however, can be drastically reduced by introducing Affine+ parameterization.}
	\label{table:one-shot}
  \begin{tabular}{llll}
\toprule
& &  \multicolumn{2}{c}{\textbf{Domains}} \\
\cmidrule(lr){3-4}
Parameter Space &   Size & Dog &      Cat \\
\midrule\midrule
Full               &  30.3M &   $20.3$ &    $7.1$ \\
SyntConv              &  23.6M &   $19.7$ &    $7.2$ \\
Affine            &   4.6M &   $70.1$ &   $27.6$ \\
Mapping           &   2.1M & $208.2$ &  $226.1$ \\
\midrule
\textbf{Affine$+$} &   5.1M &   \textbf{18.6} &    \textbf{7.0} \\
\textbf{AffineLight$+$} &   0.6M & $20.6$ &   $8.9$ \\
\textbf{StyleSpace}    &   \textbf{6.0K} & $75.8$ &  $22.0$ \\
\midrule\bottomrule
\end{tabular}
  \vspace{-0.5cm}
\end{table}

\textbf{Analysis for few-shot domains.}
For this setting, as we discussed above, we consider two datasets, AFHQ Dogs and Cats \cite{choi2020stargan}, and the results are provided in \Cref{fig:far_dom_parts} and in \Cref{table:one-shot}. See more results in \Cref{app:far_dom}.

We observe that results for Dogs and Cats are different compared to similar domains. In particular, we see that the Affine parameterization does not demonstrate the same quality as the Full parameterization. It can be seen from the degraded visual quality and visible gap in FID metric. However, it is still surprising that the generated images after adaptation have an adequate visual appearance considering that we do not fine-tune the synthesis network at all. 
SyntConv expectedly achieves results comparable with Full parameterization, and Mapping conversely shows poor quality on all datasets.

\section{Efficient and lightweight parameterizations of StyleGAN}
\label{sec:styledirections}

% \begin{figure}[!b]
%   \centering
%   \includegraphics[width=0.5\textwidth]{imgs/sim_dom_style.pdf}
%   \caption{StyleSpace parameterization performs domain adaptation comparable with Full parameterization even with $5000$ times less trained parameters.}
%   \label{fig:sim_dom_style}
% \end{figure}

\begin{figure}[!b]
  \centering
  \includegraphics[width=0.5\textwidth]{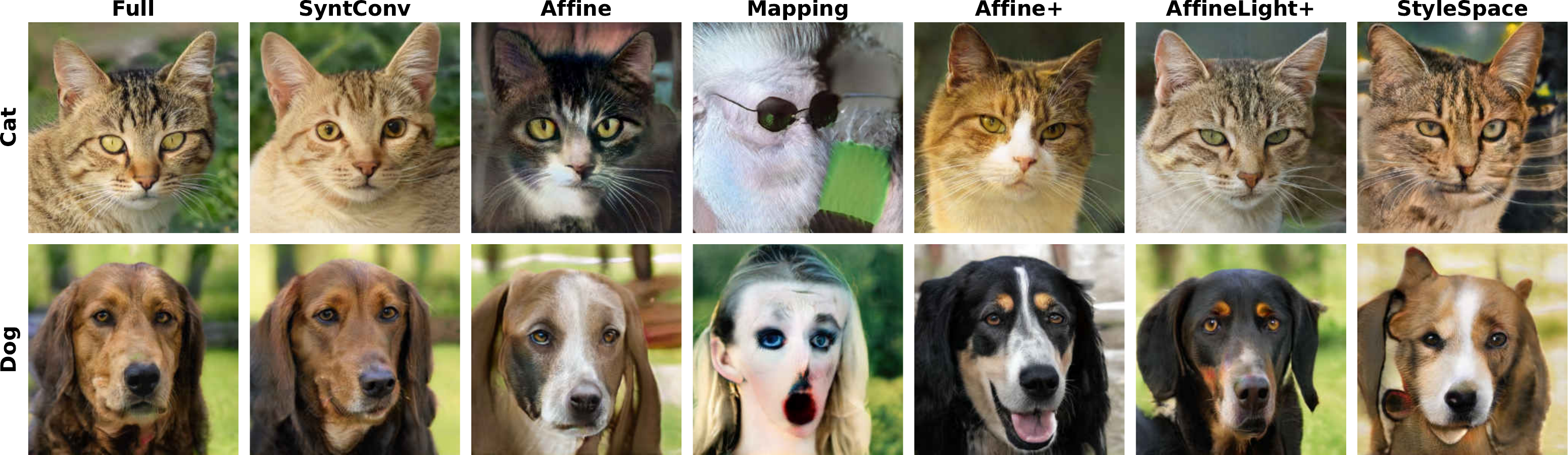}
  \caption{Domain adaptation for dissimilar domains. Affine+ parameterization produces results on par with the Full one.}
  \label{fig:far_dom_parts}
\end{figure}

\textbf{StyleSpace and StyleSpaceSparse.} Our findings from the previous section suggest that we can change the domain of generated images by modifying the style vector in the StyleSpace $\mathcal{S}$. To check this hypothesis, we will adapt StyleGAN2 by directly optimizing the direction in $\mathcal{S}$, i.e. during fine-tuning of StyleGAN we will optimize only $\Delta s^D$:
\begin{gather}
    \mathcal{L}_D\left(\left\{G_{\theta}(s(z_i) + \Delta s)\right\}_{i=1}^K\right) \rightarrow \min\limits_{\Delta s},
%\mathbb{E}\limits_Z~\mathcal{L}_D\left[G_{\theta}(s(z) + \Delta s)\right] \rightarrow \min\limits_{\Delta s},
\end{gather}
where $\Delta s = (\Delta s_1, \dots, \Delta s_N) \in \mathcal{S}$ is the optimized direction in the $\mathcal{S}$ for adapting the generator $G_{\theta}$ to the domain $D$. We call such directions $\Delta s$ as \emph{StyleDomain} directions.

Further, we explore that we can zero out most of the coordinates of StyleDomain directions without quality degradation. We use standard prunning technique when we leave 20\% of the largest absolute values in the StyleDomain vector and set the rest to zero. We call such parameterization as \textit{StyleSpaceSparse}. We examine different prunning rates and its performance in \Cref{app:style_results}. 

We apply these parameterizations to one-shot and few-shot domains and obtain the following results.

For one-shot adaptation, we provide results in \Cref{fig:sim_dom_parts} and in \Cref{table:sim_dom_style} (see more results in \Cref{app:style_results}). We observe that optimizing the StyleDomain direction $\Delta s^D$ achieves the same results both visually and quantitatively as the Full parameterization. It is new and important observation that StyleSpace allows not only image editing within domain but also generating samples from out-of domain of realistic human faces.

For few-shot domains, we observe that StyleSpace is not sufficient, which is expressed by the same significant quality degradation as for Affine parameterization. Further, we introduce a new parameterization that is efficient for more distant domains. 

% \begin{figure}[!b]
%   \centering
%   \includegraphics[width=0.5\textwidth]{imgs/far_dom_style.pdf}
%   \caption{Domain adaptation for moderately similar and dissimilar domains. StyleSpace+ parameterization closes the gap in quality for all presented domains despite having $60$ times less trainable parameters.}
%   \label{fig:far_dom_style}
% \end{figure}

\begin{figure}[!b]
  \centering
  \includegraphics[width=0.5\textwidth]{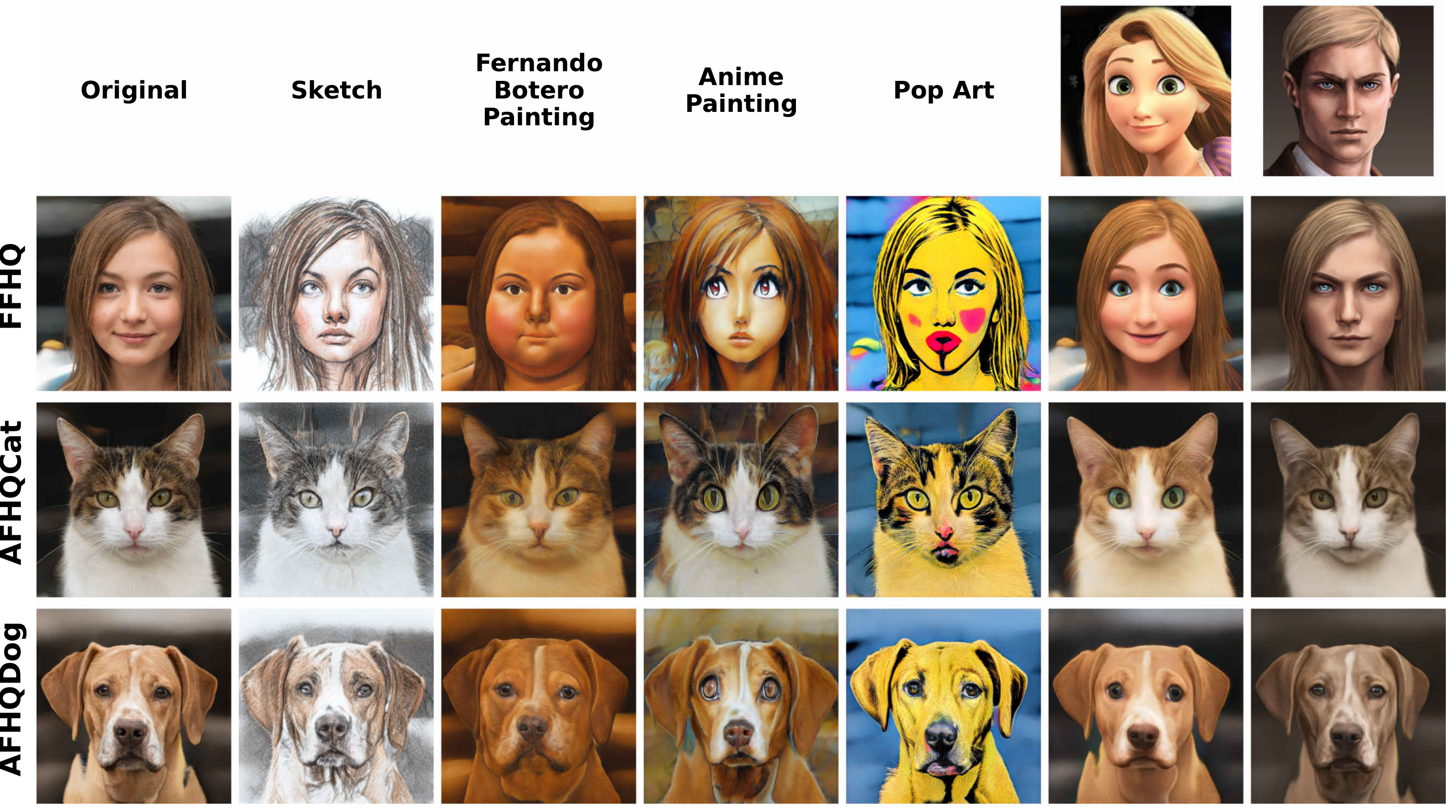}
  \caption{StyleSpace directions transfer from text-based and image-based domain adaptation to other fine-tuned models. We can successfully transfer style while preserving image content.}
  \label{fig:style_domain_transfer}
\end{figure}

\textbf{Affine$+$ and AffineLight$+$.} 
We aim to improve Affine parameterization for successfully adapting to dissimilar domains such as Dogs and Cats. We propose to extend it by adding one block from synthesis layer with specified spatial resolution. Such block has two convolutional layers with weights $\theta_1, \theta_2 \in \mathbb{R}^{512\times 512\times 3\times 3}$. Instead of fine-tuning all these weights we introduce more compact parameterization as offsets $\Delta \theta_1, \Delta \theta_2$ to these weights that are same across spatial dimensions, i.e. $\Delta \theta_1, \Delta \theta_2 \in \mathbb{R}^{512\times 512\times 1\times 1}$ (we observe that such reduction in size does affect the performance). In addition to $\Delta \theta_1, \Delta \theta_2$ we introduce the offsets to the weights $\theta^{tRGB} \in \mathbb{R}^{3\times 512\times 3\times 3}$ of the tRGB convolutional layer of the same block with similar parameterization $\Delta \theta^{tRGB} \in \mathbb{R}^{3\times 512\times 1\times 1}$. Further, we omit such detail about tRGB part in the sake of brevity. So, for this parameterization the optimization procedure has the following form:
\begin{gather}
    \mathcal{L}_D\left(\left\{G_{\theta, \Delta \theta_1, \Delta \theta_2}(s(z_i))\right\}_{i=1}^K\right) \rightarrow \min\limits_{\Delta \theta_1, \Delta \theta_2, f^A},
\end{gather}
where $G_{\Delta \theta_1, \Delta \theta_2}$ is the generator with weight offsets $\Delta \theta_1, \Delta \theta_2$ for the one block from the synthesis network.

\begin{figure}[!b]
  \centering
%   \begin{minipage}{0.5\textwidth}
%   \includegraphics[width=\textwidth]{imgs/ukiyo-e_botero_werewolf.pdf}
%   \end{minipage}\hfill
%   \begin{minipage}{0.5\textwidth}
  \includegraphics[width=0.5\textwidth]{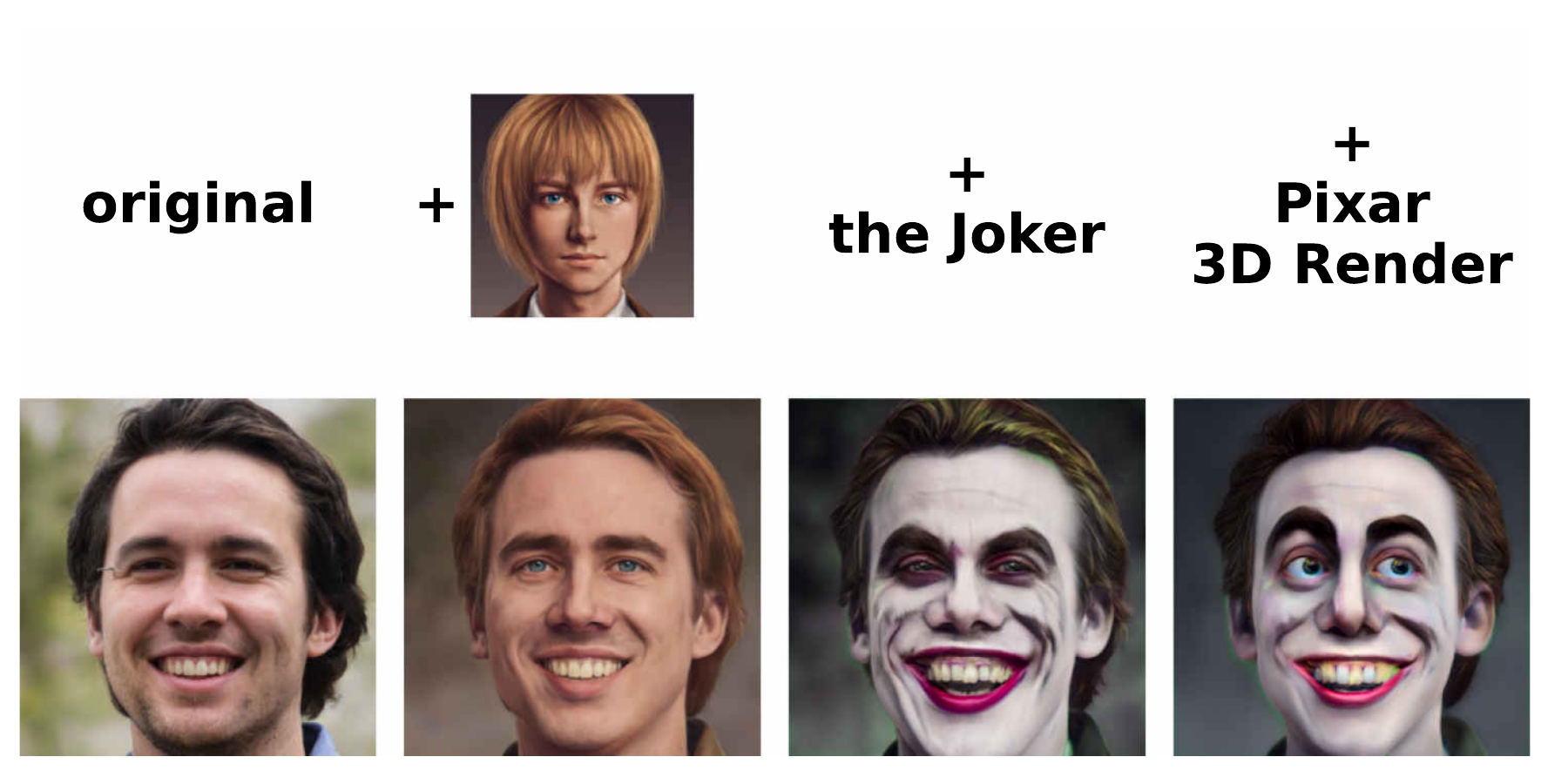}
%   \end{minipage}
  \caption{Example of mixing StyleDomain directions. We can combine different directions in order to perform adaptation into a semantically mixed domain.}
  \label{fig:style_domain_mixing}
\end{figure}

\begin{figure*}[!t]
  \centering
  \includegraphics[width=\textwidth]{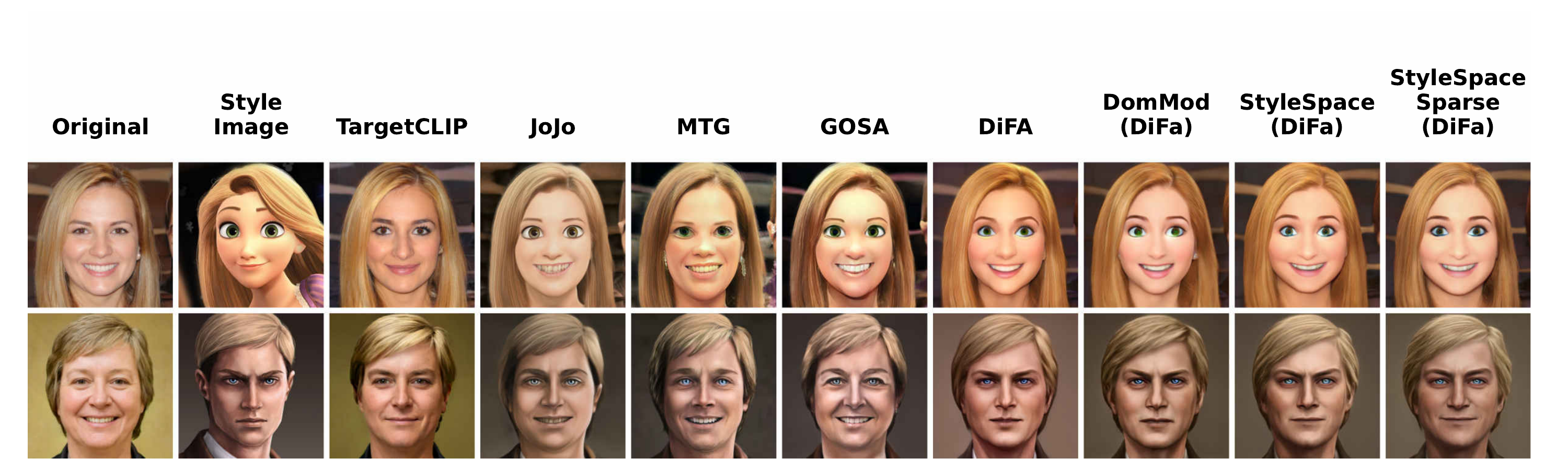}
  \caption{Comparison with baselines for one-shot image-based domain adaptation. StyleSpace and StyleSpaceSparse parameterizations achieve comparable quality as other methods while having much less trainable parameters.}
  \label{fig:one_shot_comparison}
\end{figure*}

We call such parameter space as \emph{Affine+}. We examine all blocks of the synthesis network to choose for this parameterization. We end up with the block with $64\times64$ resolution as it shows the best performance (see results of this analysis in \Cref{app:style_results}).

While Affine$+$ has already had 6 times less parameters than Full parameterization, we further reduce its size by applying low-rank decomposition to the weights of affine layers. We denote this parameterization as \textit{AffineLight$+$}. It gives us by two orders less parameters than Full parameterization with insignificant degradation in quality. We will show in \Cref{sec:experiments} that in low data regimes, this parameterization achieves comparable performance as Affine+ and outperforms other baselines. More details about applied low-rank decomposition can be found in \Cref{app:style_results}. 

We apply these two parameterizations to few-shot domains and obtain the following results.

We provide results  in \Cref{fig:far_dom_parts} and in \Cref{table:one-shot} (see more results in \Cref{app:style_results}). We see that Affine$+$ parameterization allows us
removing the performance gap from the Full parameterization both qualitatively and quantitatively. While the number
of parameters for the additional block in Affine$+$ accounts for only 2 \% of the synthesis network size. It shows that
the style vector allows adapting the generator even to more
distant domains if we just add a small part of the synthesis
network. 

We observe that AffineLight$+$ still shows adequate performance, except it has 100 times less parameters than the Full parameterization. We should notice that this parameterization is more suitable for low data regimes than those that we consider in \Cref{sec:experiments}. 

\begin{table*}[!b]
\centering
\caption{Quality and Diversity metrics \cite{alanov2022hyperdomainnet} for one-shot image-based domain adaptations with different methods. Memory denotes the memory needed for keeping adapted generators for all 12 domains for each method. StyleSpace and StyleSpaceSparse parameterizations achieve results comparable to other baselines while having much less trainable parameters. }
	\label{table:one_shot_comparison}
  \begin{tabular}{ llllllllllll }
    \toprule
    & & &  \multicolumn{2}{c}{\textbf{Titan Erwin}} & \multicolumn{2}{c}{\textbf{Disney}} & \multicolumn{2}{c}{\textbf{Across 12 domains}} \\
\cmidrule(lr){4-5} \cmidrule(lr){6-7} \cmidrule(lr){8-9}
Method & Size & Memory & Quality & Diversity & Quality & Diversity & Quality & Diversity \\
\midrule\midrule
JoJoGAN \cite{chong2022jojogan} & 30M & 1.80GB & $0.572$ & $0.292$ & $0.591$ & $0.260$ & $0.590 \pm 0.048$ & $0.257 \pm 0.025$ \\
MTG \cite{zhu2021mind} & 30M & 1.80GB & $0.607$ & $0.269$ & $0.509$ & $0.234$ & $0.586 \pm 0.054$ & $0.263 \pm 0.028$ \\
GOSA \cite{zhang2022generalized} & 30M & 1.80GB & $0.547$ & $0.283$ & $0.617$ & $0.216$ & $0.584 \pm 0.034$ & $0.252 \pm 0.030$ \\
DiFa \cite{zhang2022towards} & 30M & 1.80GB & $0.719$ & $0.226$ & $0.699$ & $0.263$ & $0.734 \pm 0.047$ & $0.215 \pm 0.038$ \\
TargetCLIP \cite{chefer2022image} & 9.0K & 420KB & $0.474$ & $0.306$ & $0.502$ & $0.333$ & $0.491 \pm 0.043$ & $0.322 \pm 0.015$ \\
DomMod (DiFa) \cite{alanov2022hyperdomainnet} & 6.0K & 280KB & $0.705$ & $0.250$ & $0.625$ & $0.294$ & $0.679 \pm 0.049$ & $0.253 \pm 0.037$ \\
\midrule
\textbf{StyleSpace (DiFa)} & 6.0K & 280KB & $0.672$ & $0.296$ & $0.627$ & $0.308$ & $0.644 \pm 0.041$ & $0.298 \pm 0.025$ \\
\textbf{StyleSpaceSparse (DiFa)} & \textbf{1.2K} & \textbf{56.4KB} & $0.659$ & $0.303$ & $0.617$ & $0.304$ & $0.638 \pm 0.046$ & $0.299 \pm 0.026$ \\
    \midrule
    \bottomrule
  \end{tabular}
\end{table*}

\textbf{Properties of the StyleDomain directions.} We investigate StyleDomain directions that adapt the generator to similar domains and explore surprising properties. The first one is mixability, i.e. we discover that StyleDomain directions can be combined with each other. In particular, we can consider several directions that correspond to different similar domains and take their sum. The resulting direction will adapt the generator to the semantically mixed domain. We provide different examples of such combinations in \Cref{fig:style_domain_mixing} (see more results in \Cref{app:styledomain}). 

The second property is transferability, i.e. we can transfer StyleDomain directions between different StyleGAN2 models.
In particular, let us consider the base generator $G_{\theta}$ pretrained on realistic faces and the generator $G_{\theta}^{Dog}$ fine-tuned to domain of dogs in the Full parameterization. We verify that we can apply StyleDomain directions to $G_{\theta}^{Dog}$ which were optimized for $G_{\theta}$. Specifically, we take StyleDomain directions that were optimized for $G_{\theta}$ to adapt it to different text-based and one-shot image-based domains (e.g. Pixar, Disney, etc.). Next, we apply these directions to generators that were fine-tuned from $G_{\theta}$ to other domains (e.g. Dogs, Cats). We provide results of this experiment in \Cref{fig:style_domain_transfer} (see more results in \Cref{app:styledomain}). 

Finally, we check that StyleDomain directions can be successfully combined with latent controls for image editing. For a more detailed exploration of this property, see in \Cref{app:styledomain}. 

%\paragraph{Revisit questions from StyleAlign paper}

\section{Experiments}
\label{sec:experiments}
\textbf{One-shot domain adaptation.} 
For image-based one-shot adaptation, we consider the main baselines: TargetCLIP \cite{chefer2022image}, JoJoGAN \cite{chong2022jojogan}, MTG \cite{zhu2021mind}, GOSA \cite{zhang2022generalized}, DiFa \cite{zhang2022towards}, DomMod \cite{alanov2022hyperdomainnet}. We apply our parameterizations StyleSpace and StyleSpaceSparse to DiFa model because this configuration achieves the highest performance (experiments with other base models can be found in \Cref{app:one_shot_adaptation}). Similarly, we apply DomMod parameterization to the DiFa model. We use StyleGAN2 as the GAN architecture and FFHQ as the source domain. For fair comparison, we strictly follow baseline default configurations in the choice of source-target adaptation setups and hyper-parameters. As target domain we use a wide range of different style images from baseline works. The full list of target images can be found in \Cref{app:one_shot_adaptation}.

\begin{figure}[!b]
  \centering
%   \begin{minipage}{0.5\textwidth}
%   \includegraphics[width=\textwidth]{imgs/ukiyo-e_botero_werewolf.pdf}
%   \end{minipage}\hfill
%   \begin{minipage}{0.5\textwidth}
  \includegraphics[width=0.5\textwidth]{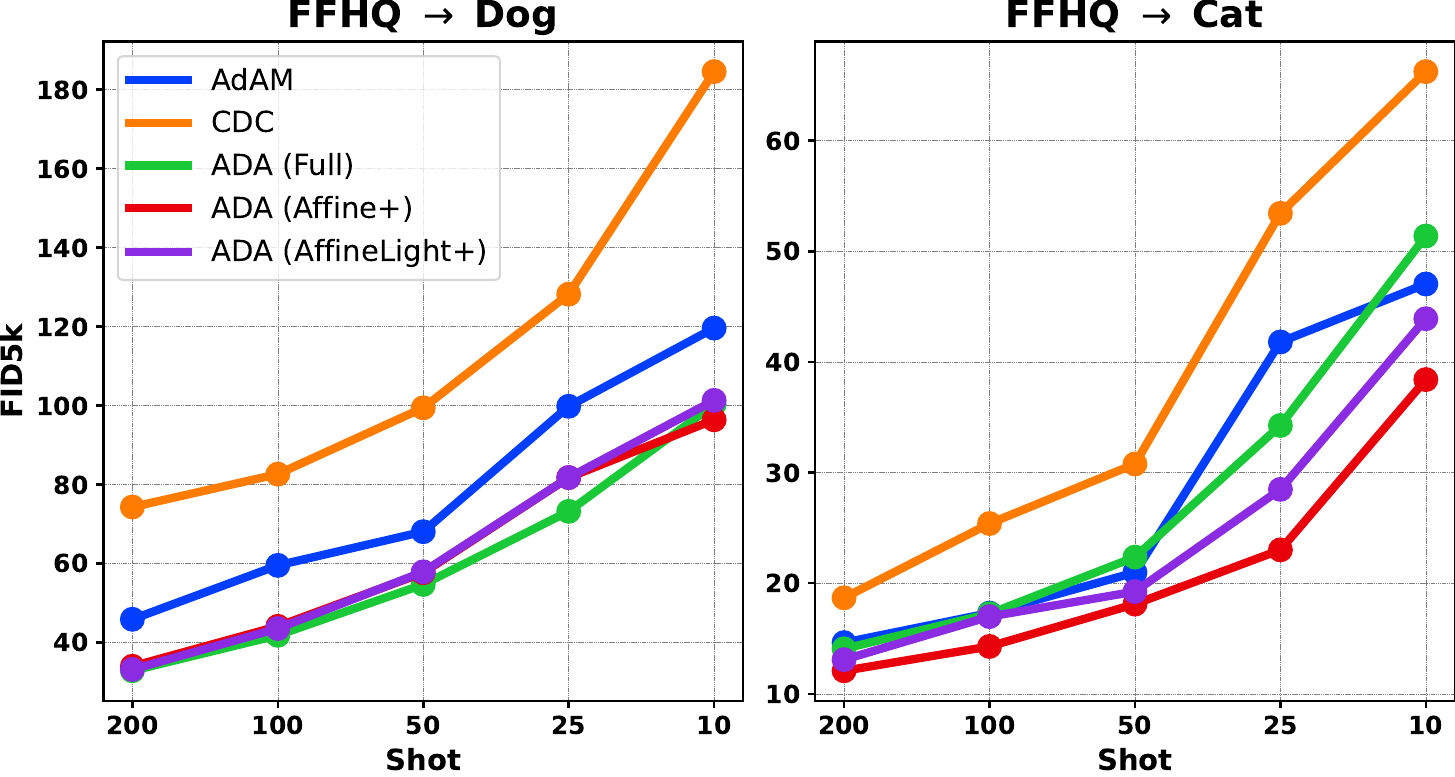}
%   \end{minipage}
  \caption{Few-shot training results for different number of shots. Proposed ADA (Affine+) and ADA (AffineLight$+$) show uniformly better performance than baselines.}
  \label{fig:few_shot_comparsion}
\end{figure}

We provide both quantitative and qualitative results in \Cref{table:one_shot_comparison} and \Cref{fig:one_shot_comparison}, correspondingly. More results and samples can be found in \Cref{app:one_shot_adaptation}. We observe that the DiFa model achieves the best Quality metric but has low Diversity compared to other methods. Our parameterizations applied to the DiFa model balance its performance in terms of these metrics. So, StyleSpace (DiFa) and StyleSpaceSparse (DiFa) achieves uniformly better results than JoJoGAN, MTG, GOSA baselines. DomMod parameterization also show good performance, but it is comparable with our StyleSpaceSparse parameterization that has 5 times less parameters. We also report the overall memory needed for keeping adapted generators for 12 domains. We observe that StyleSpaceSparse requires significantly less space than other models. It will be especially important if we scale the number of target domains to thousands and more. We note that TargetCLIP that also has small number of trainable parameters show very poor results visually and in terms of Quality metric. It has a high Diversity only because its generated images are very close to the original ones. To provide more comprehensive comparison we additionally conduct user studies and present results in \Cref{app:user_study}.

\textbf{Few-shot domain adaptation.} As main baselines for this task, we consider the vanilla StyleGAN-ADA \cite{karras2020training} (we will denote it as ADA), CDC \cite{ojha2021few} and recent SOTA method AdAM \cite{zhao2022few}. We compare our parameterizations Affine$+$ and AffineLight$+$ applied to ADA with these baselines on two datasets: Dogs and Cats \cite{choi2020stargan}. We study the efficiency of these methods with different numbers of available target samples. For fair comparison, we rigorously follow the training setups of all baselines. We note that we use the same number of training iterations and the same batch size for all methods. We observe that in the AdAM paper \cite{zhao2022few} the number of training iterations is set to a small value (less than 10K) and it causes underfitting of the vanilla ADA method. Therefore, we increase this number for all methods to 50K and use the same batch size of 4. Then, for each method, we report the best FID value that it obtains during training. See more details about training in \Cref{app:few_shot_adaptation}. 

We report results for few-shot training for different numbers of shots in \Cref{fig:few_shot_comparsion} and separately for 10 shots in \Cref{table:few_shot_comparison}. More results and samples can be found in \Cref{app:few_shot_adaptation}. Firstly, we observe that the AdAM method shows results not better than the vanilla ADA (Full) when trained for sufficient number of iterations. Secondly, we see that ADA (Affine+) and ADA (AffineLight+) based on our parameterizations achieve better results uniformly for all numbers of shots. It shows that these parameterizations is especially suitable for low data regimes. 

\begin{table}[!t]
\centering
\caption{Results for few-shot training with 10-shots. Proposed ADA (Affine+) and ADA (AffineLight$+$) achieve better performance.}
  \label{table:few_shot_comparison}
  \begin{tabular}{ llll }
    \toprule
    & &  \multicolumn{2}{c}{\textbf{Domains (10-shots)}} \\
\cmidrule(lr){3-4}
Method & Size & Cat & Dog \\
\midrule\midrule
    CDC \cite{ojha2021few}               & 30M  & $66.24$ & $184.56$  \\
    AdAM \cite{zhao2022few}              & 19M  & $47.05$ & $119.61$  \\
    ADA (Full) \cite{karras2020training} & 30M  & $51.38$ & $100.25$  \\
    \midrule
    \textbf{ADA (Affine+) }                       & 5.1M  & \textbf{38.40} & \textbf{96.38} \\
    \textbf{ADA (AffineLight$+$)}                 & \textbf{0.6M} & $43.91$ & $101.31$  \\
    \midrule
    \bottomrule
  \end{tabular}
  \vspace{-0.5cm}
\end{table}

\textbf{Cross-domain image translation.} We consider two setups of standard image-to-image problem. In the first one, we translate images from the source domain to the target domain unconditionally. In the second setup, we perform a reference-based translation, where the resulting image combines the pose and the shape of the source image with the style from a reference image. Details of these experiments can be found in \Cref{app:i2i}.

\textbf{Cross-domain image morphing.} Cross-domain morphing is a smooth transition between two images from different domains. This task is known as challenging \cite{aberman2018neural, fish2020image} and it is successfully tackled in the work \cite{wu2021stylealign} using aligned StyleGAN2 models. The idea is to interpolate between aligned generator weights to obtain a smooth transition between domains. We propose more complex image morphing by utilizing the transferability of StyleDomain directions. For example, we can apply a direction that stands for Sketch or Pixar style to Dogs domain to obtain a smooth transition between sketchy and pixar-like dog (see \Cref{fig:teaser}). See many examples of such complex cross-domain morphing in \Cref{app:morphing}. 

\section{Conclusion}
In this paper, we provide an extensive analysis of the process of StyleGAN domain adaptation. We reveal the sufficient components of the generator for successful adaptation depending on the similarity between the source and target domains. We discover the ability of StyleSpace to change the domain of generated images by StyleDomain directions. We also propose new efficient parameterizations Affine+ and AffineLight+ for few-shot adaptation that outperform existing baselines. Further, we explore and leverage the properties of StyleDomain directions. We believe that our investigation can attract more attention to the exploration of new and interesting properties of StyleSpace. 

\section{Acknowledgments}
The results of sections 1, 2, 3 were obtained by Aibek Alanov and Dmitry Vetrov with the support of the grant for research centers in the field of AI provided by the Analytical Center for the Government of the Russian Federation (ACRF) in accordance with the agreement on the provision of subsidies (identifier of the agreement 000000D730321P5Q0002) and the agreement with HSE University  No. 70-2021-00139.
This research was supported in part through computational resources of HPC facilities at HSE University.

{\small
\bibliographystyle{ieee_fullname}
\bibliography{ref}

\begin{thebibliography}{10}\itemsep=-1pt

\bibitem{abdal2019image2stylegan}
Rameen Abdal, Yipeng Qin, and Peter Wonka.
\newblock Image2stylegan: How to embed images into the stylegan latent space?
\newblock In {\em Proceedings of the IEEE/CVF International Conference on
  Computer Vision}, pages 4432--4441, 2019.

\bibitem{abdal2021styleflow}
Rameen Abdal, Peihao Zhu, Niloy~J Mitra, and Peter Wonka.
\newblock Styleflow: Attribute-conditioned exploration of stylegan-generated
  images using conditional continuous normalizing flows.
\newblock {\em ACM Transactions on Graphics (ToG)}, 40(3):1--21, 2021.

\bibitem{aberman2018neural}
Kfir Aberman, Jing Liao, Mingyi Shi, Dani Lischinski, Baoquan Chen, and Daniel
  Cohen-Or.
\newblock Neural best-buddies: Sparse cross-domain correspondence.
\newblock {\em ACM Transactions on Graphics (TOG)}, 37(4):1--14, 2018.

\bibitem{alanov2022hyperdomainnet}
Aibek Alanov, Vadim Titov, and Dmitry Vetrov.
\newblock Hyperdomainnet: Universal domain adaptation for generative
  adversarial networks.
\newblock {\em arXiv preprint arXiv:2210.08884}, 2022.

\bibitem{brock2018large}
Andrew Brock, Jeff Donahue, and Karen Simonyan.
\newblock Large scale gan training for high fidelity natural image synthesis.
\newblock {\em arXiv preprint arXiv:1809.11096}, 2018.

\bibitem{chan2021glean}
Kelvin~CK Chan, Xintao Wang, Xiangyu Xu, Jinwei Gu, and Chen~Change Loy.
\newblock Glean: Generative latent bank for large-factor image
  super-resolution.
\newblock In {\em Proceedings of the IEEE/CVF conference on computer vision and
  pattern recognition}, pages 14245--14254, 2021.

\bibitem{chefer2022image}
Hila Chefer, Sagie Benaim, Roni Paiss, and Lior Wolf.
\newblock Image-based clip-guided essence transfer.
\newblock In {\em Computer Vision--ECCV 2022: 17th European Conference, Tel
  Aviv, Israel, October 23--27, 2022, Proceedings, Part XIII}, pages 695--711.
  Springer, 2022.

\bibitem{choi2020stargan}
Yunjey Choi, Youngjung Uh, Jaejun Yoo, and Jung-Woo Ha.
\newblock Stargan v2: Diverse image synthesis for multiple domains.
\newblock In {\em Proceedings of the IEEE/CVF conference on computer vision and
  pattern recognition}, pages 8188--8197, 2020.

\bibitem{chong2022jojogan}
Min~Jin Chong and David Forsyth.
\newblock Jojogan: One shot face stylization.
\newblock In {\em Computer Vision--ECCV 2022: 17th European Conference, Tel
  Aviv, Israel, October 23--27, 2022, Proceedings, Part XVI}, pages 128--152.
  Springer, 2022.

\bibitem{fish2020image}
Noa Fish, Richard Zhang, Lilach Perry, Daniel Cohen-Or, Eli Shechtman, and
  Connelly Barnes.
\newblock Image morphing with perceptual constraints and stn alignment.
\newblock In {\em Computer Graphics Forum}, volume~39, pages 303--313. Wiley
  Online Library, 2020.

\bibitem{gal2022stylegan}
Rinon Gal, Or Patashnik, Haggai Maron, Amit~H Bermano, Gal Chechik, and Daniel
  Cohen-Or.
\newblock Stylegan-nada: Clip-guided domain adaptation of image generators.
\newblock {\em ACM Transactions on Graphics (TOG)}, 41(4):1--13, 2022.

\bibitem{goodfellow2014generative}
Ian Goodfellow, Jean Pouget-Abadie, Mehdi Mirza, Bing Xu, David Warde-Farley,
  Sherjil Ozair, Aaron Courville, and Yoshua Bengio.
\newblock Generative adversarial nets.
\newblock {\em Advances in neural information processing systems}, 27, 2014.

\bibitem{guan2020collaborative}
Shanyan Guan, Ying Tai, Bingbing Ni, Feida Zhu, Feiyue Huang, and Xiaokang
  Yang.
\newblock Collaborative learning for faster stylegan embedding.
\newblock {\em arXiv preprint arXiv:2007.01758}, 2020.

\bibitem{harkonen2020ganspace}
Erik H{\"a}rk{\"o}nen, Aaron Hertzmann, Jaakko Lehtinen, and Sylvain Paris.
\newblock Ganspace: Discovering interpretable gan controls.
\newblock {\em Advances in Neural Information Processing Systems},
  33:9841--9850, 2020.

\bibitem{heusel2017gans}
Martin Heusel, Hubert Ramsauer, Thomas Unterthiner, Bernhard Nessler, and Sepp
  Hochreiter.
\newblock Gans trained by a two time-scale update rule converge to a local nash
  equilibrium.
\newblock {\em Advances in neural information processing systems}, 30, 2017.

\bibitem{huang2021unsupervised}
Jialu Huang, Jing Liao, and Sam Kwong.
\newblock Unsupervised image-to-image translation via pre-trained stylegan2
  network.
\newblock {\em IEEE Transactions on Multimedia}, 24:1435--1448, 2021.

\bibitem{jahanian2019steerability}
Ali Jahanian, Lucy Chai, and Phillip Isola.
\newblock On the" steerability" of generative adversarial networks.
\newblock {\em arXiv preprint arXiv:1907.07171}, 2019.

\bibitem{karras2020training}
Tero Karras, Miika Aittala, Janne Hellsten, Samuli Laine, Jaakko Lehtinen, and
  Timo Aila.
\newblock Training generative adversarial networks with limited data.
\newblock {\em Advances in Neural Information Processing Systems},
  33:12104--12114, 2020.

\bibitem{karras2021alias}
Tero Karras, Miika Aittala, Samuli Laine, Erik H{\"a}rk{\"o}nen, Janne
  Hellsten, Jaakko Lehtinen, and Timo Aila.
\newblock Alias-free generative adversarial networks.
\newblock {\em Advances in Neural Information Processing Systems}, 34, 2021.

\bibitem{karras2019style}
Tero Karras, Samuli Laine, and Timo Aila.
\newblock A style-based generator architecture for generative adversarial
  networks.
\newblock In {\em Proceedings of the IEEE/CVF conference on computer vision and
  pattern recognition}, pages 4401--4410, 2019.

\bibitem{karras2020analyzing}
Tero Karras, Samuli Laine, Miika Aittala, Janne Hellsten, Jaakko Lehtinen, and
  Timo Aila.
\newblock Analyzing and improving the image quality of stylegan.
\newblock In {\em Proceedings of the IEEE/CVF conference on computer vision and
  pattern recognition}, pages 8110--8119, 2020.

\bibitem{li2020few}
Yijun Li, Richard Zhang, Jingwan Lu, and Eli Shechtman.
\newblock Few-shot image generation with elastic weight consolidation.
\newblock {\em arXiv preprint arXiv:2012.02780}, 2020.

\bibitem{liu2020towards}
Bingchen Liu, Yizhe Zhu, Kunpeng Song, and Ahmed Elgammal.
\newblock Towards faster and stabilized gan training for high-fidelity few-shot
  image synthesis.
\newblock In {\em International Conference on Learning Representations}, 2020.

\bibitem{luo2021time}
Xuan Luo, Xuaner Zhang, Paul Yoo, Ricardo Martin-Brualla, Jason Lawrence, and
  Steven~M Seitz.
\newblock Time-travel rephotography.
\newblock {\em ACM Transactions on Graphics (TOG)}, 40(6):1--12, 2021.

\bibitem{mo2020freeze}
Sangwoo Mo, Minsu Cho, and Jinwoo Shin.
\newblock Freeze the discriminator: a simple baseline for fine-tuning gans.
\newblock {\em arXiv preprint arXiv:2002.10964}, 2020.

\bibitem{nilsback2006visual}
M-E Nilsback and Andrew Zisserman.
\newblock A visual vocabulary for flower classification.
\newblock In {\em 2006 IEEE Computer Society Conference on Computer Vision and
  Pattern Recognition (CVPR'06)}, volume~2, pages 1447--1454. IEEE, 2006.

\bibitem{noguchi2019image}
Atsuhiro Noguchi and Tatsuya Harada.
\newblock Image generation from small datasets via batch statistics adaptation.
\newblock In {\em Proceedings of the IEEE/CVF International Conference on
  Computer Vision}, pages 2750--2758, 2019.

\bibitem{ojha2021few}
Utkarsh Ojha, Yijun Li, Jingwan Lu, Alexei~A Efros, Yong~Jae Lee, Eli
  Shechtman, and Richard Zhang.
\newblock Few-shot image generation via cross-domain correspondence.
\newblock In {\em Proceedings of the IEEE/CVF Conference on Computer Vision and
  Pattern Recognition}, pages 10743--10752, 2021.

\bibitem{patashnik2021styleclip}
Or Patashnik, Zongze Wu, Eli Shechtman, Daniel Cohen-Or, and Dani Lischinski.
\newblock Styleclip: Text-driven manipulation of stylegan imagery.
\newblock In {\em Proceedings of the IEEE/CVF International Conference on
  Computer Vision}, pages 2085--2094, 2021.

\bibitem{pidhorskyi2020adversarial}
Stanislav Pidhorskyi, Donald~A Adjeroh, and Gianfranco Doretto.
\newblock Adversarial latent autoencoders.
\newblock In {\em Proceedings of the IEEE/CVF Conference on Computer Vision and
  Pattern Recognition}, pages 14104--14113, 2020.

\bibitem{pinkney2020resolution}
Justin~NM Pinkney and Doron Adler.
\newblock Resolution dependent gan interpolation for controllable image
  synthesis between domains.
\newblock {\em arXiv preprint arXiv:2010.05334}, 2020.

\bibitem{radford2021learning}
Alec Radford, Jong~Wook Kim, Chris Hallacy, Aditya Ramesh, Gabriel Goh,
  Sandhini Agarwal, Girish Sastry, Amanda Askell, Pamela Mishkin, Jack Clark,
  et~al.
\newblock Learning transferable visual models from natural language
  supervision.
\newblock In {\em International Conference on Machine Learning}, pages
  8748--8763. PMLR, 2021.

\bibitem{richardson2021encoding}
Elad Richardson, Yuval Alaluf, Or Patashnik, Yotam Nitzan, Yaniv Azar, Stav
  Shapiro, and Daniel Cohen-Or.
\newblock Encoding in style: a stylegan encoder for image-to-image translation.
\newblock In {\em Proceedings of the IEEE/CVF conference on computer vision and
  pattern recognition}, pages 2287--2296, 2021.

\bibitem{robb2020few}
Esther Robb, Wen-Sheng Chu, Abhishek Kumar, and Jia-Bin Huang.
\newblock Few-shot adaptation of generative adversarial networks.
\newblock {\em arXiv preprint arXiv:2010.11943}, 2020.

\bibitem{shen2020interpreting}
Yujun Shen, Jinjin Gu, Xiaoou Tang, and Bolei Zhou.
\newblock Interpreting the latent space of gans for semantic face editing.
\newblock In {\em Proceedings of the IEEE/CVF conference on computer vision and
  pattern recognition}, pages 9243--9252, 2020.

\bibitem{song2021agilegan}
Guoxian Song, Linjie Luo, Jing Liu, Wan-Chun Ma, Chunpong Lai, Chuanxia Zheng,
  and Tat-Jen Cham.
\newblock Agilegan: stylizing portraits by inversion-consistent transfer
  learning.
\newblock {\em ACM Transactions on Graphics (TOG)}, 40(4):1--13, 2021.

\bibitem{tewari2020stylerig}
Ayush Tewari, Mohamed Elgharib, Gaurav Bharaj, Florian Bernard, Hans-Peter
  Seidel, Patrick P{\'e}rez, Michael Zollhofer, and Christian Theobalt.
\newblock Stylerig: Rigging stylegan for 3d control over portrait images.
\newblock In {\em Proceedings of the IEEE/CVF Conference on Computer Vision and
  Pattern Recognition}, pages 6142--6151, 2020.

\bibitem{tov2021designing}
Omer Tov, Yuval Alaluf, Yotam Nitzan, Or Patashnik, and Daniel Cohen-Or.
\newblock Designing an encoder for stylegan image manipulation.
\newblock {\em ACM Transactions on Graphics (TOG)}, 40(4):1--14, 2021.

\bibitem{tran2021data}
Ngoc-Trung Tran, Viet-Hung Tran, Ngoc-Bao Nguyen, Trung-Kien Nguyen, and
  Ngai-Man Cheung.
\newblock On data augmentation for gan training.
\newblock {\em IEEE Transactions on Image Processing}, 30:1882--1897, 2021.

\bibitem{tseng2021regularizing}
Hung-Yu Tseng, Lu Jiang, Ce Liu, Ming-Hsuan Yang, and Weilong Yang.
\newblock Regularizing generative adversarial networks under limited data.
\newblock In {\em Proceedings of the IEEE/CVF Conference on Computer Vision and
  Pattern Recognition}, pages 7921--7931, 2021.

\bibitem{wang2021towards}
Xintao Wang, Yu Li, Honglun Zhang, and Ying Shan.
\newblock Towards real-world blind face restoration with generative facial
  prior.
\newblock In {\em Proceedings of the IEEE/CVF Conference on Computer Vision and
  Pattern Recognition}, pages 9168--9178, 2021.

\bibitem{wang2020minegan}
Yaxing Wang, Abel Gonzalez-Garcia, David Berga, Luis Herranz, Fahad~Shahbaz
  Khan, and Joost van~de Weijer.
\newblock Minegan: effective knowledge transfer from gans to target domains
  with few images.
\newblock In {\em Proceedings of the IEEE/CVF Conference on Computer Vision and
  Pattern Recognition}, pages 9332--9341, 2020.

\bibitem{wu2021stylespace}
Zongze Wu, Dani Lischinski, and Eli Shechtman.
\newblock Stylespace analysis: Disentangled controls for stylegan image
  generation.
\newblock In {\em Proceedings of the IEEE/CVF Conference on Computer Vision and
  Pattern Recognition}, pages 12863--12872, 2021.

\bibitem{wu2021stylealign}
Zongze Wu, Yotam Nitzan, Eli Shechtman, and Dani Lischinski.
\newblock Stylealign: Analysis and applications of aligned stylegan models.
\newblock {\em arXiv preprint arXiv:2110.11323}, 2021.

\bibitem{yang2021data}
Ceyuan Yang, Yujun Shen, Yinghao Xu, and Bolei Zhou.
\newblock Data-efficient instance generation from instance discrimination.
\newblock {\em Advances in Neural Information Processing Systems},
  34:9378--9390, 2021.

\bibitem{yang2021gan}
Tao Yang, Peiran Ren, Xuansong Xie, and Lei Zhang.
\newblock Gan prior embedded network for blind face restoration in the wild.
\newblock In {\em Proceedings of the IEEE/CVF Conference on Computer Vision and
  Pattern Recognition}, pages 672--681, 2021.

\bibitem{yu2015lsun}
Fisher Yu, Ari Seff, Yinda Zhang, Shuran Song, Thomas Funkhouser, and Jianxiong
  Xiao.
\newblock Lsun: Construction of a large-scale image dataset using deep learning
  with humans in the loop.
\newblock {\em arXiv preprint arXiv:1506.03365}, 2015.

\bibitem{zhang2022towards}
Yabo Zhang, Yuxiang Wei, Zhilong Ji, Jinfeng Bai, Wangmeng Zuo, et~al.
\newblock Towards diverse and faithful one-shot adaption of generative
  adversarial networks.
\newblock In {\em Advances in Neural Information Processing Systems}, 2022.

\bibitem{zhang2022generalized}
Zicheng Zhang, Yinglu Liu, Congying Han, Tiande Guo, Ting Yao, and Tao Mei.
\newblock Generalized one-shot domain adaption of generative adversarial
  networks.
\newblock {\em arXiv preprint arXiv:2209.03665}, 2022.

\bibitem{zhao2020differentiable}
Shengyu Zhao, Zhijian Liu, Ji Lin, Jun-Yan Zhu, and Song Han.
\newblock Differentiable augmentation for data-efficient gan training.
\newblock {\em Advances in Neural Information Processing Systems},
  33:7559--7570, 2020.

\bibitem{zhao2022few}
Yunqing Zhao, Keshigeyan Chandrasegaran, Milad Abdollahzadeh, and Ngai-Man
  Cheung.
\newblock Few-shot image generation via adaptation-aware kernel modulation.
\newblock {\em arXiv preprint arXiv:2210.16559}, 2022.

\bibitem{zhao2020image}
Zhengli Zhao, Zizhao Zhang, Ting Chen, Sameer Singh, and Han Zhang.
\newblock Image augmentations for gan training.
\newblock {\em arXiv preprint arXiv:2006.02595}, 2020.

\bibitem{zhu2020domain}
Jiapeng Zhu, Yujun Shen, Deli Zhao, and Bolei Zhou.
\newblock In-domain gan inversion for real image editing.
\newblock In {\em European conference on computer vision}, pages 592--608.
  Springer, 2020.

\bibitem{zhu2016generative}
Jun-Yan Zhu, Philipp Kr{\"a}henb{\"u}hl, Eli Shechtman, and Alexei~A Efros.
\newblock Generative visual manipulation on the natural image manifold.
\newblock In {\em European conference on computer vision}, pages 597--613.
  Springer, 2016.

\bibitem{zhu2021mind}
Peihao Zhu, Rameen Abdal, John Femiani, and Peter Wonka.
\newblock Mind the gap: Domain gap control for single shot domain adaptation
  for generative adversarial networks.
\newblock {\em arXiv preprint arXiv:2110.08398}, 2021.

\end{thebibliography}
}

\clearpage
\onecolumn
\appendix

\section{Appendix}
\subsection{Domain Adaptation Losses}
\label{app:losses}
As we discuss in \Cref{sec:importance}, the optimization process for the domain adaptation is
\begin{gather}
    \mathcal{L}_D\left(\left\{G_{\theta}(s(z_i)) \right\}_{i=1}^K\right) \rightarrow \min\limits,
\end{gather}
where $\mathcal{L}_D$ is the domain adaptation loss that depends on the domain $D$. We describe each possible loss function in more detail. 

In the case of similar domains, we analyse text-based and one-shot image-based adaptation methods. For the text-based setting, we apply the optimization loss from StyleGAN-NADA \cite{gal2022stylegan} that equals
\begin{gather}
    \mathcal{L}_D\left(\left\{G_{\theta}(s(z_i)) \right\}_{i=1}^K\right) = \sum\limits_{i=1}^K \mathcal{L}_{direction}(G_{\theta}(s(z_i))), \\
    \text{ where } \; \mathcal{L}_{direction}(G_{\theta}(s(z_i))) = 1 - \dfrac{\Delta I_i^T \Delta T}{|\Delta I_i|\cdot |\Delta T|}, \label{eq:direction_loss} \\ 
    \Delta I_i = E_I(G_{\theta}(s(z_i))) - E_I(G_{\theta_0}(s(z_i))) \; (\theta_0 \text{ - pretrained weights}), \\
    \Delta T = E_T(t) - E_T(t_0) \; (t \text{ - text description of the target domain}, \: t_0 \text{ - text description of the source domain}), \\
    E_I, E_T \text{ - pretrained text and image CLIP \cite{radford2021learning} encoders, respectively}
\end{gather}

For the one-shot image-based setting, we utilize the loss function from DiFa \cite{zhang2022towards} that equals
\begin{gather}
    \mathcal{L}_D\left(\left\{G_{\theta}(s(z_i)) \right\}_{i=1}^K\right) = \mathcal{L}_{global}\left(\left\{G_{\theta}(s(z_i)) \right\}_{i=1}^K\right) + \lambda_{local}\mathcal{L}_{local}\left(\left\{G_{\theta}(s(z_i)) \right\}_{i=1}^K\right) + \lambda_{scc}\mathcal{L}_{scc}\left(\left\{G_{\theta}(s(z_i)) \right\}_{i=1}^K\right) \label{eq:mtg_loss}
\end{gather}
For more details about each loss term $\mathcal{L}_{global}, \mathcal{L}_{local}, \mathcal{L}_{scc}$ please refer to the original DiFa paper \cite{zhang2022towards}. 

In the few-shot setting for dissimilar domains, we utilize the standard StyleGAN-ADA method \cite{karras2020training} of fine-tuning GANs using differentiable augmentations \cite{karras2020training, tran2021data, zhao2020differentiable, zhao2020image}. The optimization loss equals
\begin{gather}
    \mathcal{L}_D\left(\left\{G_{\theta}(s(z_i)) \right\}_{i=1}^K\right) = \sum\limits_{i=1}^K -D_{\varphi}(T(G_{\theta}(s(z_i)))), \\
    \text{where } T \text{ - differentiable augmentation, } D_{\varphi} \text{ - discriminator that is trained using the following loss } \\
    \left[\sum\limits_{i=1}^B \max(0, 1 - D_{\varphi}(T(x_i))) +     \sum\limits_{i=1}^B \max(0, 1 + D_{\varphi}(T(G_{\theta}(s(z_i)))))\right] \rightarrow \min\limits_{\varphi}, \\
    \text{where } B \text{ - batch size}, x_1, \dots, x_B \text{ - real samples}, G_{\theta}(s(z_1))), \dots, G_{\theta}(s(z_B))) \text{ - generated samples}. 
\end{gather}

\FloatBarrier

\subsection{Analysis for One-Shot Domains}
\label{app:sim_dom}
\subsubsection{Implementation Details}
We implement our experiments using PyTorch\footnote{\hyperlink{https://pytorch.org}{https://pytorch.org}} deep learning framework. For StyleGAN2 \cite{karras2020analyzing} architecture in the case of similar domains, we use the popular PyTorch implementation \footnote{\hyperlink{https://github.com/rosinality/stylegan2- pytorch}{https://github.com/rosinality/stylegan2- pytorch}}. We attach all source code that reproduces our experiments for the setting with similar domains as a part of the supplementary material named "adaptation\_to\_one\_shot\_domains". We also provide configuration files to run each experiment.

% \subsubsection{Licenses and Data Privacy}
% Tables \ref{tbl:model_licenses}, \ref{tbl:data_licenses} provide sources and licenses of the models and datasets we used in our work. 

% \begin{table}[h]
%     \centering
%     \caption{Models used in our work, their sources and licenses.}
%     \begin{tabular}{l c c}
%         Model & Source & License \\ \hline
%         StyleGAN2 & \cite{karras2020analyzing} & \nvsrc \\ 
%         pSp & \cite{richardson2021encoding} & \mitlic \\
%         e4e & \cite{tov2021designing} & \mitlic \\
%         StyleCLIP & \cite{patashnik2021styleclip} & \mitlic \\
%         CLIP & \cite{radford2021learning} & \mitlic \\
%         StyleGAN2-pytorch & \cite{rosinalitySG2} & \mitlic \\
%         StyleGAN-ADA & \cite{karras2020training} & \href{https://nvlabs.github.io/stylegan2-ada-pytorch/license.html}{Nvidia Source Code License}
%     \end{tabular}\label{tbl:model_licenses}
% \end{table}

\subsubsection{Hyperparameters}
%Like in HyperDomainNet paper \cite{alanov2022hyperdomainnet} in Appendix A.3.1.

For the text-based domain adaptation we use the optimization loss from \Cref{eq:direction_loss}. As pretrained CLIP Vision-Transformer \cite{radford2021learning} we utilize "ViT-B/32", "ViT-B/16" models.
We use batch size of 4 and different number of iterations depending on the domain (from 250 to 300 iterations). As optimizer we apply the ADAM Optimizer with parameters we provide in \Cref{tab:hyper_opt_td_im2im}. 

For the one-shot image-based domain adaptation we apply the optimization loss from \Cref{eq:mtg_loss}. We set loss coefficients as $\lambda_{cw} = 0.5, \lambda_{rc} = 30, \lambda_{rr} = 10$. We use batch of 4 and 300 number of iterations. We utilize the ADAM Optimizer with parameters we provide in Table~\ref{tab:hyper_opt_td_im2im}.

\begin{table*}[!h]
    \centering
    \caption{ADAM optimizer hyperparameters for each setting.}
    \begin{tabular}{llll}
    \toprule
         Parameter Space & lr & betas & weight\_decay \\
         \midrule\midrule
         SyntConv & 0.002 & (0.0, 0.999) & 0 \\
         Full & 0.002 & (0.0, 0.999) & 0 \\
         Affine & 0.01 & (0.0, 0.999) & 0 \\
         Mapping & 0.3 & (0.0, 0.999) & 0 \\
         StyleSpace & 0.05 & (0.9, 0.999) & 0 \\
    \bottomrule
    \end{tabular}
    \label{tab:hyper_opt_td_im2im}
\end{table*}

\subsubsection{Quantitative Results}
%To provide quantitative comparison for one-shot domain adaptation (text based and image based) we use metrics "Quality" and "Diversity" which were proposed in \cite{alanov2022hyperdomainnet}.

For quantitative comparisons, we utilize Quality and Diversity metrics from \cite{alanov2022hyperdomainnet}. The Quality is measured as mean cosine similarity between CLIP embedding of the text description of the target domain and embeddings of generated images from the target domain. In the case of one-shot image-based domain adaptation, the embedding of the target domain is calculated as the CLIP embedding of the reference style image.
Diversity metric is estimated as mean pairwise cosine distance between CLIP embeddings of generated images from the target domain. As the CLIP image encoder we use only  ViT-L/14 that is not applied during training (in the training we use ViT-B/16, ViT-B/32 image encoders).

For every metric we set the number of synthesized images to \textbf{1000}. We evaluate each metric five times with different generated images to estimate error bars. However, we obtain that the error bar is less than $10^{-5}$ for all settings. So, we omit them in Tables.  

\begin{figure}[!h]
    \centering
  \includegraphics[width=\textwidth]{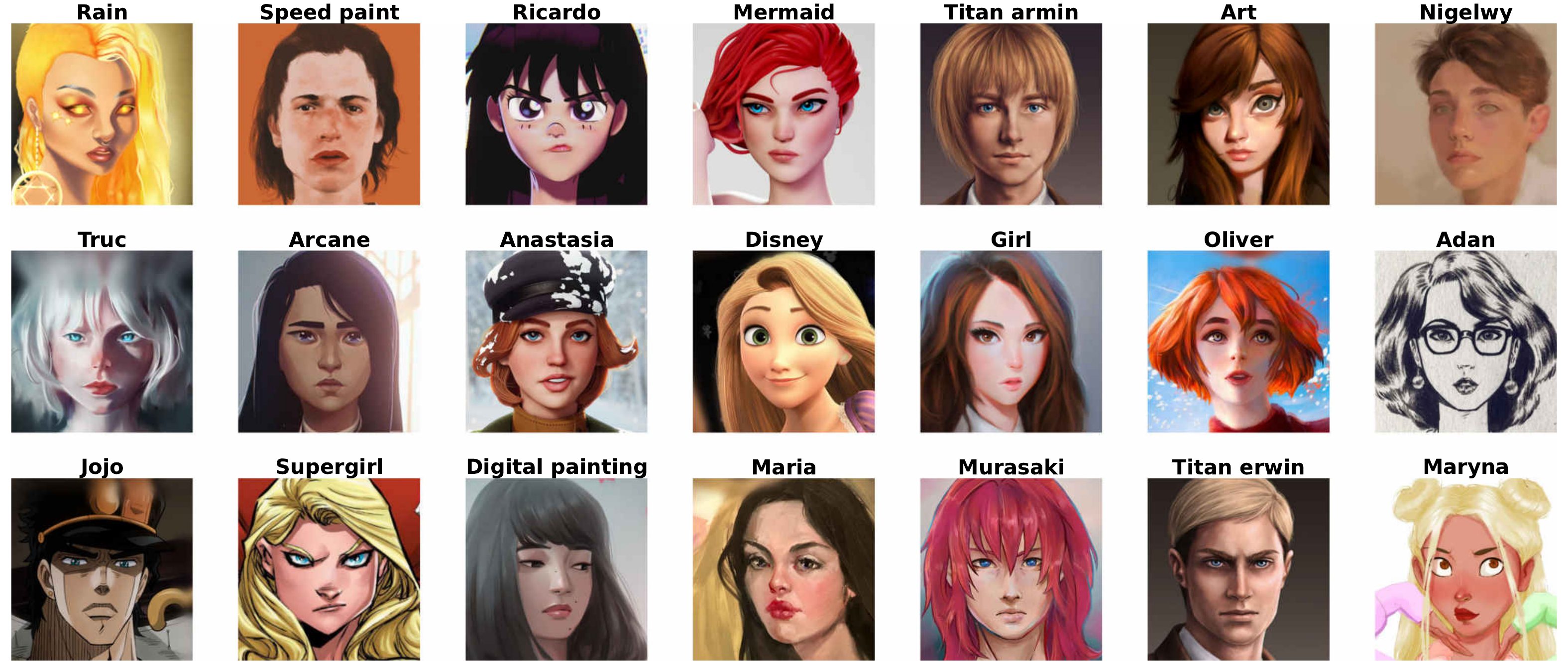}
  \caption{Correspondence between names and style images for one-shot image-based domain adaptation.}
  \label{fig:name_to_style_image_mapping}
  \vspace{-0.5cm}
\end{figure}

We provide calculated metrics in \Cref{table:sim_domains_sg_tuned_parts} for a wide range of domains (text-based and image-based). Please see \Cref{fig:name_to_style_image_mapping} for the correspondence between the name of style image and its appearance. From the results in \Cref{table:sim_domains_sg_tuned_parts} we see that all four parameterizations Full, SyntConv, Affine and Mapping achieve the comparable Quality but the Mapping has significantly lower Diversity. It indicates that the Mapping parameterization collapsed to one image that maximizes the CLIP score (it can be also seen from qualitative results, see \Cref{appx:sim_qual_res}). At the same time, we observe that Affine parameterization demonstrates comparable performance to Full and SyntConv in terms of both Quality and Diversity. It confirms our results from the main part in \Cref{sec:importance} that Affine is sufficient for similar domains. 

%Calculated metrics for different parameterizations and domains are provided in Table~\ref{table:sim_domains_sg_tuned_parts}.

%(StyleGAN optimization comparison) and Table~\ref{table:sim_domains_stylespace} (StyleSpace offsets). To calculate std over each metric several runs were launched. For each target domain std is $< 1e-5$. Therefore variance is not pointed out in tables.

%For image based adaptation corresponding style images could be found on Figure~\ref{fig:name_to_style_image_mapping}.

\addtolength{\tabcolsep}{-3pt}  
\begin{table*}[!h]
\centering
\caption{Quality and Diversity metrics \cite{alanov2022hyperdomainnet} for text-based and one-shot image-based domain adaptation. Affine parameterization achieves comparable results as Full and SyntConv parameterizations.}
	\label{table:sim_domains_sg_tuned_parts}
\begin{tabular}{p{0.15\textwidth}cccccccc}
\toprule
 & \multicolumn{4}{c}{\textbf{Quality}} & \multicolumn{4}{c}{\textbf{Diversity}} \\
 \cmidrule(lr){2-5} \cmidrule(lr){6-9}
 & Full & SyntConv & Affine & Mapping & Full & SyntConv & Affine & Mapping \\
\midrule\midrule
\textbf{Size} & 30.3M & 23.6M & 4.6M & 2.1M & 30.3M & 23.6M & 4.6M & 2.1M \\
\midrule
\textbf{Text Domains} \\
\midrule
Anime Painting & $0.25$ & $0.25$ & $0.26$ & $0.22$ & $0.25$ & $0.25$ & $0.23$ & $0.13$ \\
Pixar Render & $0.27$ & $0.26$ & $0.27$ & $0.26$ & $0.21$ & $0.21$ & $0.20$ & $0.09$ \\
Sketch & $0.21$ & $0.19$ & $0.19$ & $0.18$ & $0.30$ & $0.31$ & $0.29$ & $0.14$ \\
Ukiyo-e Painting & $0.22$ & $0.22$ & $0.22$ & $0.23$ & $0.22$ & $0.22$ & $0.23$ & $0.07$ \\
Botero Painting & $0.31$ & $0.29$ & $0.29$ & $0.23$ & $0.23$ & $0.24$ & $0.24$ & $0.11$ \\
Pop Art & $0.25$ & $0.24$ & $0.24$ & $0.26$ & $0.27$ & $0.27$ & $0.33$ & $0.07$ \\
Werewolf & $0.26$ & $0.25$ & $0.22$ & $0.27$ & $0.13$ & $0.14$ & $0.17$ & $0.07$ \\
Zombie & $0.24$ & $0.23$ & $0.24$ & $0.24$ & $0.12$ & $0.12$ & $0.13$ & $0.07$ \\
The Joker & $0.25$ & $0.24$ & $0.24$ & $0.25$ & $0.17$ & $0.17$ & $0.17$ & $0.13$ \\
Disney Princess & $0.23$ & $0.22$ & $0.23$ & $0.20$ & $0.24$ & $0.28$ & $0.28$ & $0.11$ \\
Cubism & $0.23$ & $0.21$ & $0.19$ & $0.19$ & $0.22$ & $0.22$ & $0.25$ & $0.07$ \\
Tolkien Elf & $0.27$ & $0.26$ & $0.28$ & $0.28$ & $0.22$ & $0.23$ & $0.22$ & $0.10$ \\
Impressionism & $0.27$ & $0.25$ & $0.25$ & $0.24$ & $0.25$ & $0.24$ & $0.25$ & $0.09$ \\
Claude Monet & $0.21$ & $0.19$ & $0.19$ & $0.22$ & $0.24$ & $0.25$ & $0.27$ & $0.07$ \\
Modigliani & $0.27$ & $0.25$ & $0.23$ & $0.26$ & $0.20$ & $0.21$ & $0.24$ & $0.12$ \\
The Thanos & $0.23$ & $0.22$ & $0.23$ & $0.28$ & $0.29$ & $0.31$ & $0.31$ & $0.12$ \\
Edvard Munch Painting & $0.25$ & $0.23$ & $0.24$ & $0.22$ & $0.20$ & $0.22$ & $0.24$ & $0.09$ \\
Dali Painting & $0.26$ & $0.24$ & $0.25$ & $0.25$ & $0.22$ & $0.23$ & $0.22$ & $0.14$ \\
\midrule
\textbf{Over 18 domains} & $0.24 \pm 0.02$ & $0.23 \pm 0.02$ & $0.23 \pm 0.02$ & $0.23 \pm 0.02$ & $0.22 \pm 0.04$ & $0.22 \pm 0.04$ & $0.23 \pm 0.04$ & $0.099 \pm 0.02$ \\
\midrule
\textbf{Image Domains} \\
\midrule
Adan & $0.76$ & $0.74$ & $0.56$ & $0.72$ & $0.17$ & $0.20$ & $0.35$ & $0.06$ \\
Arcane & $0.70$ & $0.72$ & $0.59$ & $0.63$ & $0.23$ & $0.22$ & $0.33$ & $0.09$ \\
Art & $0.77$ & $0.79$ & $0.68$ & $0.77$ & $0.20$ & $0.18$ & $0.30$ & $0.04$ \\
Anastasia & $0.66$ & $0.65$ & $0.55$ & $0.58$ & $0.24$ & $0.28$ & $0.36$ & $0.14$ \\
Digital painting & $0.71$ & $0.71$ & $0.60$ & $0.56$ & $0.22$ & $0.22$ & $0.33$ & $0.36$ \\
Disney & $0.70$ & $0.71$ & $0.56$ & $0.72$ & $0.26$ & $0.26$ & $0.36$ & $0.08$ \\
Girl & $0.80$ & $0.81$ & $0.65$ & $0.74$ & $0.17$ & $0.17$ & $0.32$ & $0.06$ \\
Jojo & $0.71$ & $0.68$ & $0.55$ & $0.70$ & $0.21$ & $0.20$ & $0.35$ & $0.11$ \\
Maria & $0.68$ & $0.68$ & $0.61$ & $0.63$ & $0.26$ & $0.27$ & $0.32$ & $0.14$ \\
Maryna & $0.67$ & $0.66$ & $0.48$ & $0.57$ & $0.24$ & $0.25$ & $0.38$ & $0.09$ \\
Mermaid & $0.78$ & $0.79$ & $0.65$ & $0.50$ & $0.18$ & $0.17$ & $0.32$ & $0.35$ \\
Murasaki & $0.71$ & $0.70$ & $0.52$ & $0.70$ & $0.22$ & $0.23$ & $0.37$ & $0.11$ \\
Nigelwy & $0.70$ & $0.67$ & $0.58$ & $0.57$ & $0.25$ & $0.27$ & $0.34$ & $0.15$ \\
Oliver & $0.68$ & $0.67$ & $0.53$ & $0.67$ & $0.24$ & $0.26$ & $0.37$ & $0.13$ \\
Rain & $0.75$ & $0.75$ & $0.61$ & $0.72$ & $0.20$ & $0.21$ & $0.33$ & $0.09$ \\
Ricardo & $0.66$ & $0.63$ & $0.52$ & $0.68$ & $0.25$ & $0.27$ & $0.37$ & $0.11$ \\
Speed paint & $0.76$ & $0.75$ & $0.66$ & $0.51$ & $0.24$ & $0.26$ & $0.31$ & $0.35$ \\
Supergirl & $0.82$ & $0.83$ & $0.53$ & $0.66$ & $0.15$ & $0.14$ & $0.38$ & $0.07$ \\
Titan armin & $0.72$ & $0.56$ & $0.60$ & $0.44$ & $0.23$ & $0.36$ & $0.33$ & $0.35$ \\
Titan erwin & $0.72$ & $0.74$ & $0.65$ & $0.65$ & $0.23$ & $0.21$ & $0.31$ & $0.10$ \\
Truc & $0.73$ & $0.72$ & $0.60$ & $0.71$ & $0.23$ & $0.24$ & $0.33$ & $0.07$ \\
\midrule
\textbf{Over 21 domains} & $0.72 \pm 0.04$ & $0.71 \pm 0.06$ & $0.58 \pm 0.05$ & $0.64 \pm 0.09$ & $0.22 \pm 0.03$ & $0.23 \pm 0.05$ & $0.34 \pm 0.02$ & $0.15 \pm 0.10$ \\
    \midrule\bottomrule
    \end{tabular}
  \vspace{-0.5cm}
\end{table*}
\addtolength{\tabcolsep}{3pt}

\subsubsection{Qualitative Results}
\label{appx:sim_qual_res}

\begin{figure*}[!h]
  \centering
  \includegraphics[width=\textwidth]{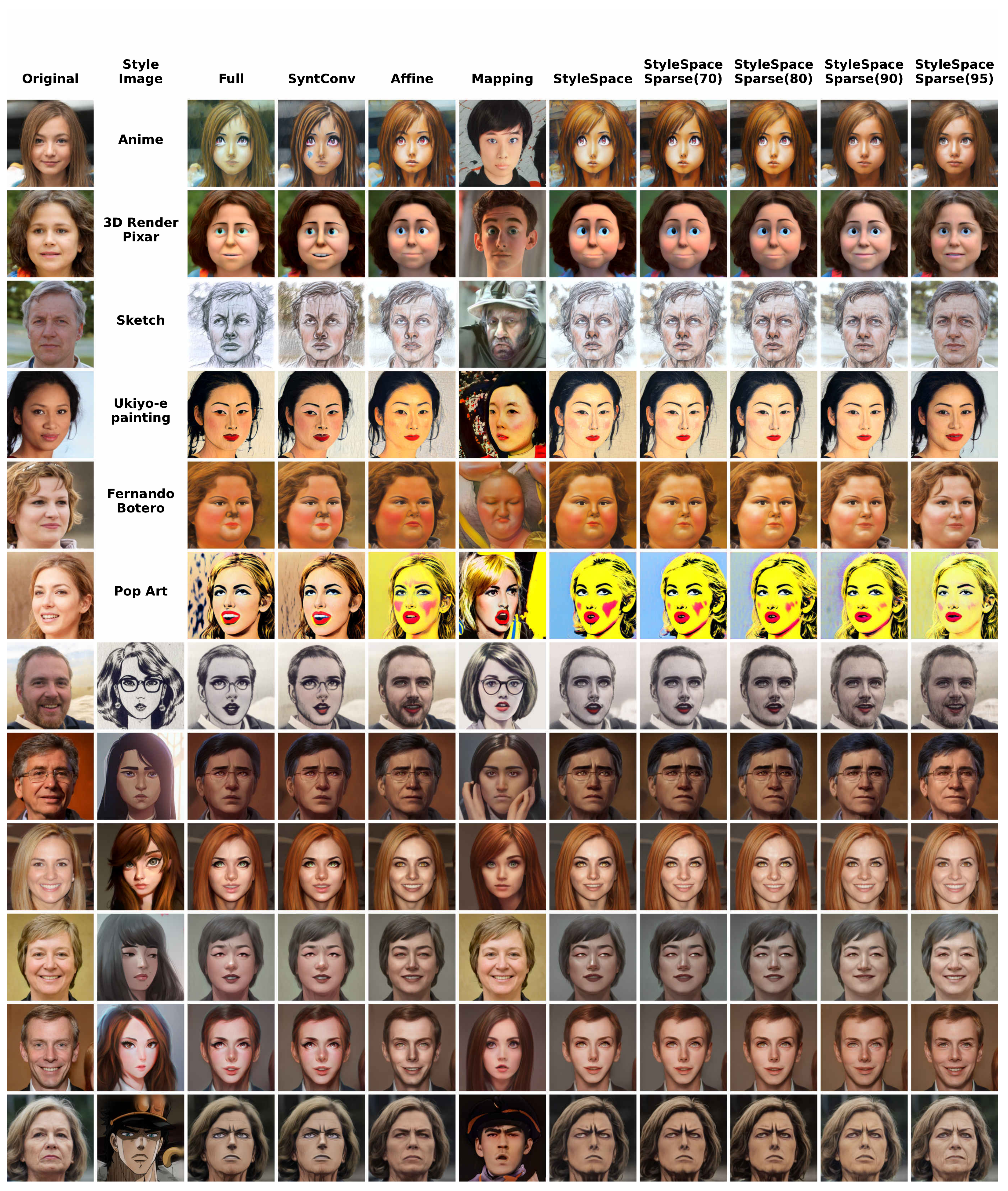}
  \caption{Text-based and one-shot image-based domain adaptation for different parameterizations. Affine, StyleSpace and StyleSparse (70, 80) achieve visual quality on par with Full and SyntConv parameterizations.}
  \label{fig:a2_1}
  \vspace{-0.5cm}
\end{figure*}

\begin{figure*}[!h]
  \centering
  \includegraphics[width=\textwidth]{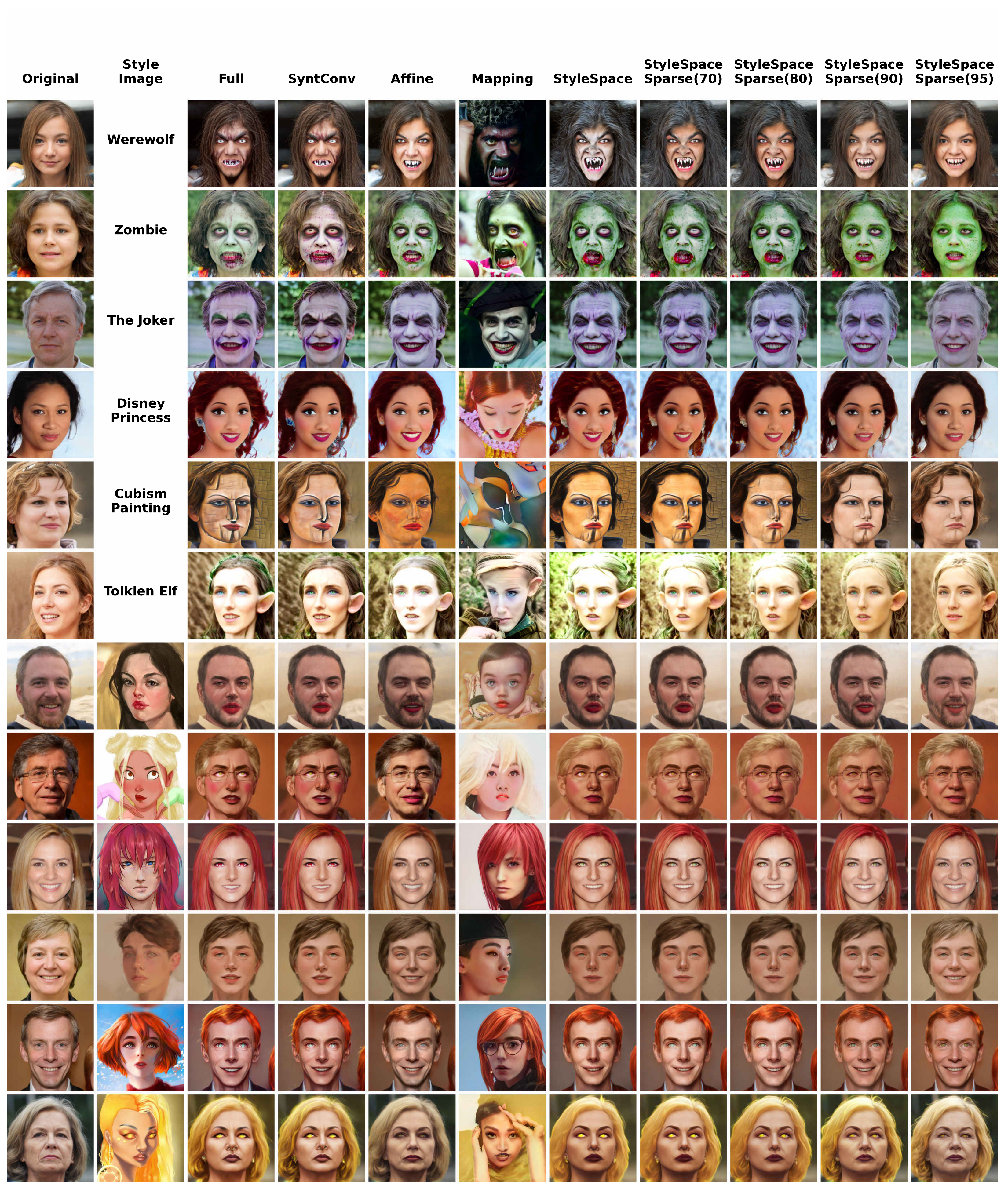}
  \caption{Text-based and one-shot image-based domain adaptation for different parameterizations. Affine, StyleSpace and StyleSparse (70, 80) achieve visual quality on par with Full and SyntConv parameterizations.}
  \label{fig:a2_2}
  \vspace{-0.5cm}
\end{figure*}

\begin{figure*}[!h]
  \centering
  \includegraphics[width=\textwidth]{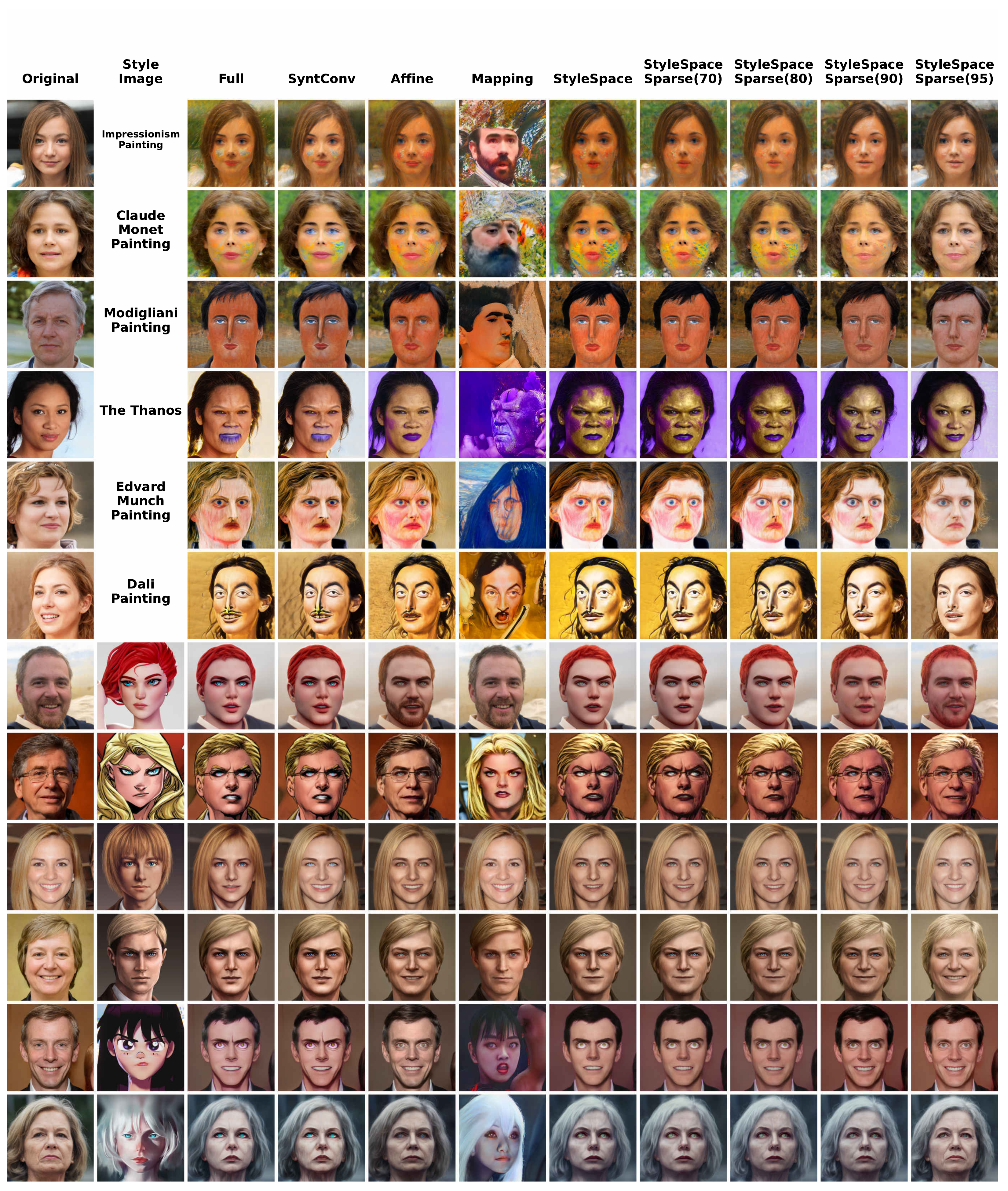}
  \caption{Text-based and one-shot image-based domain adaptation for different parameterizations. Affine, StyleSpace and StyleSparse (70, 80) achieve visual quality on par with Full and SyntConv parameterizations.}
  \label{fig:a2_3}
  \vspace{-0.5cm}
\end{figure*}

We provide more samples for qualitative comparison of different parameterizations. \Cref{fig:a2_1,fig:a2_2,fig:a2_3} are extensions of \Cref{fig:sim_dom_parts} from the \Cref{sec:importance} that show more text-based and image-based domains. We observe that Full, SyntConv and Affine parameterizations demonstrate comparable visual quality and diversity. While the Mapping parameterization obtains poor adaptation quality and collapsed to one image almost for each domain.
We see that these figures agree with the quantitative results and it confirms the conclusions from \Cref{sec:importance}. 

%In this section we provide extra examples for comparison of different parameterizations domain adaptation. Figures [\ref{fig:fig3_like_0}, \ref{fig:fig3_like_1}] show additional examples on Figure~\ref{fig:sim_dom_parts}. Figures [\ref{fig:fig5_like_0}, \ref{fig:fig5_like_1}] show additional examples on Figure~\ref{fig:sim_dom_style}. In addition for every part there is single figure with few examples: 

% \begin{itemize}
%     \item Affine -- \ref{fig:app_affine_sim_examples}
%     \item Full -- \ref{fig:app_full_sim_examples}
%     \item SyntConv -- \ref{fig:app_syntconv_sim_examples}
%     \item Mapping -- \ref{fig:app_mapping_sim_examples}
% \end{itemize}

% \begin{figure*}[!h]
%   \centering
%   \includegraphics[width=\textwidth]{imgs/appendix_sim_affine.pdf}
%   \caption{Text-based and one-shot image-based domain adaptation for Affine parameterization. Uncurated random samples for each domain.}
%   \label{fig:app_affine_sim_examples}
%   \vspace{-0.5cm}
% \end{figure*}

% \begin{figure*}[!h]
%   \centering
%   \includegraphics[width=\textwidth]{imgs/appendix_sim_full.pdf}
%   \caption{Text-based and one-shot image-based domain adaptation for Full parameterization. Uncurated random samples for each domain.}
%   \label{fig:app_full_sim_examples}
%   \vspace{-0.5cm}
% \end{figure*}

% \begin{figure*}[!h]
%   \centering
%   \includegraphics[width=\textwidth]{imgs/appendix_sim_conv.pdf}
%   \caption{Text-based and one-shot image-based domain adaptation for SyntConv parameterization. Uncurated random samples for each domain.}
%   \label{fig:app_syntconv_sim_examples}
%   \vspace{-0.5cm}
% \end{figure*}

% \begin{figure*}[!h]
%   \centering
%   \includegraphics[width=\textwidth]{imgs/appendix_sim_mapping.pdf}
%   \caption{Text-based and one-shot image-based domain adaptation for Mapping parameterization. Uncurated random samples for each domain.}
%   \label{fig:app_mapping_sim_examples}
%   \vspace{-0.5cm}
% \end{figure*}

\FloatBarrier

\subsection{Analysis for Few-Shot Domains}
\label{app:far_dom}
\subsubsection{Additional Few-Shot Domains}
In Appendix, we examine more few-shot domains in addition to Dogs and Cats: MetFaces \cite{karras2020training}, Mega cartoon dataset \cite{pinkney2020resolution}, Ukiyo-e faces \cite{pinkney2020resolution}. We also consider even more dissimilar domains: LSUN Car and Church \cite{yu2015lsun}, Flowers \cite{nilsback2006visual}. 

\subsubsection{Implementation Details}
% About frameworks, networks and source code. 

We use a modified version of the StyleGAN2-ADA Pytorch implementation\footnote{\hyperlink{https://github.com/NVlabs/stylegan2-ada-pytorch}{https://github.com/NVlabs/stylegan2-ada-pytorch}} for training all experiments with few-shot domains. We downscale images from Metfaces, Mega, Ukiyo-e and Flowers datasets to match $512\times512$ resolution and upscale LSUN Car and Church datasets to the same resolution. Also, we use only 10K random subsets from LSUN Car and LSUN Church datasets and 1K random subset from Flowers dataset. We attach all source code that reproduces our experiments
for the setting with moderately similar and dissimilar domains as a part of the supplementary material named ”adaptation\_to few\_shot\_domains”.

\subsubsection{Hyperparameters}
% Like in HyperDomainNet paper \cite{alanov2022hyperdomainnet} in Appendix A.3.1.

We fine-tune all models to a child domain from the FFHQ512 checkpoint. Specifically, we use model config-f and the default hyper-parameters for \texttt{stylegan2} config, except learning rate, from the official Nvidia StyleGAN2-ADA Pytorch implementation. For the Affine+ (and for all models in Table~\ref{table:a.3.Affine+}), StyleSpace and AffineLight+ parameterizations, we increase the generator learning rate from $0.002$ to $0.02$. For other parameterizations, we leave the learning rate unchanged. Also, we use \texttt{bgc} augmentation for all domains and parameterizations. We train models on a single Nvidia V100 GPU for 241K images, but for Mega and Ukiyo-e datasets we use intermediate checkpoints with 40K and 100K images respectfully.

\subsubsection{Quantitative Results}
% Describe how we compute metrics. 
% Provide here tables. 
% Enumerate all domains for which we compute metrics.

For quantitative comparison, we compute FID5k (denoted as FID) using $5000$ generated images and all training images. We also compute FID50k and KID50k (denoted as KID) using $50000$ generated images and all training images. 
These metrics in Tables~[\ref{table:a.3.FID50k},~\ref{table:a.3.KID50k}] reaffirm our observation that the Affine+ parameterization can close the gap with the Full parameterization for most of the domains. For the Car domain, the quality gap remains, although ten times less than for the Affine parameterization.

\begin{table*}[!h]
\centering
\caption{FID50k scores for few-shot domains. Affine+ has results comparable to the Full or SyntConv parameterizations for all domains except Car dataset.}
	\label{table:a.3.FID50k}
    \begin{tabular}{llllllllll}
    \toprule
    Parameter Space &   Size & Metfaces &     Mega &   Ukiyoe &      Dog &      Cat &      Car &   Church &  Flowers \\
    \midrule\midrule
    Full     &  30.3M &   $20.0$ &   $80.4$ &   $19.6$ &   $18.9$ &    $5.9$ &   $20.9$ &   $15.0$ &   $14.8$ \\
    SyntConv &  23.6M &   $21.5$ &   $81.5$ &   $21.2$ &   $20.0$ &    $6.1$ &   $22.6$ &   $15.4$ &   $15.0$ \\
    Affine+  &   5.1M &   $20.2$ &   $77.2$ &   $24.0$ &   $17.1$ &    $5.6$ &   $38.2$ &   $16.2$ &   $15.6$ \\
    Affine   &   4.6M &   $24.3$ &  $107.9$ &   $60.3$ &   $68.0$ &   $27.0$ &  $110.4$ &   $85.9$ &   $40.4$ \\
    Mapping  &   2.1M &   $53.4$ &  $136.2$ &  $155.6$ &  $206.2$ &  $225.6$ &  $237.1$ &  $263.5$ &  $234.3$ \\
    \midrule\bottomrule
    \end{tabular}
%   \vspace{-0.5cm}
\end{table*}
\begin{table*}[!h]
\centering
\caption{KID$\times10^{3}$ metric for few-shot domains. Affine+ has results comparable to the Full or SyntConv parameterizations for all domains except Car dataset.}
	\label{table:a.3.KID50k}
    \begin{tabular}{llllllllll}
    \toprule
    Parameter Space &   Size & Metfaces &     Mega &   Ukiyoe &      Dog &      Cat &      Car &   Church &  Flowers \\
    \midrule\midrule
    Full     &  30.3M &    $2.6$ &  $12.8$ &   $11.5$ &    $9.1$ &    $1.8$ &    $7.8$ &    $7.8$ &    $4.6$ \\
    SyntConv &  23.6M &    $3.4$ &  $13.2$ &   $13.5$ &   $10.7$ &    $1.7$ &    $9.8$ &    $7.2$ &    $4.9$ \\
    Affine+  &   5.1M &    $3.5$ &  $14.1$ &   $16.3$ &    $7.8$ &    $1.2$ &   $16.3$ &    $7.8$ &    $4.2$ \\
    Affine   &   4.6M &    $4.5$ &  $28.8$ &   $52.8$ &   $56.0$ &   $17.7$ &   $88.0$ &   $77.2$ &   $24.2$ \\
    Mapping  &   2.1M &   $22.6$ &  $52.7$ &  $161.9$ &  $169.4$ &  $223.2$ &  $240.3$ &  $298.5$ &  $224.2$ \\
    \midrule\bottomrule
    \end{tabular}
%   \vspace{-0.5cm}
\end{table*}

Exact Affine+ parameterization was selected by analyzing three different domains: Metfaces, Dog and Church. Table \ref{table:a.3.Affine+} demonstrates that Affine+64 has the best performance on these datasets and therefore is selected for the further analysis.

\begin{table*}[!h]
\centering
\caption{Comparison of different block resolutions for the Affine+ parameterization. Affine+64 shows the best performance for the moderately similar and for the distant domains. It is denoted as Affine+ parameterization for the rest of the text.}
	\label{table:a.3.Affine+}
	
    \begin{tabular}{lcccccccc}
    \toprule
    Dataset & Affine+4 & Affine+8 & Affine+16 & Affine+32 & Affine+64 & Affine+128 & Affine+256 & Affine+512 \\
    \midrule\midrule
    Size     &     4.9M &     5.1M &      5.1M &      5.1M &    5.1M &       4.8M &       4.6M &       4.6M \\
    \midrule
    Metfaces &   $22.8$ &   $23.0$ &    $\textbf{22.0}$ &    $22.5$ &  $\textbf{22.1}$ &     $23.0$ &     $23.2$ &     $23.2$ \\
    Dog      &   $24.9$ &   $24.3$ &    $21.6$ &    $19.5$ &  $\textbf{18.6}$ &     $18.9$ &     $21.8$ &     $25.7$ \\
    Church   &   $23.2$ &   $23.4$ &    $22.3$ &    $20.5$ &  $\textbf{18.8}$ &     $20.1$ &     $21.3$ &     $22.7$ \\
    \midrule\bottomrule
    \end{tabular}
  \vspace{-0.5cm}
\end{table*}

\subsubsection{Qualitative Results}
% Provide here samples and describe them. 
% Enumerate all domains for which we provide figures.

\begin{figure}[!h]
  \centering
  \includegraphics[width=0.95\textwidth]{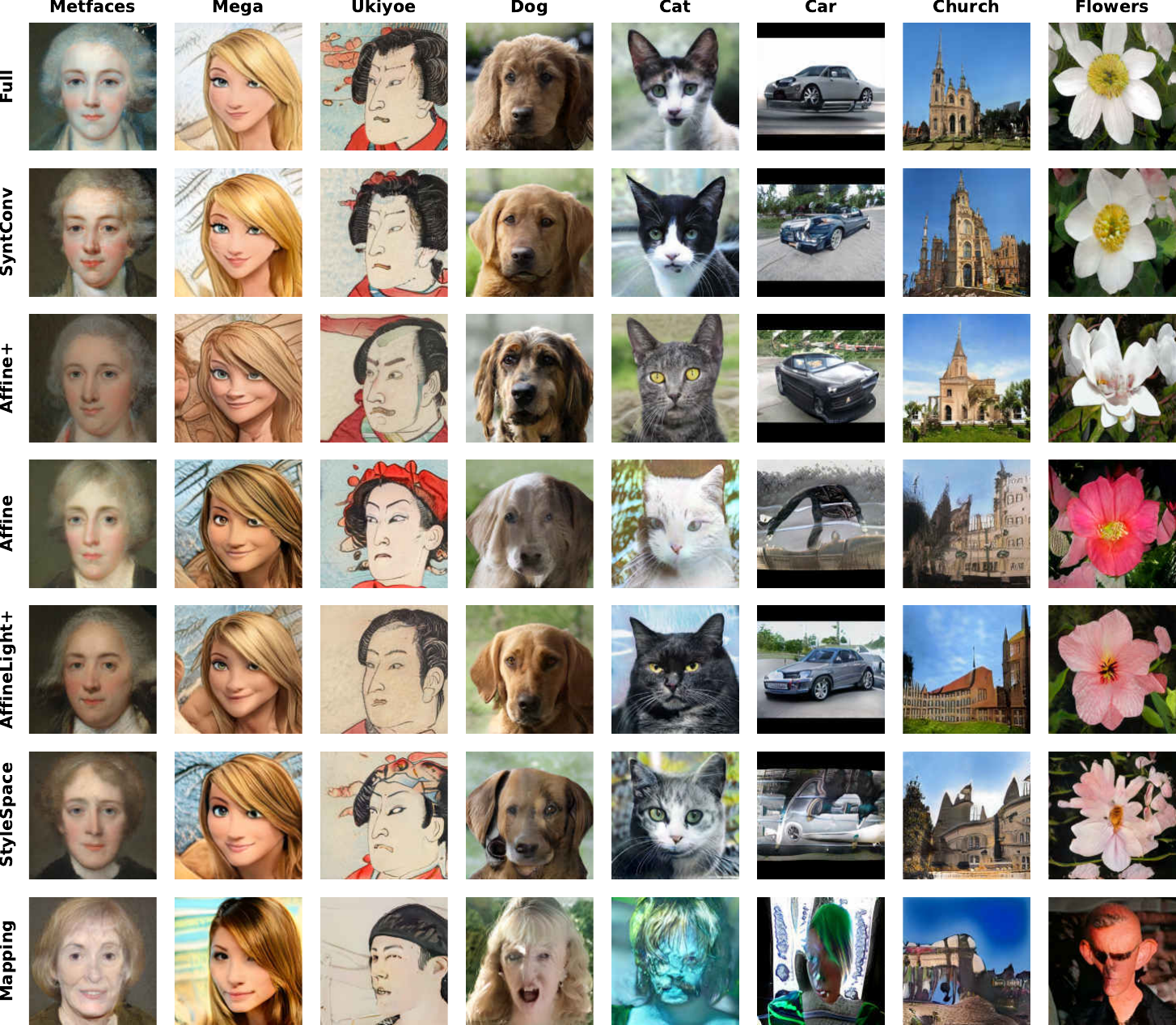}
  \caption{Domain adaptation for few-shot domains. Affine$+$ and AffineLight$+$ parameterizations produce results on par with Full one even for dissimilar domains (Cat, Church).}
  \label{fig:fig_A.3.1}
\end{figure}

\begin{figure}[!h]
  \centering
  \includegraphics[width=0.95\textwidth]{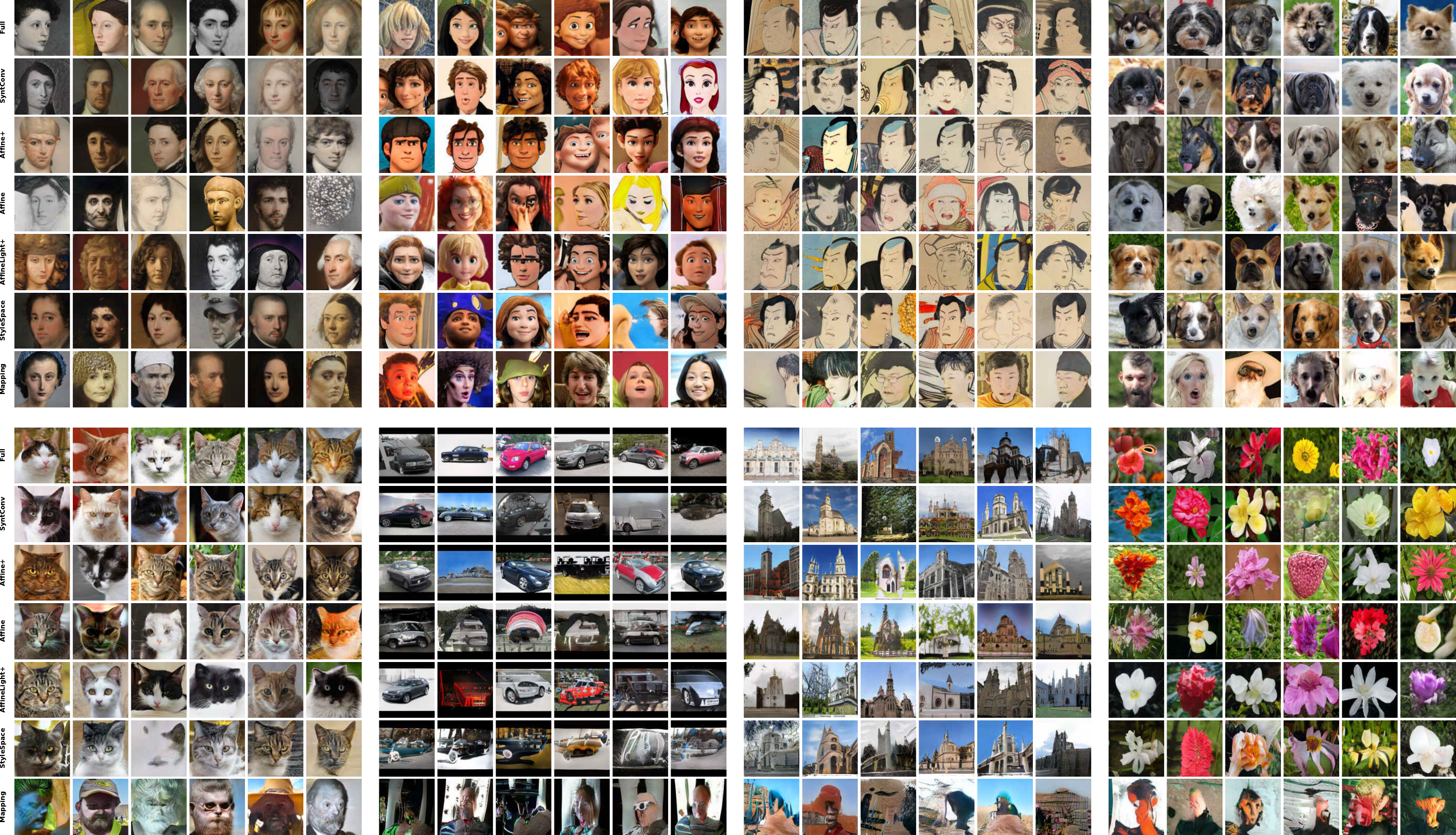}
  \caption{Domain adaptation for few-shot domains.  Uncurated results for each dataset and generator.}
  \label{fig:fig_A.3_4.Samples}
\end{figure}

In this section, we provide generated images for all datasets (Metfaces, Mega, Ukiyo-e, Dog, Cat, Car, Church and Flowers) alongside with uncurated results for parameterizations and datasets.\\*\indent Figures~[\ref{fig:fig_A.3.1},~\ref{fig:fig_A.3_4.Samples}] show that Affine+ parameterization has comparable visual quality and diversity with Full and SyntConv parameterizations for all domains (including dissimilar ones) while having less trainable parameters by the factor of five. At the same time, Affine or Mapping parameterizations fail even on moderately similar domains (Cat, Dog).

\FloatBarrier

\subsection{Results for the Proposed Parameterizations}
\label{app:style_results}
\subsubsection{Analysis for StyleSpace and StyleSpaceSparse}
\textbf{StyleSpaceSparse.} As we note in \Cref{sec:styledirections} we can zero out most of the coordinates of StyleDomain directions without quality degradation. In particular, we apply the standard prunning technique when we leave some percent of the largest absolute values in the StyleDomain vector $\Delta s$ and set the rest to zero: 
\begin{gather}
    \Delta s_i = \begin{cases}
    \Delta s_i, & \text{ if } |\Delta s_i| \geqslant R_q, \\
    0, & \text{ else,}
    \end{cases} \quad \text{where } i = 1, \dots, |\mathcal{S}|, \; R_q = \text{perc}(\{|\Delta s_1|, \dots, |\Delta s_{|\mathcal{S}|}|\}, q). 
\end{gather}
As $\text{perc}(values, q)$ we denote the function that returns the $q$-th percentile of $values$. 

In experiments, we examine different values of prunning rate $q$: $70\%, 80\%, 90\%, 95\%$. We provide quantitative and qualitative performance comparison of these rates in \Cref{table:rate_comparison} and in \Cref{fig:a2_1,fig:a2_2,fig:a2_3}. We observe that $70\%$ almost does not degrade Quality metric and even improve Diversity metric. The strongest rate $95\%$ has the highest Diversity, however, it diminishes Quality significantly. Considering the balance between these two metrics and the parameterization size, we choose $80\%$ rate as the best trade-off. Therefore, in all other experiments we utilize this prunning rate unless other value is explicitly specified.

\begin{table*}[!h]
\centering
\caption{Quality and Diversity metrics \cite{alanov2022hyperdomainnet} for different prunning rates of StyleSpaceSparse parameterization. StyleSpaceSparse with 80\% prunning rate achieves the most balanced performance in terms of Quality and Diversity. }
	\label{table:rate_comparison}
  \begin{tabular}{ lllllll }
    \toprule
\cmidrule(lr){3-4}
Method & Size & Quality & Diversity \\
\midrule\midrule
\textbf{Across 18 text domains} & & & \\
\midrule
Full & 30M & $0.249 \pm 0.024$ & $0.221 \pm 0.046$ \\
StyleSpace & 6048 & $0.265 \pm 0.030$ & $0.182 \pm 0.038$ \\
StyleSpaceSparse (70\%) & 1814 & $0.264 \pm 0.029$ & $0.187 \pm 0.037$ \\
StyleSpaceSparse (80\%) & 1209 & $0.258 \pm 0.028$ & $0.210 \pm 0.041$ \\
StyleSpaceSparse (90\%) & 604 & $0.241 \pm 0.025$ & $0.264 \pm 0.044$ \\
StyleSpaceSparse (95\%) & 302 & $0.208 \pm 0.022$ & $0.318 \pm 0.033$ \\
\midrule
\textbf{Across 21 image domains} & & & \\
\midrule
Full & 30M & $0.719 \pm 0.044$ & $0.214 \pm 0.037$ \\
StyleSpace & 6048 & $0.641 \pm 0.039$ & $0.290 \pm 0.025$ \\
StyleSpaceSparse (70\%) & 1814 & $0.640 \pm 0.043$ & $0.298 \pm 0.032$ \\
StyleSpaceSparse (80\%) & 1209 & $0.631 \pm 0.041$ & $0.307 \pm 0.025$ \\
StyleSpaceSparse (90\%) & 604 & $0.607 \pm 0.044$ & $0.323 \pm 0.023$ \\
StyleSpaceSparse (95\%) & 302 & $0.585 \pm 0.054$ & $0.335 \pm 0.028$ \\
    \midrule
    \bottomrule
  \end{tabular}
\end{table*}

\noindent
\textbf{More results.} We provide more results on capabilities of StyleSpace and StyleSpaceSparse parameterizations for the case of one-shot domains. At first, we evaluate these parameterizations for a large number of different domains in \Cref{table:stylespace_comparison}. We see that for all considered domains StyleSpace and StyleSpaceSparse achieve results comparable to the Full parameterization in terms of both Quality and Diversity metric. It shows that StyleSpace and StyleSpaceSparse allows us to successfully adapt the whole generator to text-based and one-shot domains. Qualitative results in \Cref{fig:a2_1,fig:a2_2,fig:a2_3} confirm our conclusions.

% \begin{figure*}[!h]
%   \centering
%   \includegraphics[width=\textwidth]{imgs/app_fig5_0.pdf}
%   \caption{Text-based and one-shot image-based domain adaptation for different parameterizations. StyleSpace achieves visual quality on par with Full parameterization.}
%   \label{fig:fig5_like_0}
%   \vspace{-0.5cm}
% \end{figure*}

% \begin{figure*}[!h]
%   \centering
%   \includegraphics[width=\textwidth]{imgs/app_fig5_ext_0.pdf}
%   \caption{Text-based and one-shot image-based domain adaptation for different parameterizations. StyleSpace achieves visual quality on par with Full parameterization.}
%   \label{fig:fig5_like_1}
%   \vspace{-0.5cm}
% \end{figure*}

\begin{table*}[!h]
\centering
\caption{Quality and Diversity metrics \cite{alanov2022hyperdomainnet} for text-based and one-shot image-based domain adaptation. StyleSpace and StyleSpaceSparse parameterizations achieve comparable results as Full parameterization.}
	\label{table:stylespace_comparison}
	
    \begin{tabular}{lcccccc}
\toprule
 & \multicolumn{3}{c}{\textbf{Quality}} & \multicolumn{3}{c}{\textbf{Diversity}} \\
 \cmidrule(lr){2-4} \cmidrule(lr){5-7}
 & Full & StyleSpace & StyleSpaceSparse & Full & StyleSpace & StyleSpaceSparse \\
\midrule\midrule
Size & 30.3M & 6.0K & 1.2K & 30.3M & 6.0K & 1.2K \\
\midrule
\textbf{Text Domains} \\
\midrule
Anime Painting & $0.25$ & $0.271$ & $0.269$ & $0.25$ & $0.182$ & $0.190$ \\
Claude Monet & $0.21$ & $0.218$ & $0.208$ & $0.24$ & $0.259$ & $0.277$ \\
Cubism & $0.23$ & $0.237$ & $0.234$ & $0.22$ & $0.221$ & $0.256$ \\
Dali Painting & $0.26$ & $0.272$ & $0.266$ & $0.22$ & $0.183$ & $0.208$ \\
Disney Princess & $0.23$ & $0.272$ & $0.266$ & $0.24$ & $0.183$ & $0.208$ \\
Edvard Munch Painting & $0.25$ & $0.252$ & $0.249$ & $0.2$ & $0.161$ & $0.165$ \\
Fernando Botero Painting & $0.31$ & $0.332$ & $0.322$ & $0.23$ & $0.184$ & $0.212$ \\
Impressionism & $0.27$ & $0.255$ & $0.256$ & $0.25$ & $0.245$ & $0.237$ \\
Modigliani & $0.27$ & $0.285$ & $0.278$ & $0.2$ & $0.189$ & $0.211$ \\
Pixar Render & $0.27$ & $0.266$ & $0.260$ & $0.21$ & $0.190$ & $0.192$ \\
Pop Art & $0.25$ & $0.264$ & $0.260$ & $0.27$ & $0.245$ & $0.260$ \\
Sketch & $0.21$ & $0.196$ & $0.202$ & $0.3$ & $0.267$ & $0.273$ \\
The Joker & $0.25$ & $0.246$ & $0.249$ & $0.17$ & $0.175$ & $0.190$ \\
The Thanos & $0.23$ & $0.269$ & $0.266$ & $0.29$ & $0.232$ & $0.228$ \\
Tolkien Elf & $0.27$ & $0.293$ & $0.286$ & $0.22$ & $0.181$ & $0.210$ \\
Ukiyo-e Painting & $0.22$ & $0.260$ & $0.272$ & $0.22$ & $0.171$ & $0.177$ \\
Werewolf & $0.26$ & $0.252$ & $0.257$ & $0.13$ & $0.125$ & $0.120$ \\
Zombie & $0.24$ & $0.241$ & $0.247$ & $0.12$ & $0.131$ & $0.144$ \\
\midrule
\textbf{Over 18 domains} & $0.25 \pm 0.02$ & $0.26 \pm 0.03$ & $0.26 \pm 0.03$ & $0.22 \pm 0.05$ & $0.20 \pm 0.04$ & $0.21 \pm 0.04$ \\
\midrule
\textbf{Image Domains} \\
\midrule
Adan & $0.762$ & $0.613$ & $0.603$ & $0.168$ & $0.312$ & $0.315$ \\
Arcane & $0.702$ & $0.612$ & $0.596$ & $0.227$ & $0.310$ & $0.307$ \\
Art & $0.766$ & $0.695$ & $0.699$ & $0.202$ & $0.279$ & $0.278$ \\
Anastasia & $0.645$ & $0.560$ & $0.546$ & $0.286$ & $0.349$ & $0.352$ \\
Digital painting & $0.708$ & $0.642$ & $0.661$ & $0.217$ & $0.293$ & $0.282$ \\
Disney & $0.699$ & $0.627$ & $0.617$ & $0.263$ & $0.308$ & $0.304$ \\
Girl & $0.802$ & $0.706$ & $0.711$ & $0.165$ & $0.264$ & $0.262$ \\
Jojo & $0.707$ & $0.628$ & $0.626$ & $0.209$ & $0.275$ & $0.282$ \\
Maria & $0.683$ & $0.646$ & $0.646$ & $0.265$ & $0.300$ & $0.303$ \\
Maryna & $0.673$ & $0.596$ & $0.595$ & $0.239$ & $0.308$ & $0.308$ \\
Mermaid & $0.775$ & $0.689$ & $0.687$ & $0.177$ & $0.281$ & $0.280$ \\
Murasaki & $0.715$ & $0.625$ & $0.621$ & $0.220$ & $0.312$ & $0.316$ \\
Nigelwy & $0.696$ & $0.620$ & $0.620$ & $0.248$ & $0.310$ & $0.310$ \\
Oliver & $0.676$ & $0.610$ & $0.609$ & $0.241$ & $0.318$ & $0.318$ \\
Rain & $0.750$ & $0.713$ & $0.712$ & $0.197$ & $0.258$ & $0.260$ \\
Ricardo & $0.659$ & $0.578$ & $0.576$ & $0.252$ & $0.321$ & $0.319$ \\
Speed paint & $0.759$ & $0.668$ & $0.672$ & $0.245$ & $0.314$ & $0.318$ \\
Supergirl & $0.821$ & $0.688$ & $0.670$ & $0.149$ & $0.261$ & $0.265$ \\
Titan armin & $0.720$ & $0.616$ & $0.599$ & $0.226$ & $0.320$ & $0.327$ \\
Titan erwin & $0.719$ & $0.672$ & $0.659$ & $0.226$ & $0.296$ & $0.303$ \\
Truc & $0.735$ & $0.664$ & $0.665$ & $0.226$ & $0.298$ & $0.296$ \\
\midrule
\textbf{Over 21 domains} & $0.72 \pm 0.04$ & $0.64 \pm 0.04$ & $0.64 \pm 0.04$ & $0.22 \pm 0.03$ & $0.30 \pm 0.02$ & $0.30 \pm 0.02$ \\
    \midrule\bottomrule
    \end{tabular}
  \vspace{-0.5cm}
\end{table*}

% \begin{figure*}[!h]
%   \centering
%   \includegraphics[width=\textwidth]{imgs/appendix_sim_s_delta.pdf}
%   \caption{Text-based and one-shot image-based domain adaptation for StyleSpace parameterization. Uncurated random samples for each domain.}
%   \label{fig:app_s_delta_sim_examples}
%   \vspace{-0.5cm}
% \end{figure*}
\FloatBarrier
\subsubsection{Analysis for Affine$+$ and AffineLight$+$}
\textbf{AffineLight$+$.} As we describe in \Cref{sec:styledirections}, Affine$+$ consists of $f^A, \Delta \theta_1, \theta_2$ where $f^A$ - affine layers, $\Delta \theta_1, \theta_2 \in \mathbb{R}^{512\times512\times1\times1}$ - offsets to weights of two convolutional layers from the synthesis part. To further lightweight this parameterization we reduce its size by applying low-rank decomposition to the weights of affine layers $f^A = \{f_1^A, \dots, f_N^A\}$. 

Each affine layer $f_i^A$ is a linear map with bias: $f_i^A(w) = W_i^Tw + b_i$, where $W_i \in \mathbb{R}^{512\times512}, \; b_i \in \mathbb{R}^{512}$ (these dimensions can differ depending on the layer but we omit these details for brevity). At first, we propose to optimize only offset $\Delta W_i$ to weight $W_i$. Secondly, we apply low-rank decomposition to this offset $\Delta W_i = A_i^TB_i$, where $A_i \in \mathbb{R}^{512\times k}, B_i \in \mathbb{R}^{k\times512}$. In experiments, we examine different low-rank $k$ values and we choose $k = 5$ as the optimal choice regarding its performance and number of trainable parameters. Overall, AffineLight$+$ parameterization has $600K$ parameters when $k = 5$. 

\noindent
\textbf{More results.}
We provide additional results for the Affine+, AffineLight+ and StyleSpace  parameterizations for few-shot domains.

We report FID50k and KID50k scores in Tables~[\ref{table:a.4.FID50k},~\ref{table:a.4.KID50k}]. It can be seen that the results are similar to Table~\ref{table:one-shot}, where the StyleSpace works much worse than the Full parameterization, but adding weight offsets can improve quality for all datasets. So, Affine+ and AffineLight+ parameterizations achieve comparable scores for the moderately similar domains (Metfaces, Mega, Ukiyoe, Dog, Cat) as the Full parameterization while having 60 times less parameters. For dissimilar domains the gap is larger, but it is much less than for StyleSpace parameterization. \Cref{fig:fig_A.3.1,fig:fig_A.3_4.Samples} show that Affine+ and AffineLight+ parameterizations can generate good quality samples even for dissimilar domains. On the other hand, StyleSpace parameterization works well only for the most similar domains, like Metfaces and Ukiyo-e.

% \begin{table*}[!h]
% \centering
% \caption{FID50k scores for domain adaptation in $S$ and $S+$. $S+$ parameterization can be competitive for moderately similar domains}
% 	\label{table:a.4.FID50k}
	
%     \begin{tabular}{llllllllll}
%     \toprule
%     Parameter Space &   Size & Metfaces &     Mega &   Ukiyoe &      Dog &      Cat &      Car &   Church &  Flowers \\
%     \midrule\midrule
%     Full        &  30.3M &   $20.0$ &   $80.4$ &  $19.6$ &  $18.9$ &   $5.9$ &   $20.9$ &  $15.0$ &  $14.8$ \\
%     Affine+     &   5.1M &   $20.2$ &   $77.2$ &  $24.0$ &  $17.1$ &   $5.6$ &   $38.2$ &  $16.2$ &  $15.6$ \\
%     StyleSpace+ &   0.5M &   $23.5$ &   $82.5$ &  $29.1$ &  $28.0$ &   $7.8$ &   $41.7$ &  $25.7$ &  $19.9$ \\
%     StyleSpace  &   9.0K &   $28.6$ &  $102.0$ &  $48.8$ &  $74.2$ &  $20.6$ &  $132.3$ &  $63.9$ &  $36.8$ \\
%     \midrule\bottomrule
%     \end{tabular}
%   \vspace{-0.5cm}
% \end{table*}

\begin{table*}[!h]
\centering
\caption{FID50k scores for domain adaptation using Affine+ and AffineLight+ parameterizations. These parameterizations achieve comparable results with Full parameterization.}
	\label{table:a.4.FID50k}
	
    \begin{tabular}{llllllllll}
    \toprule
    Parameter Space &   Size & Metfaces &     Mega &   Ukiyoe &      Dog &      Cat &      Car &   Church &  Flowers \\
    \midrule\midrule
    Full         &  30.3M &   $20.0$ &   $80.4$ &  $19.6$ &  $18.9$ &   $5.9$ &   $20.9$ &  $15.0$ &  $14.8$ \\
    Affine+      &   5.1M &   $20.2$ &   $77.2$ &  $24.0$ &  $17.1$ &   $5.6$ &   $38.2$ &  $16.2$ &  $15.6$ \\
    AffineLight+ &   0.6M &   $21.3$ &   $69.5$ &  $21.4$ &  $19.0$ &   $7.3$ &   $35.2$ &  $18.6$ &  $15.9$ \\
    StyleSpace   &   9.0K &   $28.6$ &  $102.0$ &  $48.8$ &  $74.2$ &  $20.6$ &  $132.3$ &  $63.9$ &  $36.8$ \\
    \midrule\bottomrule
    \end{tabular}
  \vspace{-0.5cm}
\end{table*}
% \begin{table*}[!h]
% \centering
% \caption{KID$\times 10^{3}$ scores for domain adaptation in $S$ and $S+$. $S+$ parameterization can be competitive for moderately similar domains}
% 	\label{table:a.4.KID50k}
%     \begin{tabular}{llllllllll}
%     \toprule
%     Parameter Space &   Size & Metfaces &     Mega &   Ukiyoe &      Dog &      Cat &      Car &   Church &  Flowers \\
%     \midrule\midrule
%     Full        &  30.3M &    $2.6$ &  $12.8$ &  $11.5$ &   $9.1$ &   $1.8$ &    $7.8$ &   $7.8$ &   $4.6$ \\
%     Affine+     &   5.1M &    $3.5$ &  $14.1$ &  $16.3$ &   $7.8$ &   $1.2$ &   $16.3$ &   $7.8$ &   $4.2$ \\
%     StyleSpace+ &   0.5M &    $5.2$ &  $16.1$ &  $21.3$ &  $16.4$ &   $2.4$ &   $21.8$ &  $14.4$ &   $8.8$ \\
%     StyleSpace  &   9.0K &    $6.5$ &  $24.6$ &  $41.8$ &  $65.7$ &  $11.6$ &  $132.6$ &  $55.6$ &  $21.6$ \\
%     \midrule\bottomrule
%     \end{tabular}
%   \vspace{-0.5cm}
% \end{table*}

\begin{table*}[!h]
\centering
\caption{KID$\times 10^{3}$ scores for domain adaptation using Affine+ and AffineLight+ parameterizations. These parameterizations achieve comparable results with Full parameterization.}
	\label{table:a.4.KID50k}
    \begin{tabular}{llllllllll}
    \toprule
    Parameter Space &   Size & Metfaces &     Mega &   Ukiyoe &      Dog &      Cat &      Car &   Church &  Flowers \\
    \midrule\midrule
    Full         &  30.3M &    $2.6$ &  $12.8$ &  $11.5$ &   $9.1$  &   $1.8$  &   $7.8$   &   $7.8$ &   $4.6$ \\
    Affine+      &   5.1M &    $3.5$ &  $14.1$ &  $16.3$ &   $7.8$  &   $1.2$  &   $16.3$  &   $7.8$ &   $4.2$ \\
    AffineLight+ &   0.6M &    $4.2$ &   $9.5$ &  $12.0$ &   $8.6$  &   $1.8$  &   $13.4$  &   $9.5$ &   $4.4$ \\
    StyleSpace   &   9.0K &    $6.5$ &  $24.6$ &  $41.8$ &   $65.7$ &   $11.6$ &   $132.6$ &  $55.6$ &  $21.6$ \\
    \midrule\bottomrule
    \end{tabular}
  \vspace{-0.5cm}
\end{table*}

% \begin{figure}[!h]
%   \centering
%   \includegraphics[width=0.95\textwidth]{imgs/Figure_A.4.1.pdf}
%   \caption{Domain adaptation for moderately similar and dissimilar domains. StyleSpace+ parameterization closes the gap in quality for all presented domains despite having $60$ times less trainable parameters.}
%   \label{fig:fig_A.4.1}
% \end{figure}

\FloatBarrier

\subsection{Properties of StyleDomain Directions}
\label{app:styledomain}
\subsubsection{Transferability of StyleDomain Directions to Other Aligned StyleGAN Models}

\begin{figure*}[!h]
  \centering
  \includegraphics[width=\textwidth]{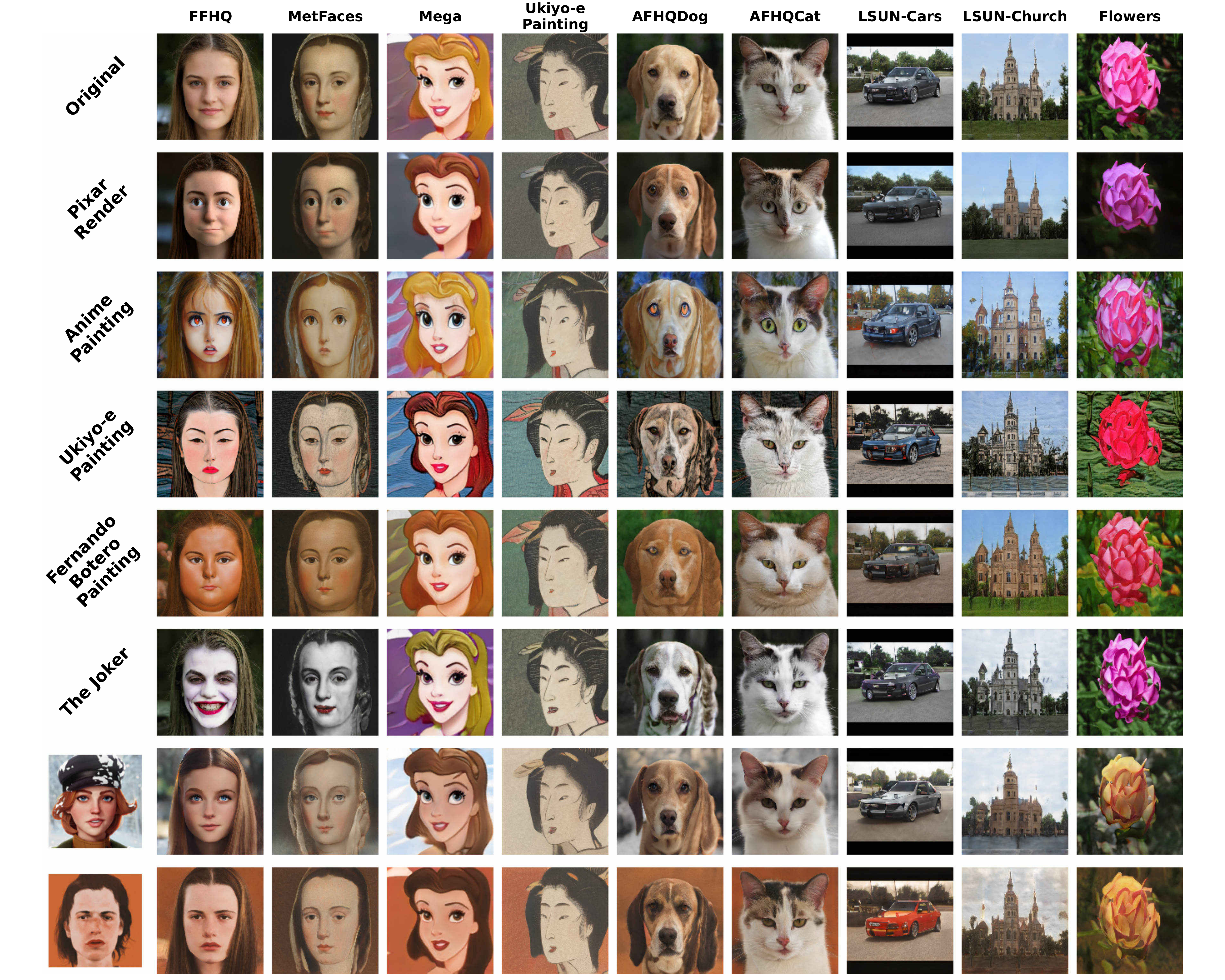}
  \caption{Examples of transferability of StyleDomain directions optimized for similar domains (text-based and one-shot image-based) to aligned generators fine-tuned to other domains (MetFaces, Mega, Dog, etc.). We observe successful transferability for moderately similar domains and reasonable quality for dissimilar ones.}
  \label{fig:app_transfer}
  \vspace{-0.5cm}
\end{figure*}

\begin{figure*}[!h]
  \centering
  \includegraphics[width=\textwidth]{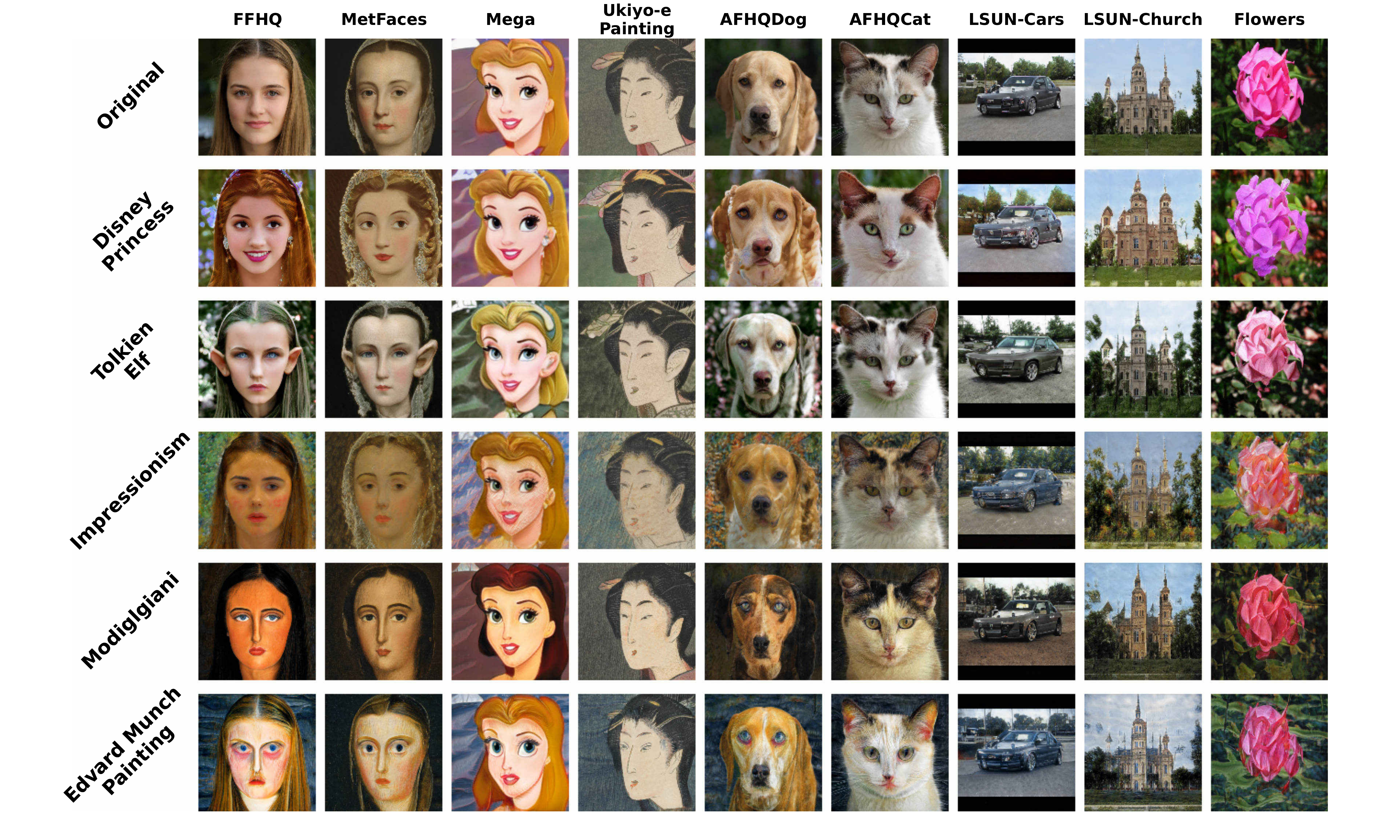}
  \caption{Examples of transferability of StyleDomain directions optimized for similar domains (text-based and one-shot image-based) to aligned generators fine-tuned to other domains (MetFaces, Mega, Dog, etc.). We observe successful transferability for moderately similar domains and reasonable quality for dissimilar ones.}
  \label{fig:app_transfer_ext}
  \vspace{-0.5cm}
\end{figure*}

\textbf{StyleDomain directions of similar domains.} We illustrate the ability of StyleDomain directions optimized for the base generator (e.g. pretrained on FFHQ) to be transferable to other aligned generators fine-tuned to other domains (e.g. MetFaces, Mega, Dogs, etc.). We apply StyleDomain directions that correspond to a wide range of text-based and image-based domains to different aligned generators in \Cref{fig:app_transfer,fig:app_transfer_ext}. We observe that transferability is successful for moderately similar domains and reasonable for dissimilar domains. 

%This section illustrates ability of StyleSpace offsets model for domain adaptation trained on some source domain (e.g. FFHQ) to transfer its properties on other domains (e.g. MetFaces) that are fine-tuned from source domain. Figures [\ref{fig:app_transfer}, \ref{fig:app_transfer_ext}] represents other domains that are not included into main paper.

\FloatBarrier

\subsubsection{Combinations of StyleDomain Directions for Mixed Domain Adaptation}
\begin{figure*}[!h]
  \centering
  \includegraphics[width=\textwidth]{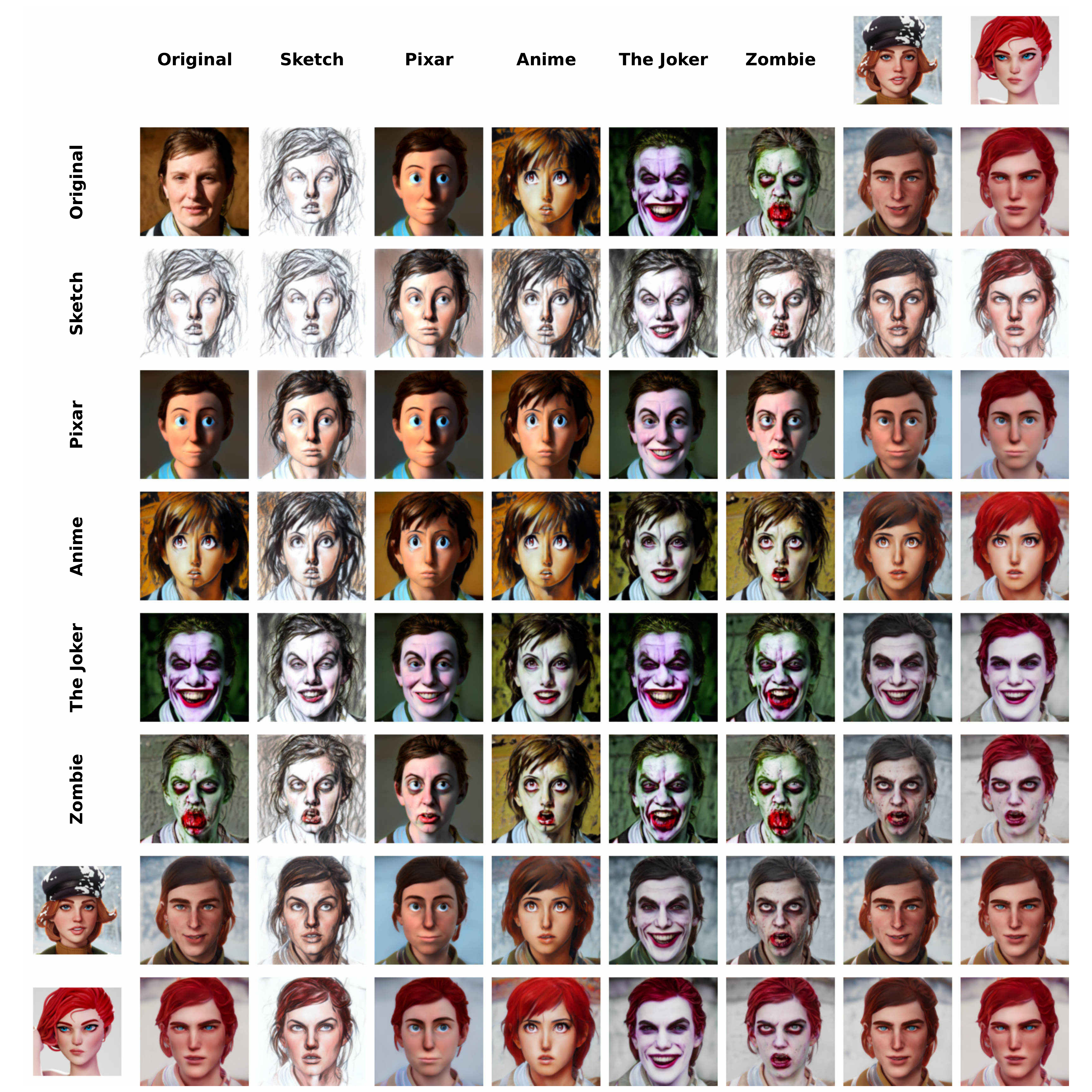}
  \caption{Pairwise combinations of different StyleDomain directions that correspond to text-based or image-based domains. We observe semantically meaningful mixed domains.}
  \label{fig:app_transfer_combinations}
  \vspace{-0.5cm}
\end{figure*}

\begin{figure*}[!h]
  \centering
  \includegraphics[width=\textwidth]{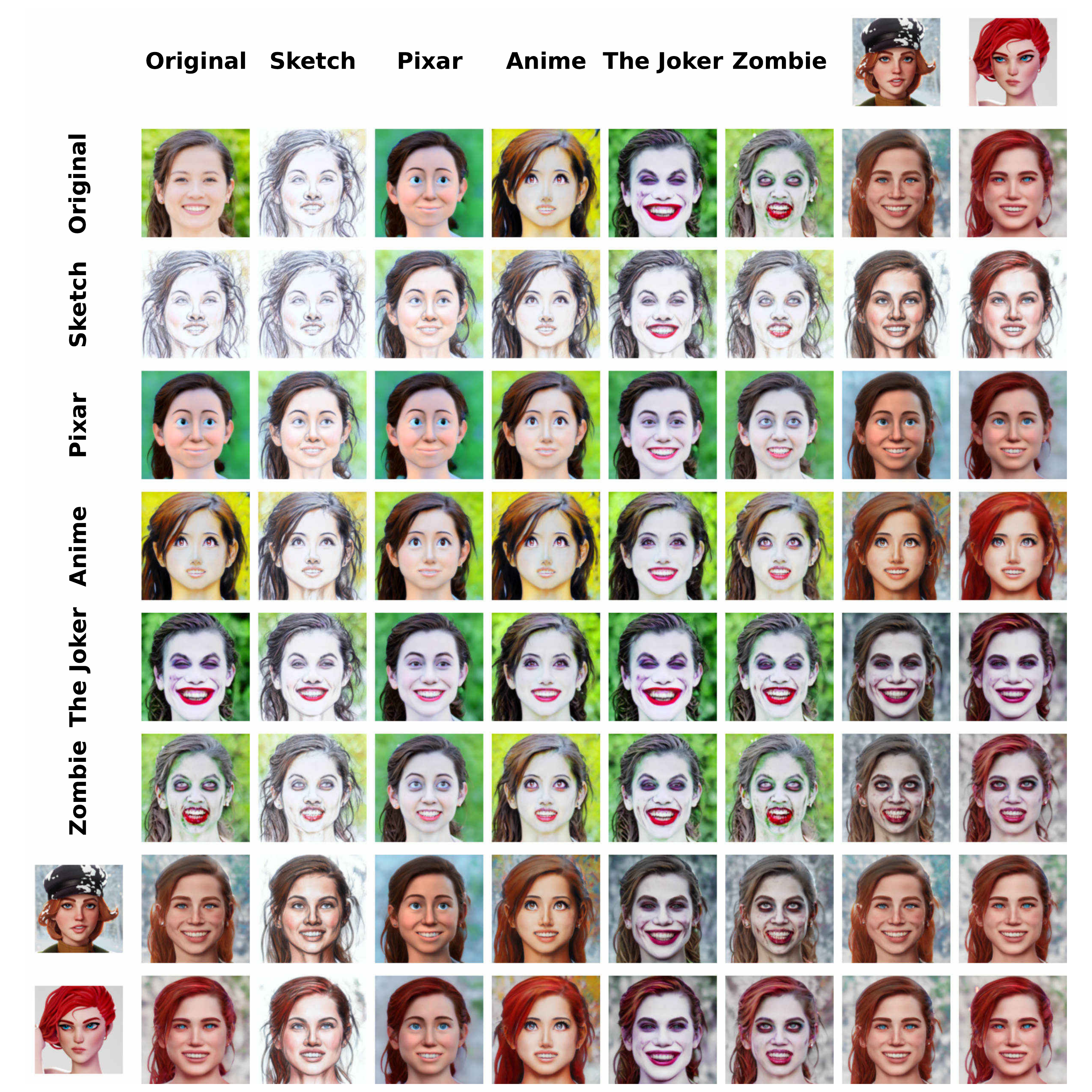}
  \caption{Pairwise combinations of different StyleDomain directions that correspond to text-based or image-based domains. We observe semantically meaningful mixed domains.}
  \label{fig:app_transfer_combinations2}
  \vspace{-0.5cm}
\end{figure*}

\begin{figure*}[!h]
  \centering
  \begin{minipage}{0.5\textwidth}
  \includegraphics[width=\textwidth]{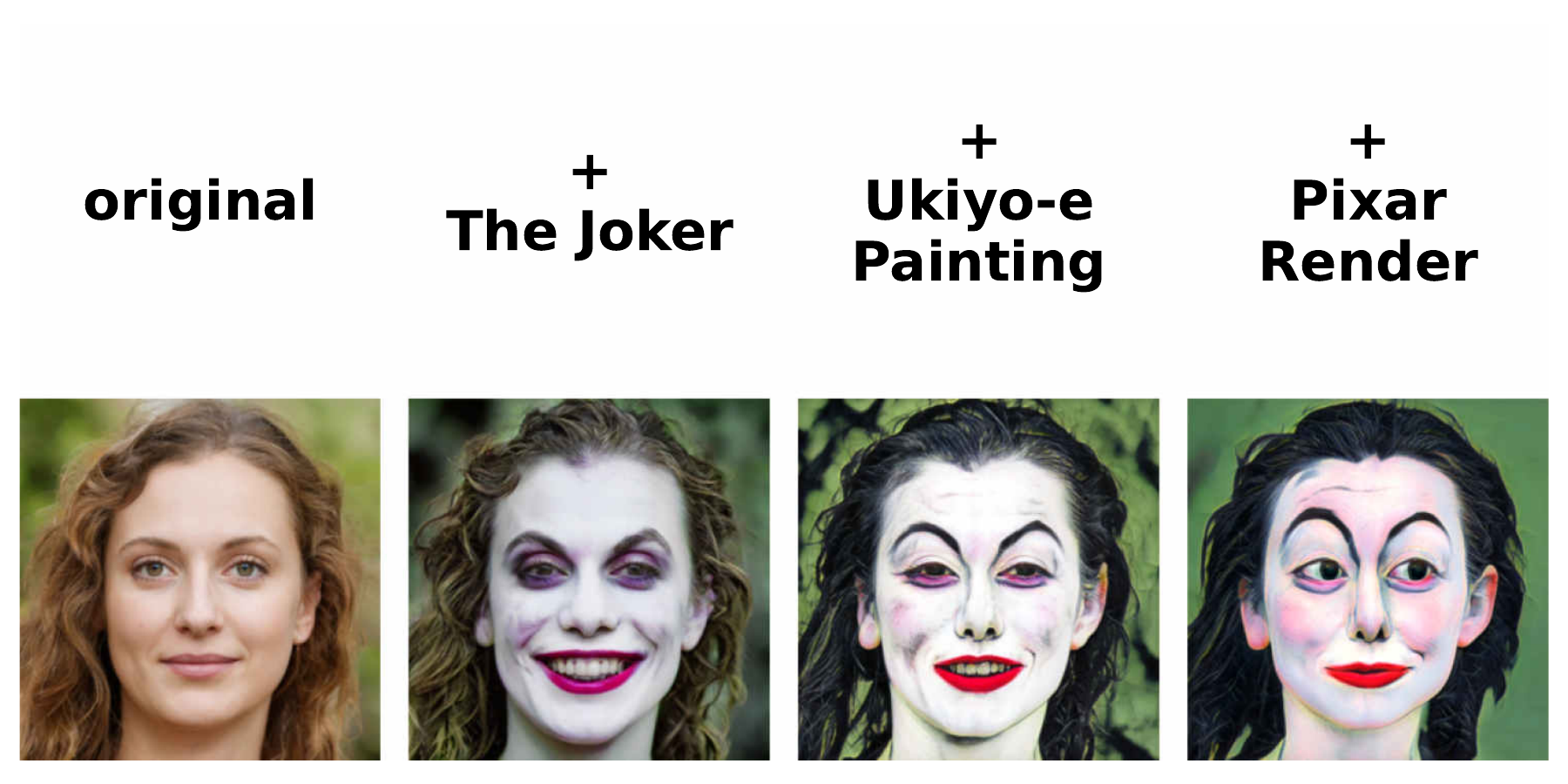}
  \end{minipage}%
  \begin{minipage}{0.5\textwidth}
  \includegraphics[width=\textwidth]{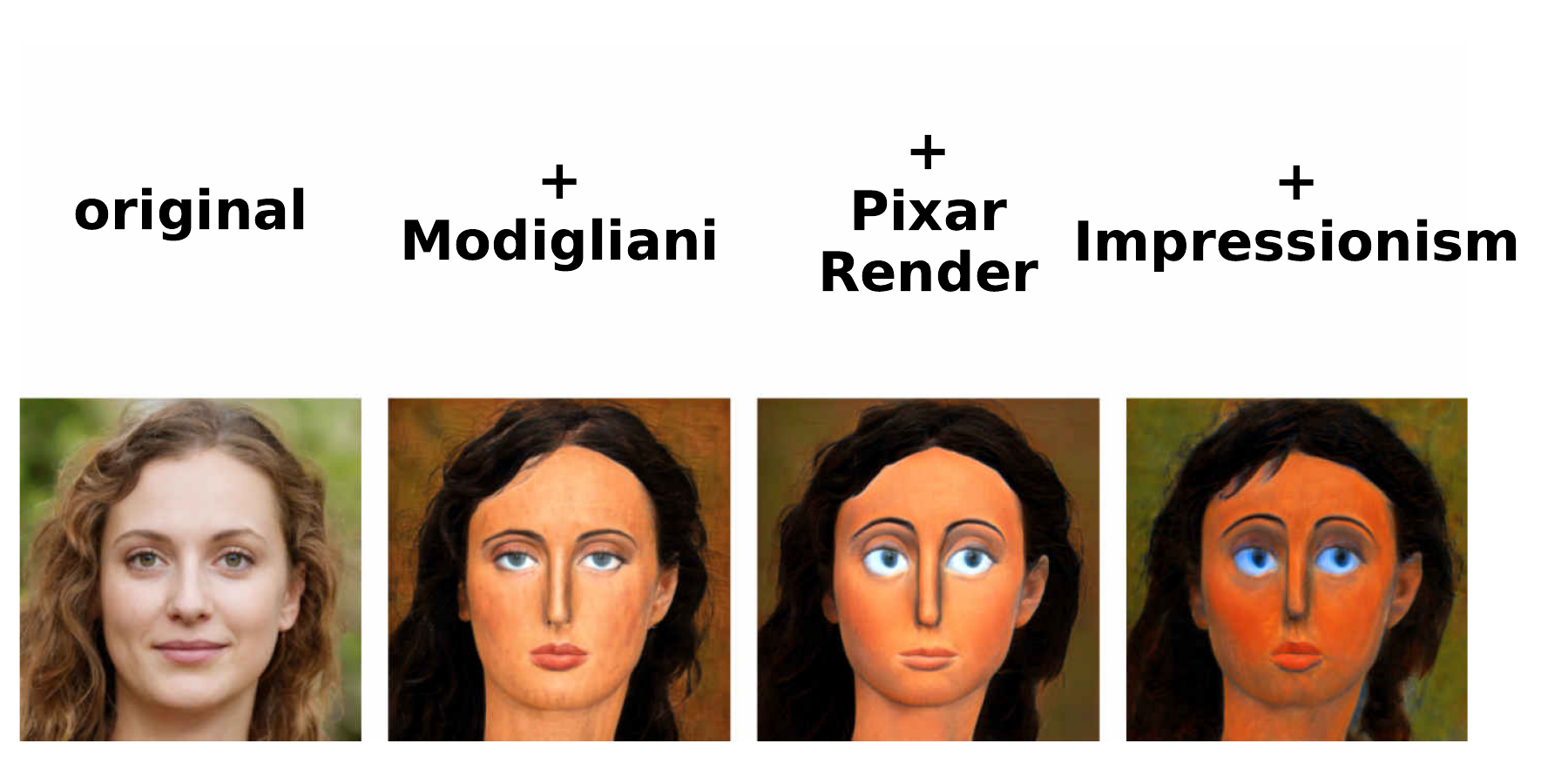}
  \end{minipage}
  \begin{minipage}{0.5\textwidth}
  \includegraphics[width=\textwidth]{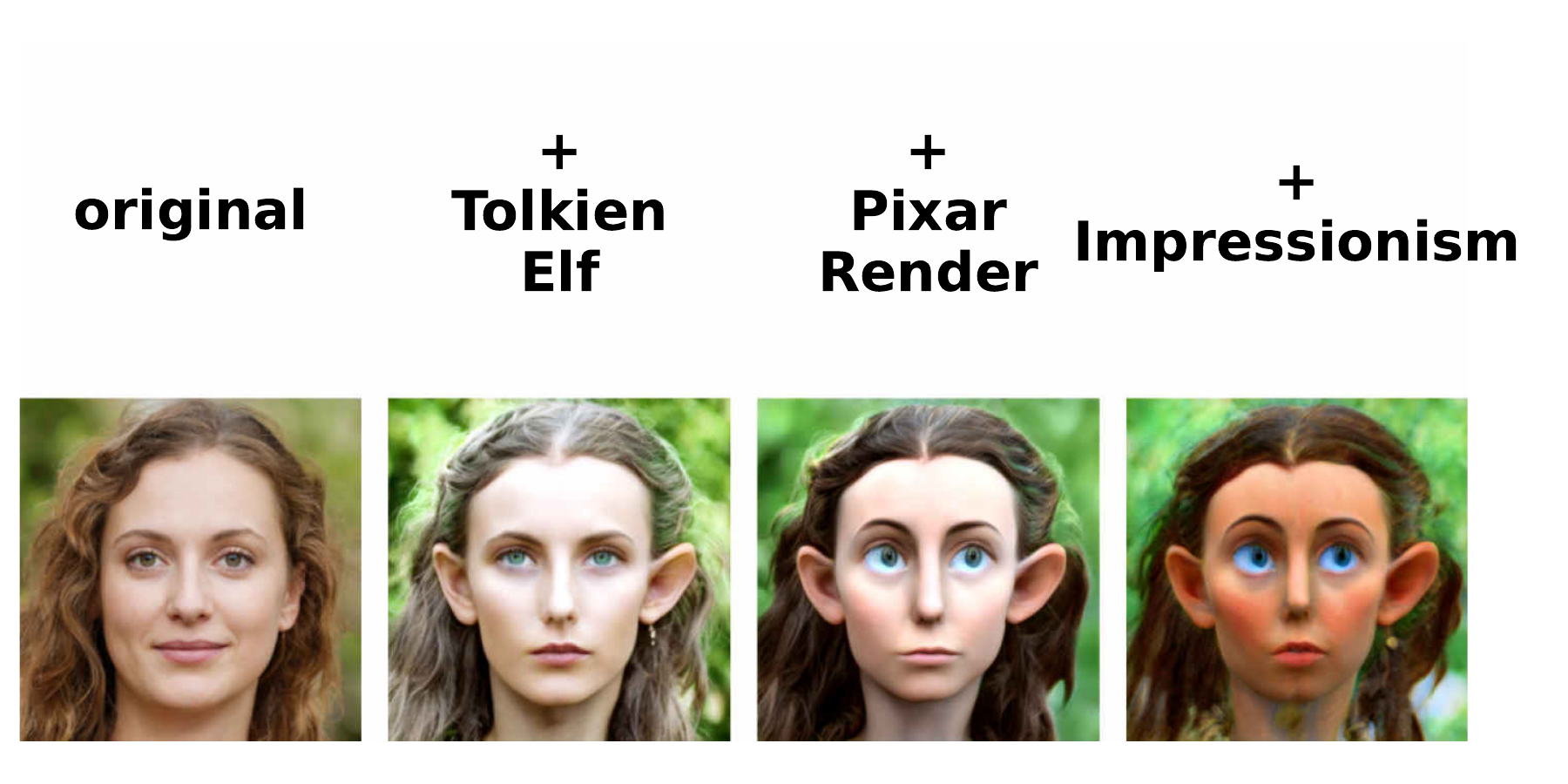}
  \end{minipage}%
  \begin{minipage}{0.5\textwidth}
  \includegraphics[width=\textwidth]{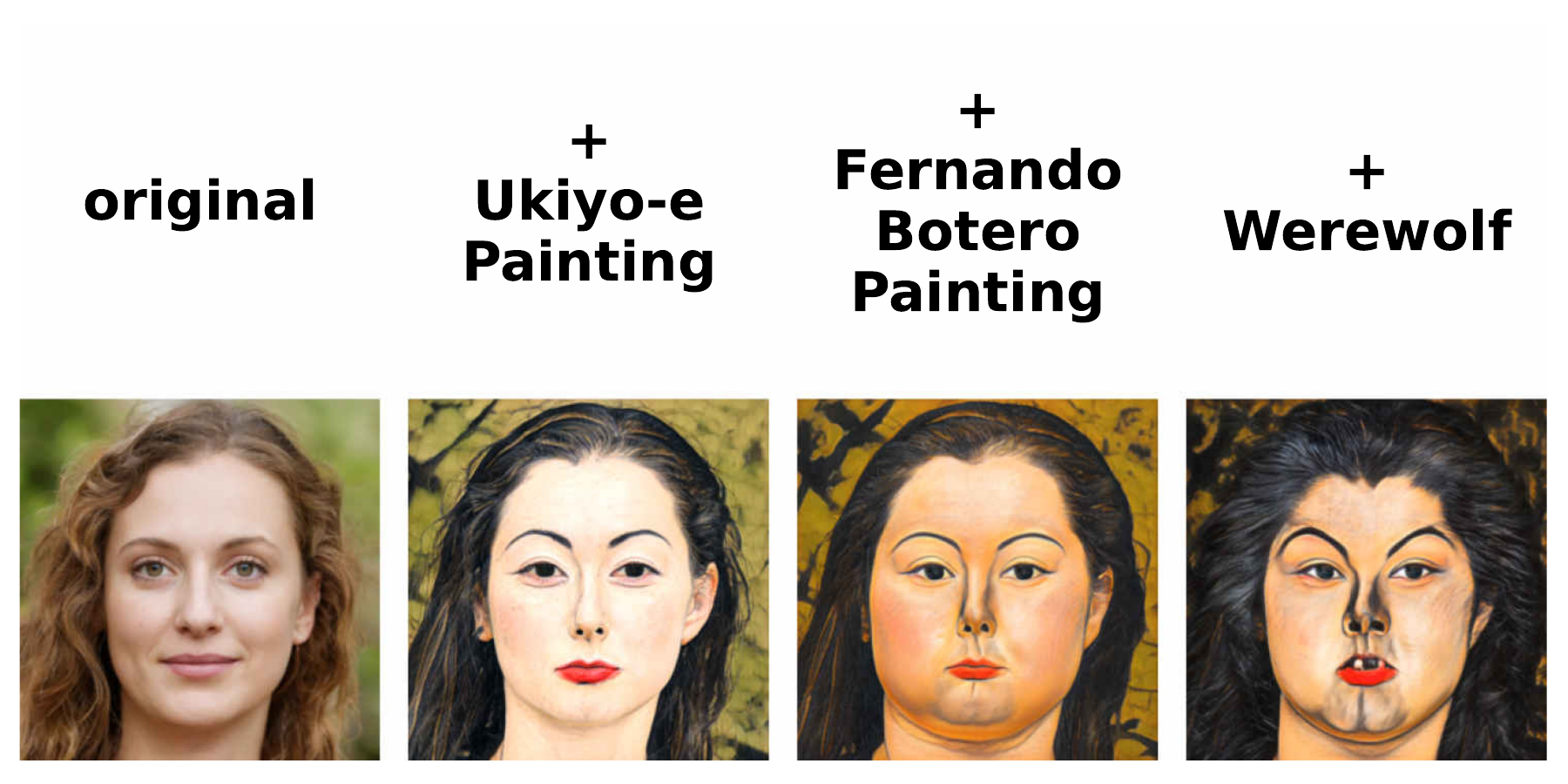}
  \end{minipage}
  \begin{minipage}{0.5\textwidth}
  \includegraphics[width=\textwidth]{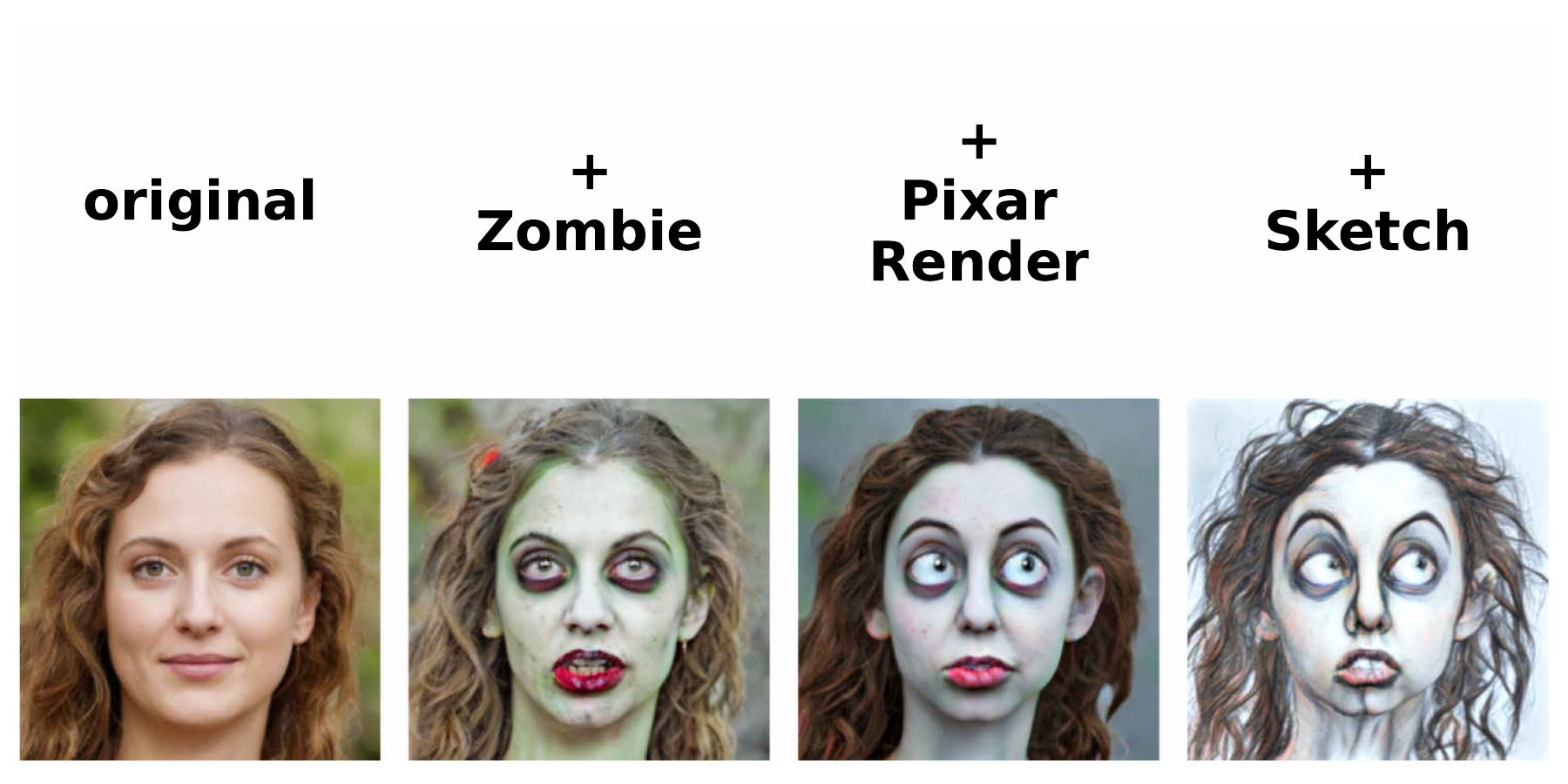}
  \end{minipage}%
  \begin{minipage}{0.5\textwidth}
  \includegraphics[width=\textwidth]{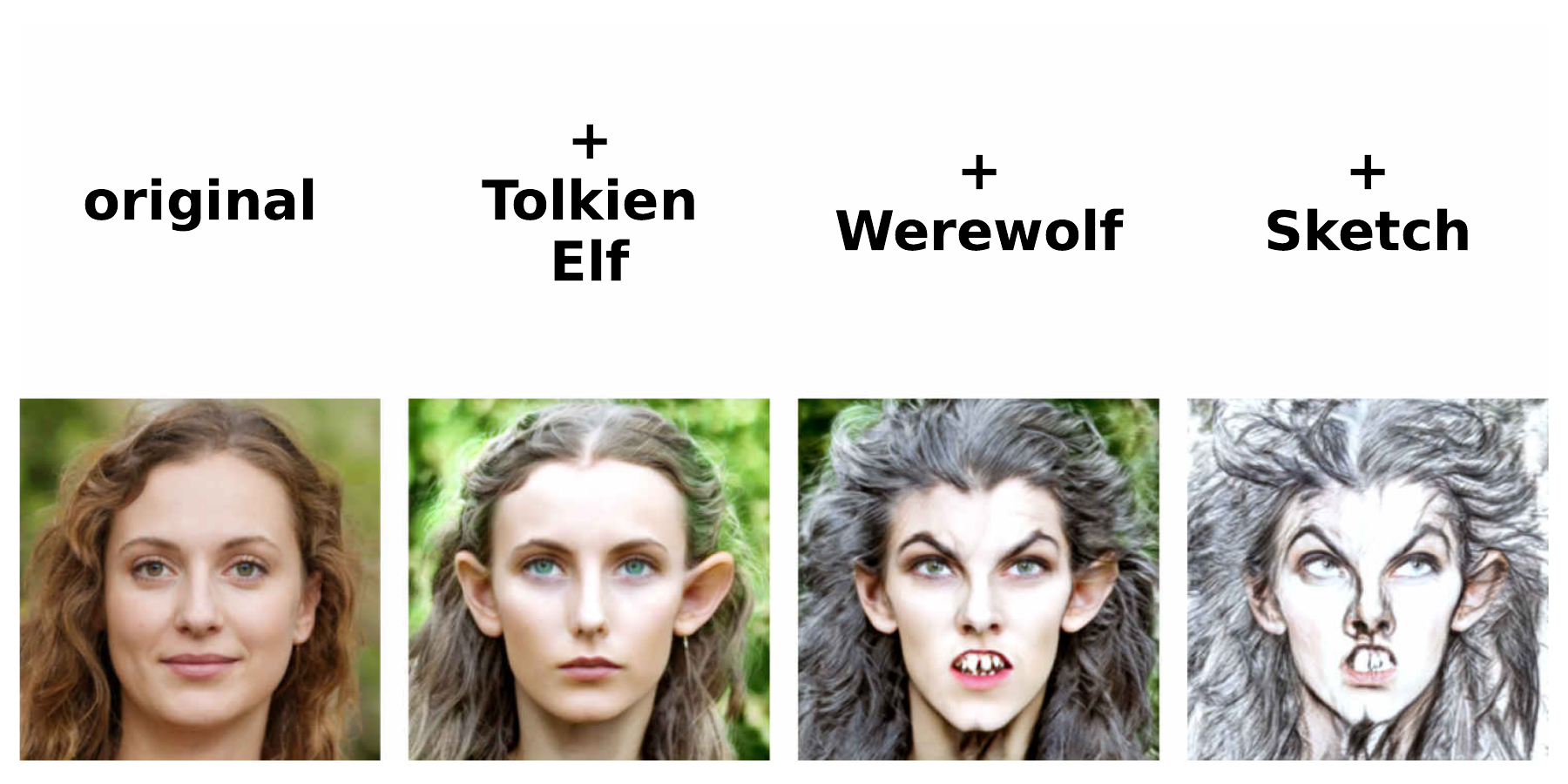}
  \end{minipage}
  \begin{minipage}{0.5\textwidth}
  \includegraphics[width=\textwidth]{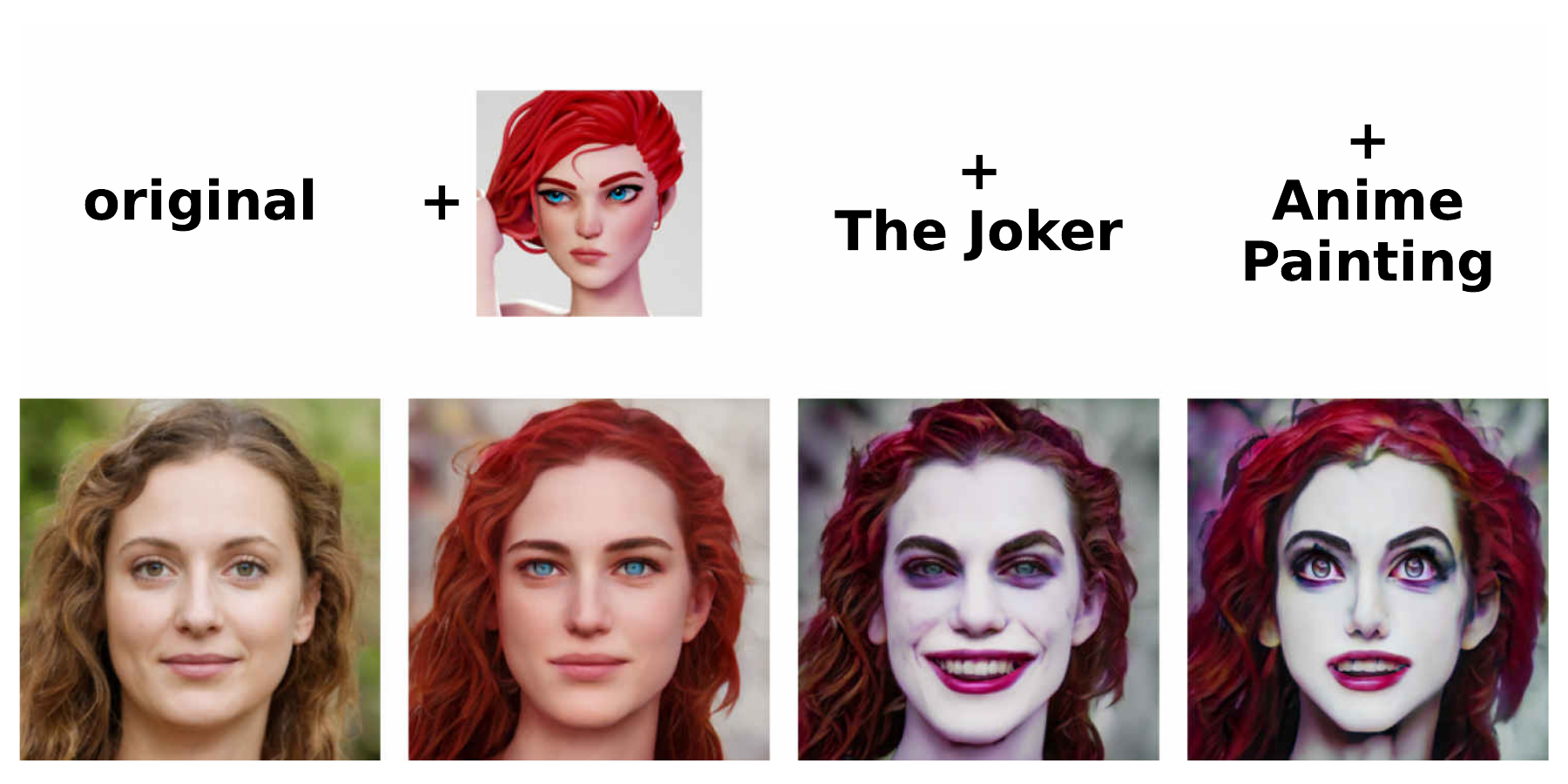}
  \end{minipage}%
  \begin{minipage}{0.5\textwidth}
  \includegraphics[width=\textwidth]{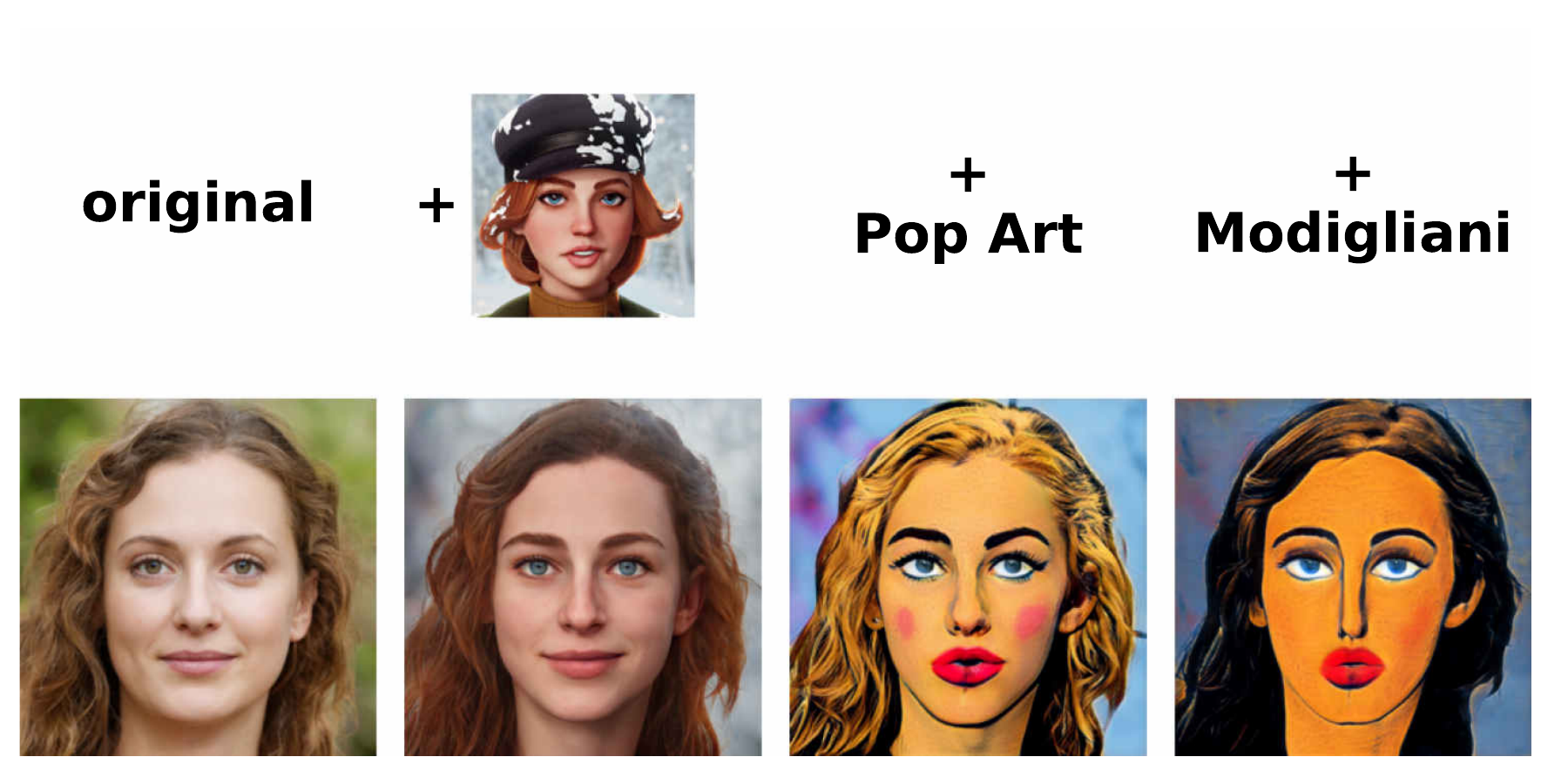}
  \end{minipage}
  \caption{Triple combinations of different StyleDomain directions that correspond to text-based or image-based domains. We observe semantically meaningful mixed domains.}
  \label{fig:triple_combinations}
\end{figure*}

We present more examples of the StyleDomain directions property to be combined with each other to obtain new mixed domains. In \Cref{fig:app_transfer_combinations,fig:app_transfer_combinations2} we consider all pairwise combinations between different text-based and image-based domains. We observe that new obtained mixed domains have meaningful semantic blending of the corresponding two domains. We also consider combinations of three different domains and provide several examples in \Cref{fig:triple_combinations}. We see that these mixed domains also have a recognisable semantic sense.

%This section illustrates mixability of StyleSpace offsets model between different domains. Figures [\ref{fig:app_transfer_combinations}, \ref{fig:app_transfer_combinations2}] represents other mixtures that are not included into main paper.

\FloatBarrier

\subsubsection{Semantic Editing for Similar Domains}
We demonstrate how StyleDomain directions can be combined with latent transformations for semantic editing. 

We utilize the StyleSpace \cite{wu2021stylespace} editing method for "Black hair", "Smile" and "Lipstick" attributes. InterFaceGan \cite{shen2020interpreting} editing method is used for "Yaw", "Rejuvenation", "Aging", "Gender".

\begin{figure*}[!h]
  \centering
  \includegraphics[width=0.9\textwidth]{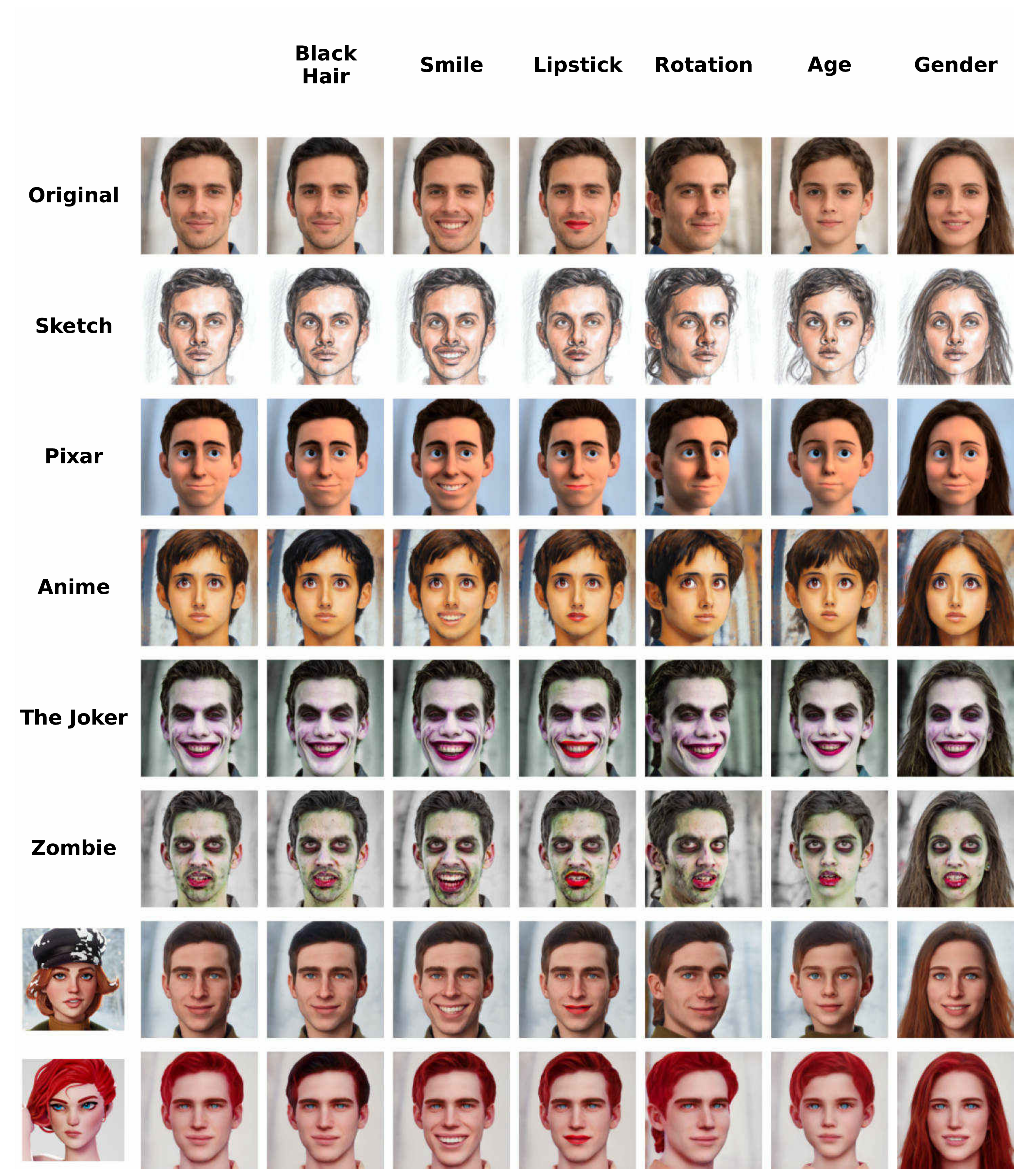}
  \caption{Combination of StyleDomain directions with semantic editing. We observe that they are successfully combined.}
  \label{fig:app_editing_0}
  \vspace{-0.5cm}
\end{figure*}

\begin{figure*}[!h]
  \centering
  \includegraphics[width=\textwidth]{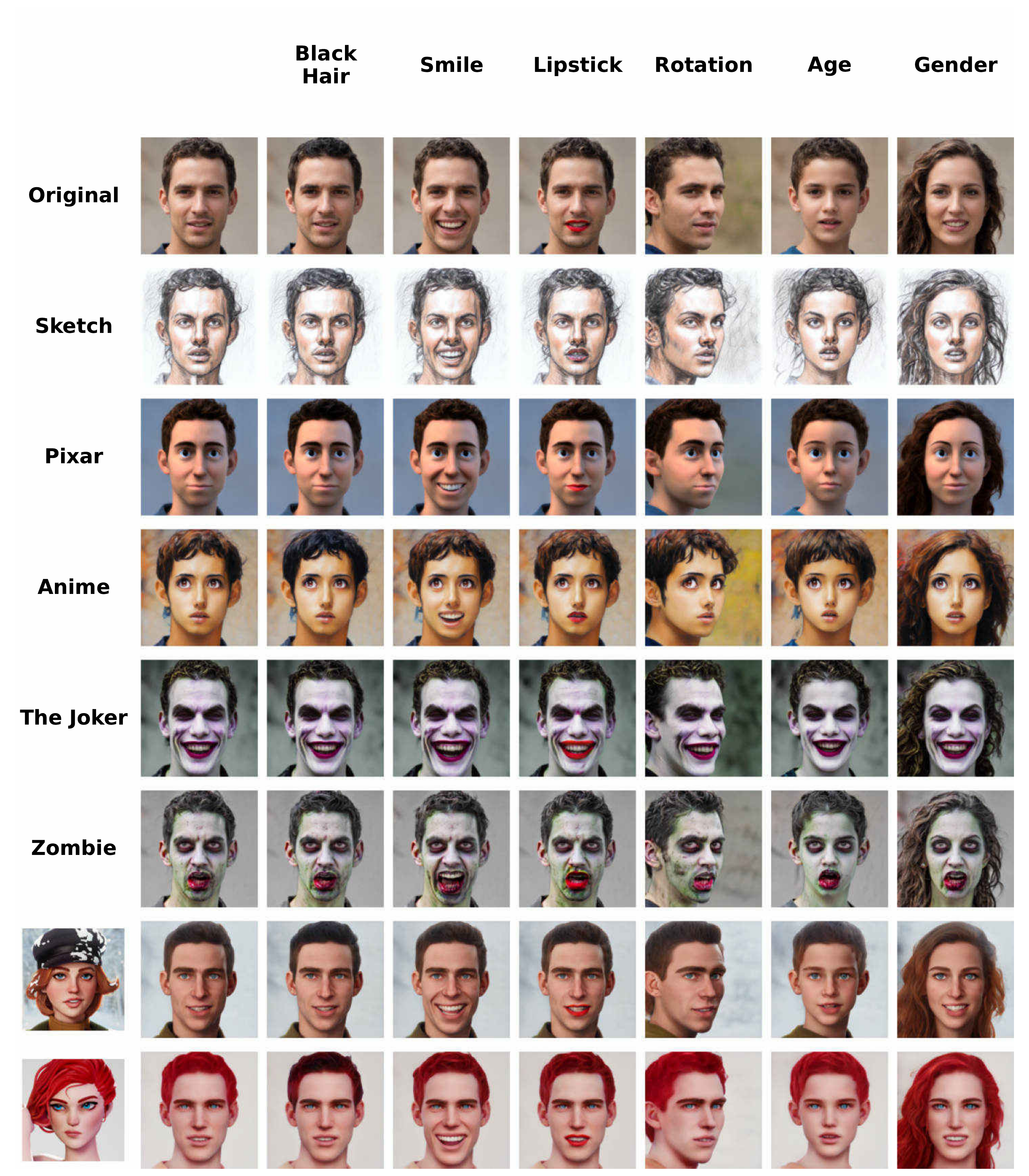}
  \caption{Combination of StyleDomain directions with semantic editing. We observe that they are successfully combined.}
  \label{fig:app_editing_1}
  \vspace{-0.5cm}
\end{figure*}

\Cref{fig:app_editing_0,fig:app_editing_1} show results of combination of StyleDomain directions for different text-based and image-based domains and editing of mentioned attributes. We see that domain and modification directions are orthogonal, so they are combined successfully.  

%This section illustrates combination of StyleSpace offsets inference and editing via latent transformation. Figures [\ref{fig:app_editing_0}, \ref{fig:app_editing_1}, \ref{fig:app_editing_2}] represents editing cases that are not included into main paper. For attributes "Black hair", "Smile" and "Lipstick" StyleSpace \cite{wu2021stylespace} editing method is used. For "Rotation" and "Age" InterFaceGan \cite{shen2020interpreting} method is used. For "Gender" StyleFlow \cite{abdal2021styleflow} is used.

\FloatBarrier

\subsubsection{Semantic Editing for Moderately Similar and Dissimilar Domains}
We analyze the ability to be semantically edited for different parameterizations in the case of moderately similar and dissimilar domains.

%In this section, we analyze latent transformations of fine-tuned models. 

We use the StyleSpace \cite{wu2021stylespace} editing method for "Blond Hair", "Black Hair", "Smile", "Lipstick", "Short Hair", "Wavy Hair" and "Gaze" attributes. The StyleFlow \cite{abdal2021styleflow} editing method is used for "Yaw", "Rejuvenation", "Aging", "Gender". Since the models for modifications imply using of images in $1024 \times 1024$ resolution, we train separate models in this resolution from FFHQ1024 checkpoint. We use exactly the same setup as described in Appendix~\ref{app:far_dom} for training these models.

\begin{figure*}[!h]
  \centering
  \includegraphics[width=0.9\textwidth]{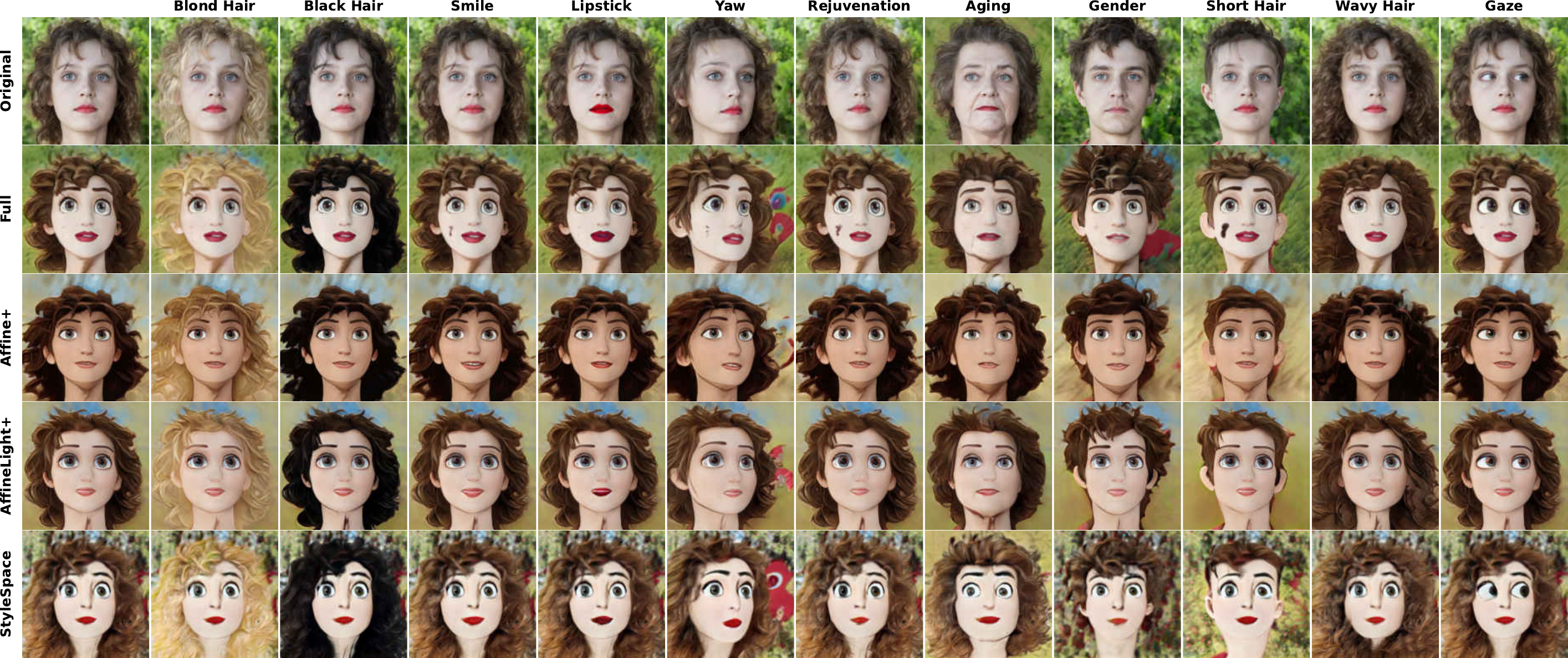}
  \caption{Examples of different parameterizations mixed with semantic editing for Mega. Top row represents attribute names. We can control style (color of hair or lips) as well as the shape (age, gender, hair length and yaw) for all parameterizations.}
  \label{fig:app_editing_Mega}
  \vspace{-0.5cm}
\end{figure*}

\begin{figure*}[!h]
  \centering
  \includegraphics[width=0.9\textwidth]{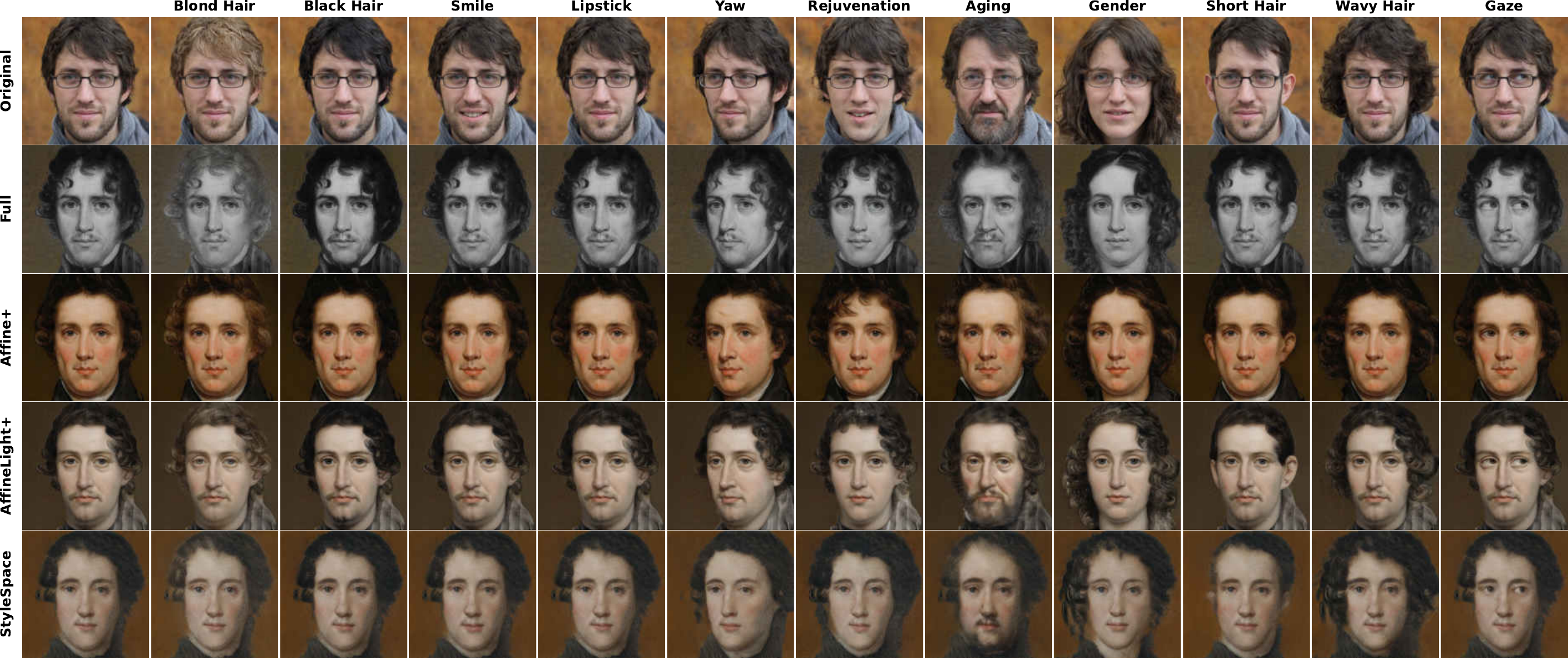}
  \caption{Examples of different parameterizations mixed with semantic editing for Metfaces. Top row represents attribute names.}
  \label{fig:app_editing_Metfaces}
  \vspace{-0.5cm}
\end{figure*}

\begin{figure*}[!h]
  \centering
  \includegraphics[width=0.96\textwidth]{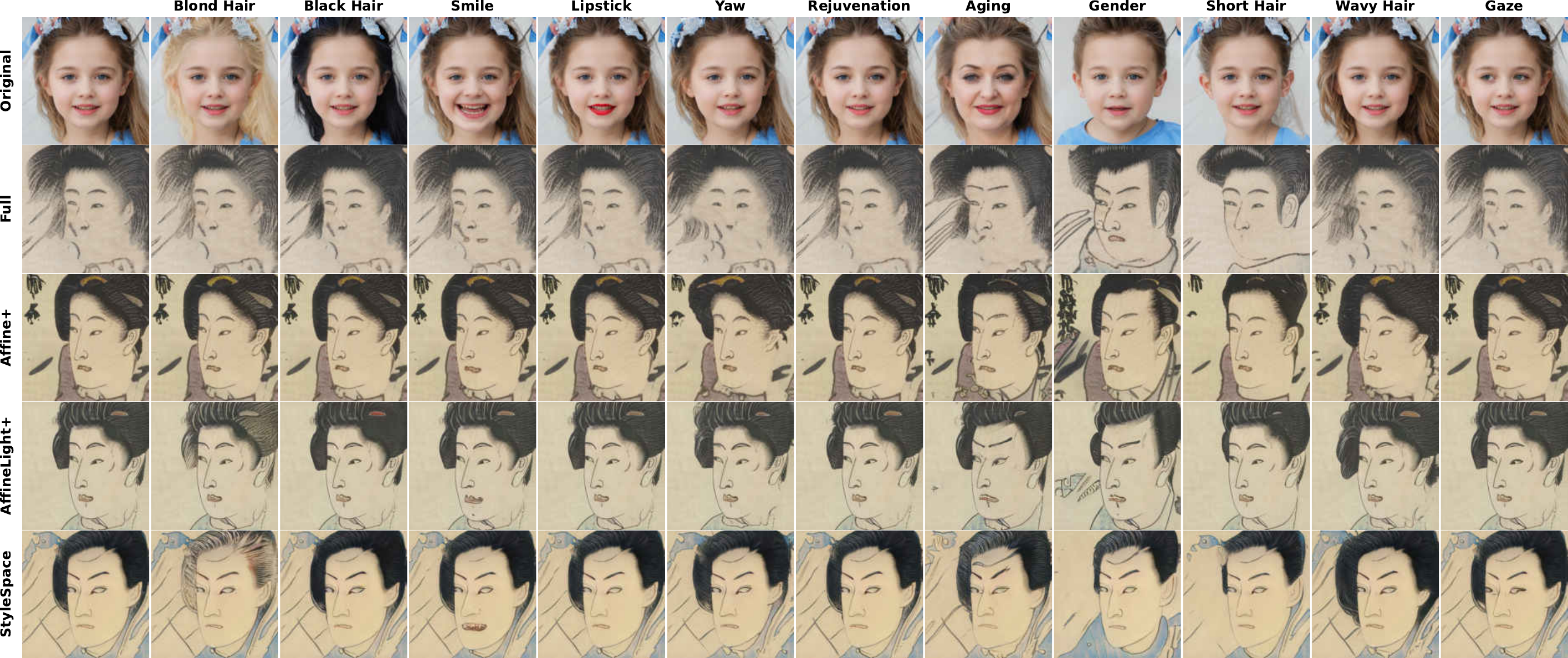}
  \caption{Examples of different parameterizations mixed with semantic editing for Ukiyo-e. Top row represents attribute names. Most of the transformations are inoperable regardless of the exact choice of parameterization.}
  \label{fig:app_editing_Ukiyoe}
  \vspace{-0.2cm}
\end{figure*}

\begin{figure}[!h]
{
    \begin{tabular}{cc}
        \includegraphics[width=0.48\textwidth]{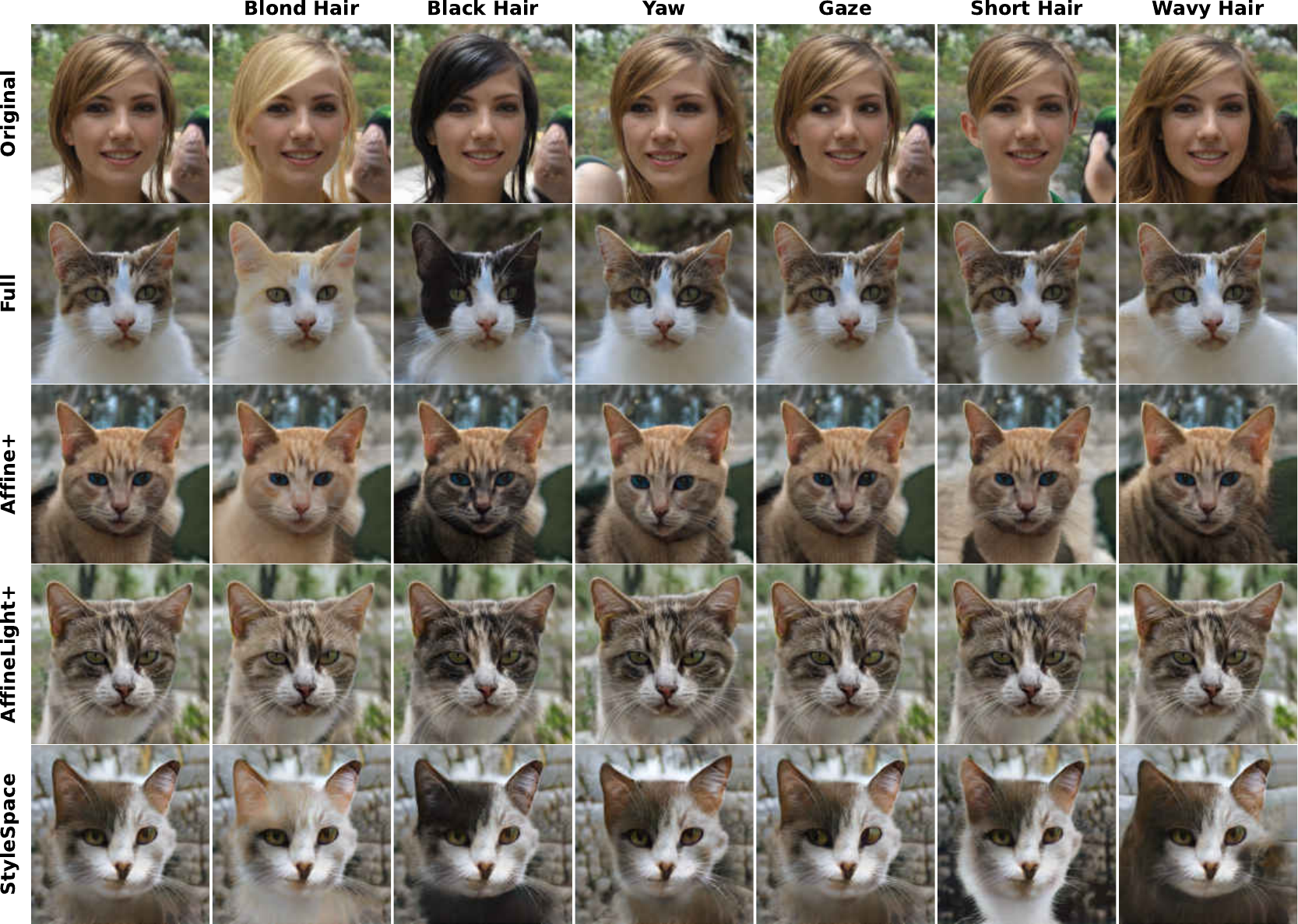} &
        \includegraphics[width=0.48\textwidth]{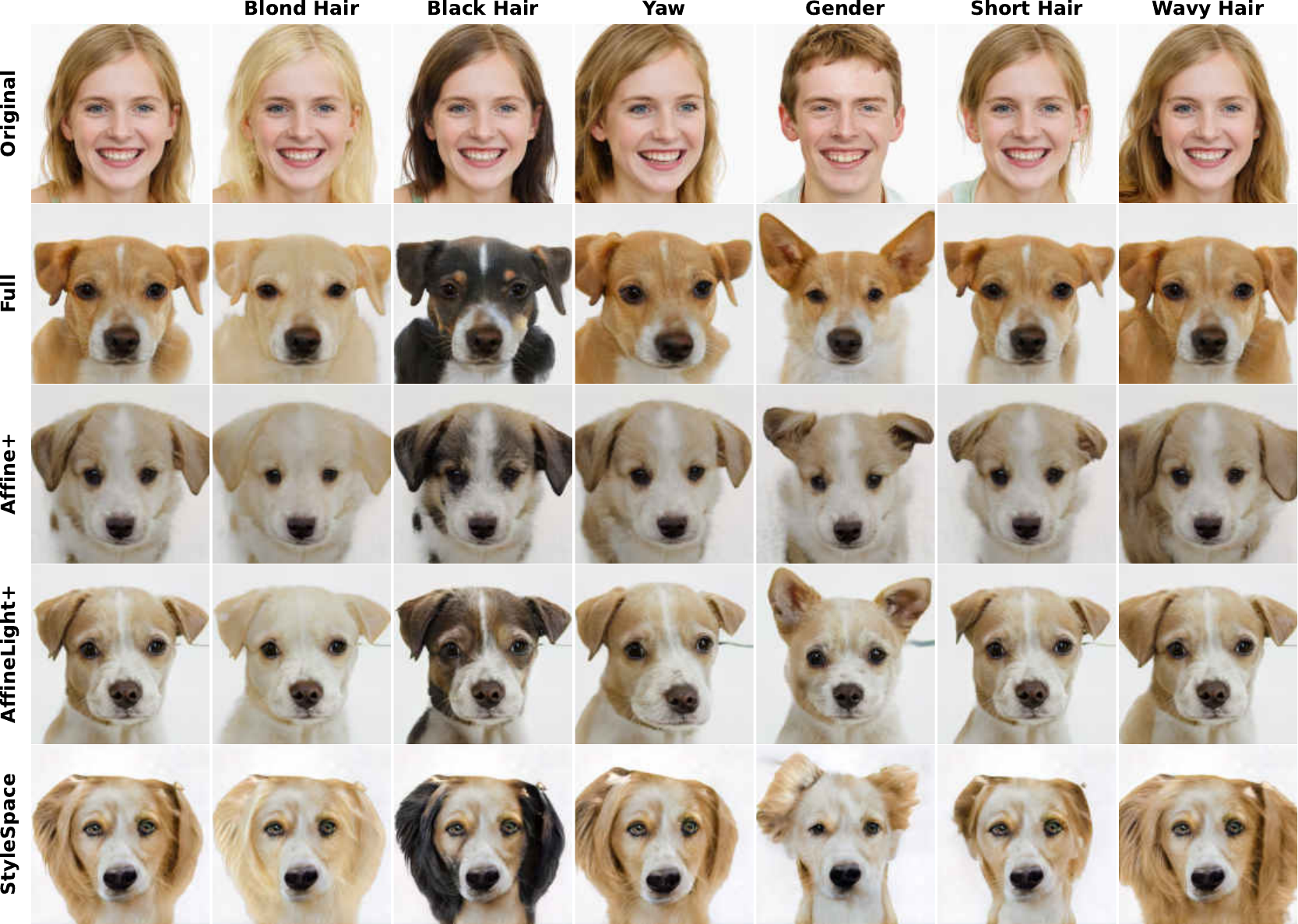} \\
    \end{tabular}
}
\centering
\caption{Examples of different parameterizations mixed with semantic editing for Cat and Dog. Top row represents attribute names. Only working latent transformations are left.}
\label{fig:app_editing_Cat_Dog}
\end{figure}

As can be seen from Figure~\ref{fig:app_editing_Mega}, we can utilize existing latent transformations for models fine-tuned with different parameterizations. Notably, we can successfully apply simple style modifications (change of hair color) as well as complex shape transformations (change in age and gender). For more distant domains in Figures~[\ref{fig:app_editing_Metfaces},~\ref{fig:app_editing_Ukiyoe},~\ref{fig:app_editing_Cat_Dog}] some controls become much less pronounced ("Smile") or even inoperable ("Lipstick").

We observe a negligible difference in the quality of modifications between different parameterizations. So, we can conclude that the ability to perform latent transformations mostly depends on the target dataset rather than the choice of parameterization.

% https://github.com/betterze/StyleSpace/blob/main/StyleSpace_single.ipynb

\FloatBarrier

\subsection{One-Shot Domain Adaptation}
\label{app:one_shot_adaptation}
\subsubsection{Setup of Baselines}
For image-based one-shot adaptation, we consider the main baselines: TargetCLIP \cite{chefer2022image}, JoJoGAN \cite{chong2022jojogan}, MTG \cite{zhu2021mind}, GOSA \cite{zhang2022generalized}, DiFa \cite{zhang2022towards}, DomMod \cite{alanov2022hyperdomainnet}. For each method we use its official source code and its default hyperparameters. We train all methods for 300 iterations by ADAM optimization. For ADAM we use the same hyperparameters for all methods: $lr = 0.002, betas = (0.0, 0.999), wd = 0$. 

\subsubsection{Analysis of Different Base Models for DomMod and StyleSpace}
DomMod \cite{alanov2022hyperdomainnet} and StyleSpace are parameterizations that can be applied to any other baseline method. We examine which model is optimal for these parameterizations. We consider four methods, JoJoGAN, MTG, GOSA and DiFa, as the state-of-the-art one-shot adaptation methods. We omit TargetCLIP because it shows poor performance. 

So, we apply DomMod and StyleSpace parameterizations to JoJoGAN, MTG, GOSA and DiFa methods and provide results in \Cref{table:base_model_comparison} and in Figure. We observe that original DiFa method achieves the best Quality while having less diversity than other three baselines. However, if we train it with DomMod and StyleSpace parameterizations we obtain well-balanced performance in terms of Quality and Diversity. For other base models these parameterizations show worse Quality without significant improvement of Diversity. Therefore, for both parameterizations we choose DiFa as the base model. 

\begin{table*}[!h]
\centering
\caption{Quality and Diversity metrics \cite{alanov2022hyperdomainnet} for DomMod \cite{alanov2022hyperdomainnet} and StyleSpace parameterizations with different base models. We observe these parameterizations applied to DiFa model achieve the most balanced performance in terms of Quality and Diversity. }
	\label{table:base_model_comparison}
  \begin{tabular}{ lllllll }
    \toprule
    & & \multicolumn{2}{c}{\textbf{Across 21 domains}} \\
\cmidrule(lr){3-4}
Method & Size & Quality & Diversity \\
\midrule\midrule
JoJoGAN \cite{chong2022jojogan} & 30M & $0.602 \pm 0.049$ & $0.246 \pm 0.026$ \\
DomMod (JoJoGAN) \cite{alanov2022hyperdomainnet} & 6K & $0.478 \pm 0.081$ & $0.296 \pm 0.072$ \\
StyleSpace (JoJoGAN) & 6K & $0.479 \pm 0.083$ & $0.308 \pm 0.058$ \\
\midrule
MTG \cite{zhu2021mind} & 30M & $0.591 \pm 0.049$ & $0.258 \pm 0.025$ \\
DomMod (MTG) \cite{alanov2022hyperdomainnet} & 6K & $0.583 \pm 0.050$ & $0.268 \pm 0.027$ \\
StyleSpace (MTG) & 6K & $0.573 \pm 0.052$ & $0.279 \pm 0.025$ \\
\midrule
GOSA \cite{zhang2022generalized} & 30M & $0.592 \pm 0.046$ & $0.244 \pm 0.029$ \\
DomMod (GOSA) \cite{alanov2022hyperdomainnet} & 6K & $0.555 \pm 0.042$ & $0.189 \pm 0.020$ \\
StyleSpace (GOSA) & 6K & $0.540 \pm 0.032$ & $0.246 \pm 0.026$ \\
\midrule
DiFa \cite{zhang2022towards} & 30M & $0.722 \pm 0.045$ & $0.221 \pm 0.034$ \\
DomMod (DiFa) \cite{alanov2022hyperdomainnet} & 6K & $0.685 \pm 0.045$ & $0.245 \pm 0.033$ \\
StyleSpace (DiFa) & 6K & $0.641 \pm 0.041$ & $0.299 \pm 0.022$ \\
    \midrule
    \bottomrule
  \end{tabular}
\end{table*}

\begin{figure*}[!h]
  \centering
  \includegraphics[width=\textwidth]{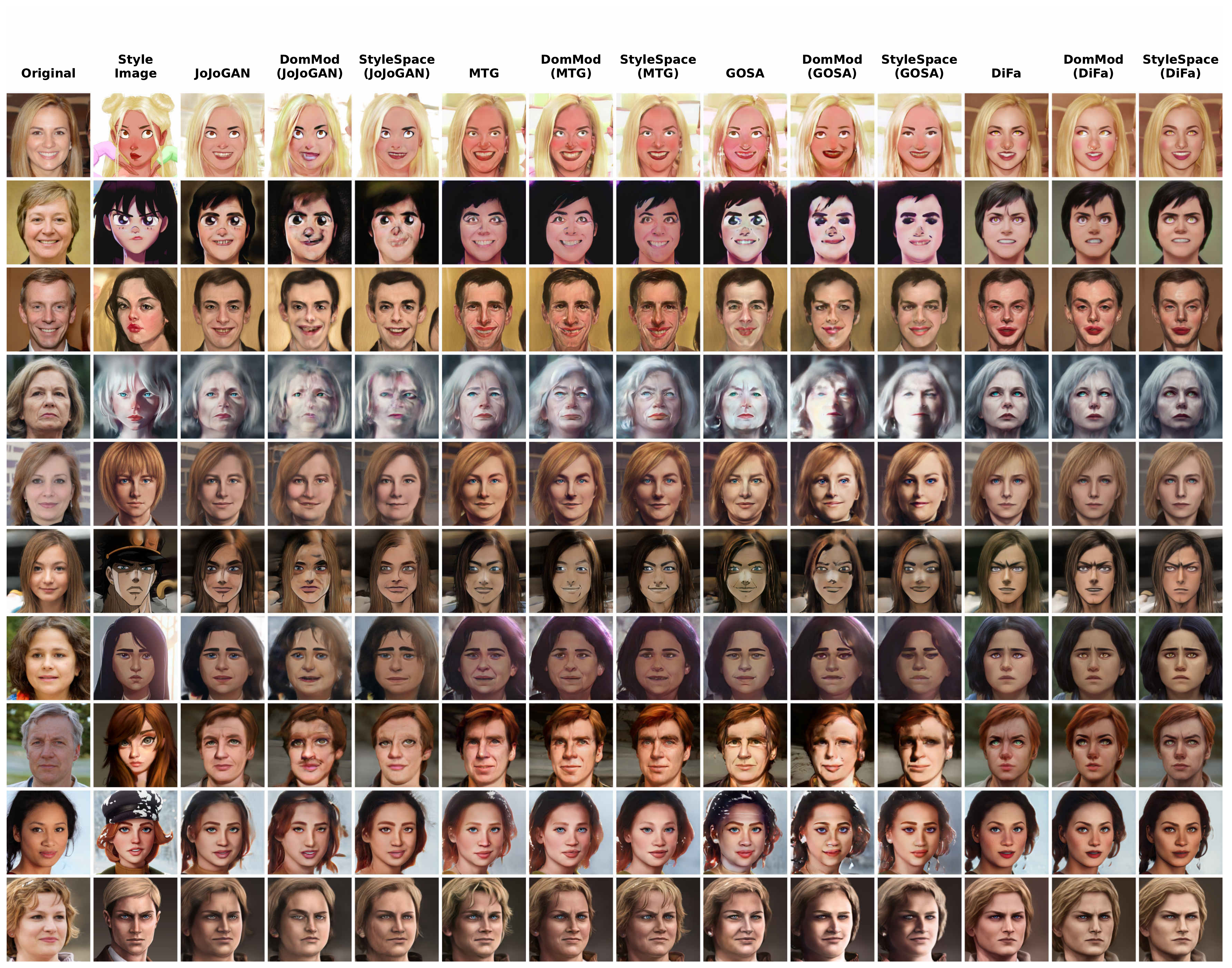}
  \caption{Text-based and one-shot image-based domain adaptation for DomMod \cite{alanov2022hyperdomainnet} and StyleSpace parameterizations with different base models.}
  \label{fig:fig3_like_0}
  \vspace{-0.5cm}
\end{figure*}

\begin{figure*}[!h]
  \centering
  \includegraphics[width=\textwidth]{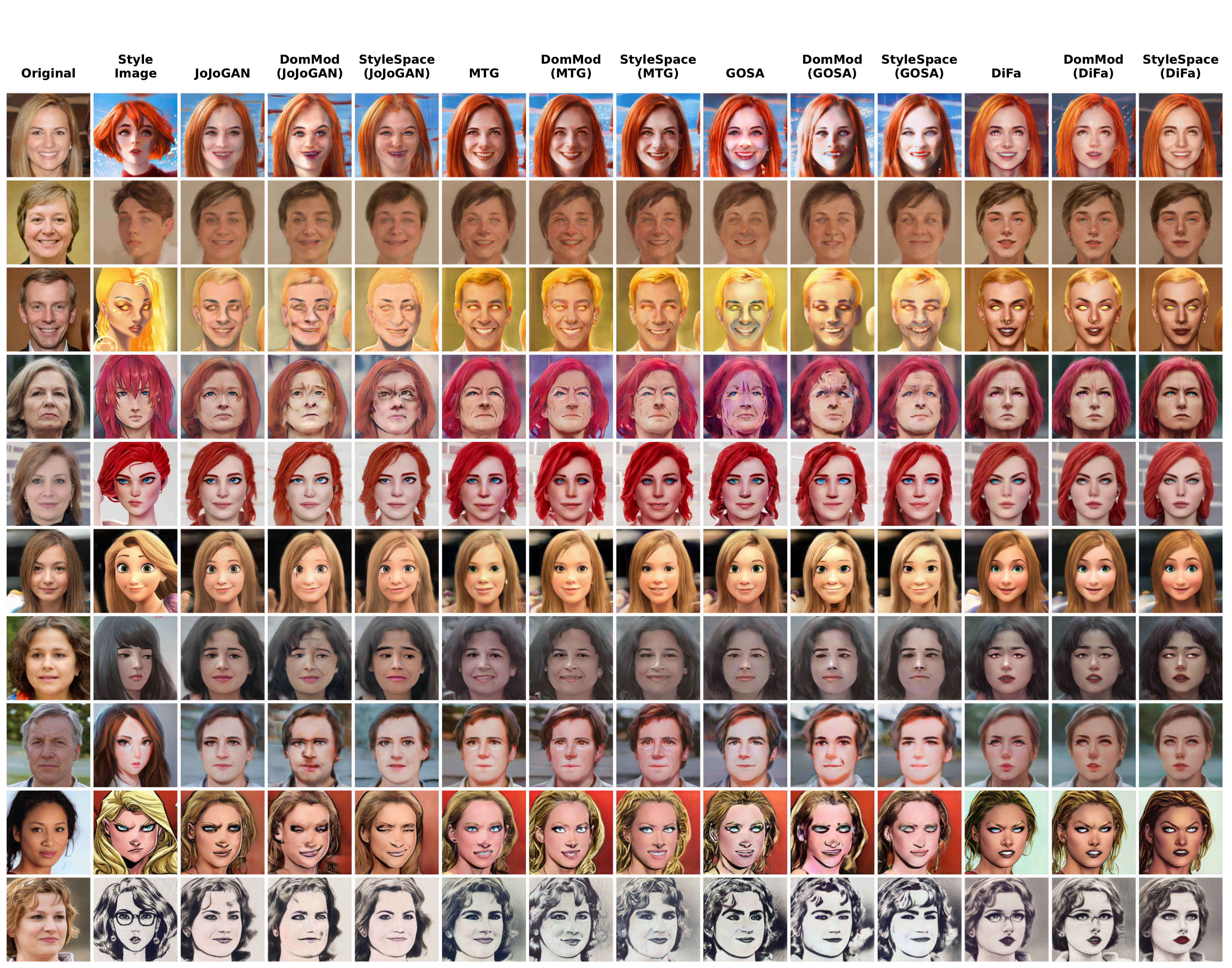}
  \caption{Text-based and one-shot image-based domain adaptation for DomMod \cite{alanov2022hyperdomainnet} and StyleSpace parameterizations with different base models.}
  \label{fig:fig3_like_0}
  \vspace{-0.5cm}
\end{figure*}

\subsubsection{Full Comparison with Baselines}
\begin{figure*}[!h]
  \centering
  \includegraphics[width=\textwidth]{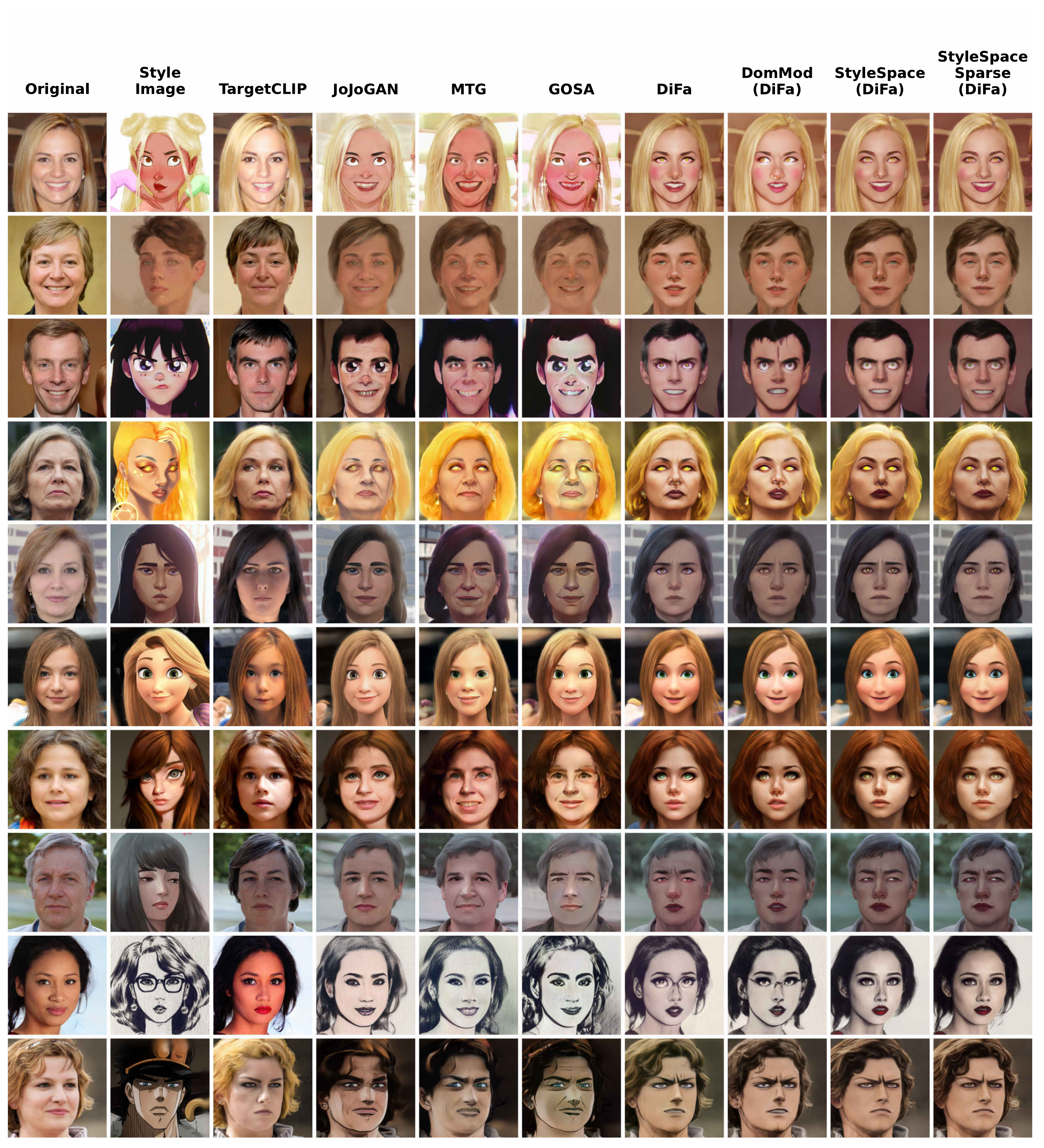}
  \caption{Comparison with baselines for one-shot image-based domain adaptation. StyleSpace and StyleSpaceSparse parameterizations achieve comparable quality as other methods while having much less trainable parameters.}
  \label{fig:a6_comparison_1}
  \vspace{-0.5cm}
\end{figure*}

\begin{figure*}[!h]
  \centering
  \includegraphics[width=\textwidth]{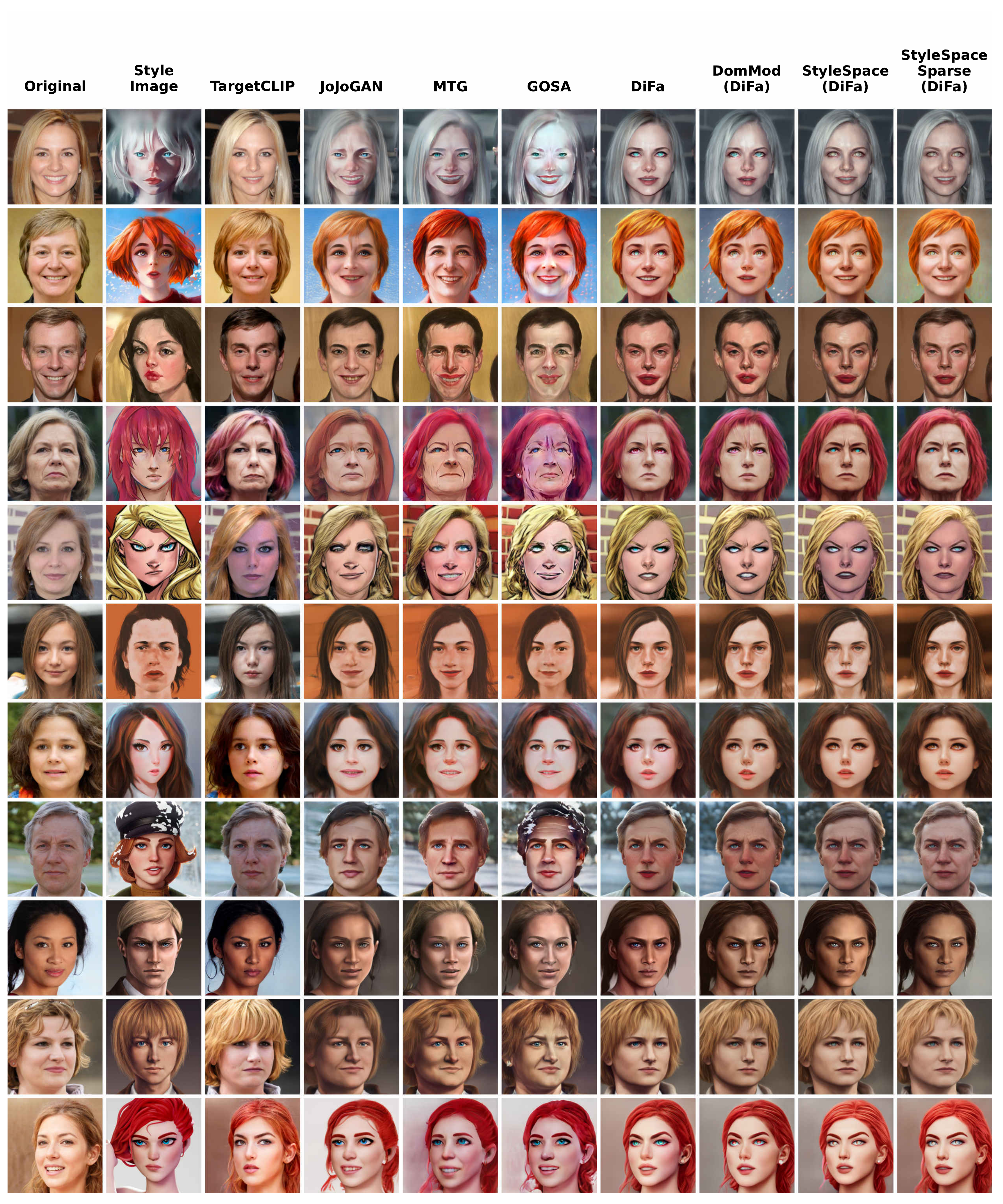}
  \caption{Comparison with baselines for one-shot image-based domain adaptation. StyleSpace and StyleSpaceSparse parameterizations achieve comparable quality as other methods while having much less trainable parameters.}
  \label{fig:a6_comparison_2}
  \vspace{-0.5cm}
\end{figure*}

Overall, we compare our StyleSpace and StyleSpaceSparse parameterizations applied to DiFa model with all other baselines on more (21) one-shot domains. We provide both quantitative and qualitative results in \Cref{table:full_one_shot_comparison} and in \Cref{fig:a6_comparison_1,fig:a6_comparison_2}, respectively. We observe similar results as in \Cref{sec:experiments}. StyleSpace and StyleSpaceSparse parameterizations achieve better results both in Quality and Diversity than JoJoGAN, MTG and GOSA baselines while having several orders less trainable parameters. 

\begin{table*}[!b]
\centering
\caption{Quality and Diversity metrics \cite{alanov2022hyperdomainnet} for one-shot image-based domain adaptation baselines. Memory denotes the memory needed for keeping adapted generators for all 21 domains for each method. StyleSpace and StyleSpaceSparse parameterizations achieve results comparable to other baselines while having much less trainable parameters. }
	\label{table:full_one_shot_comparison}
  \begin{tabular}{ llllllll }
    \toprule
    & & &  \multicolumn{2}{c}{\textbf{Across 21 domains}} \\
\cmidrule(lr){4-5}
Method & Size & Memory & Quality & Diversity \\
\midrule\midrule
JoJoGAN \cite{chong2022jojogan} & 30M & 3.15GB & $0.590 \pm 0.048$ & $0.257 \pm 0.025$ \\
MTG \cite{zhu2021mind} & 30M & 3.15GB & $0.586 \pm 0.054$ & $0.263 \pm 0.028$ \\
GOSA \cite{zhang2022generalized} & 30M & 3.15GB & $0.584 \pm 0.034$ & $0.252 \pm 0.030$ \\
DiFa \cite{zhang2022towards} & 30M & 3.15GB & $0.734 \pm 0.047$ & $0.215 \pm 0.038$ \\
TargetCLIP \cite{chefer2022image} & 9.0K & 735KB & $0.491 \pm 0.043$ & $0.322 \pm 0.015$ \\
DomMod (DiFa) \cite{alanov2022hyperdomainnet} & 6.0K & 490KB & $0.679 \pm 0.049$ & $0.253 \pm 0.037$ \\
\midrule
\textbf{StyleSpace (DiFa)} & 6.0K & 490KB & $0.644 \pm 0.041$ & $0.298 \pm 0.025$ \\
\textbf{StyleSpaceSparse (DiFa)} & \textbf{1.2K} & \textbf{98.7KB} & $0.638 \pm 0.046$ & $0.299 \pm 0.026$ \\
    \midrule
    \bottomrule
  \end{tabular}
\end{table*}

\FloatBarrier

\subsubsection{User Study}
\label{app:user_study}
To complement quantitative metrics we conduct user studies to provide a comprehensive comparsion with baselines. We consider only DiFa, GOSA, MTG and JoJoGAN methods as the most competitive baselines. We invited 133 participants to
evaluate these methods. In the study we use 21 style domains from the paper. Given an original face and a style face (randomly selected),
we show the results generated by two models for comparison and ask the user to choose which stylization result (1) preserves the person identity better ('Identity') and (2) transfers the style image patterns more consistently ('Domainness'). We collected 320 votes in total.

The results of user studies are presented in \Cref{tab:user_studies}. We observe that our StyleSpace (DiFa) model achieve comparable quality in terms of both Identity and Domainess. In particular, StyleSpace (DiFa) is preferred in $60\%$ for Identity and in $40\%$ for Domainness in comparison with the original DiFa method. %We observe that these results are well aligned with the objective metrics we report in the paper.  

\begin{table}[!h]
\caption{Statistical results of user studies. Percentage shows what proportion of users picked our StyleSpace (DiFa) model in paired comparison with each corresponding baseline.}
\label{tab:user_studies}
\begin{adjustbox}{center}
\begin{tabular}{l|cccc}
\toprule
Model    &  DiFa &  GOSA & MTG & JoJoGAN      \\
\midrule
Identity        &  $60\%$  &  $45\%$ & $39\%$ & $44\%$  \\
Domainess    &  $40\%$  &  $46\%$ & $52\%$ & $46\%$  \\
\bottomrule
\end{tabular}
\end{adjustbox}
\end{table}
\FloatBarrier

\subsection{Few-Shot Domain Adaptation}
\label{app:few_shot_adaptation}
\subsubsection{Setup}
We consider three main baselines for few-shot domain adaptation: StyleGAN-ADA \cite{karras2020training} (denoted as ADA), CDC \cite{ojha2021few} and AdAM \cite{zhao2022few}. We compare our parameterizations Affine$+$ and AffineLight$+$ applied to ADA with these baselines. For each method we use its official source code and its default hyperparameters. For these models we consider 5 few-shot settings with different available shots: 10, 25, 50, 100, 200. We train all methods for 50K number of iterations with the same batch size of 4. Then, for each method, we report the best FID value that it obtains during training. 

For few-shot experiments we consider four datasets: Dogs and Cats \cite{choi2020stargan}, LSUN Car and Church \cite{yu2015lsun}. 

\subsubsection{Results}
\begin{figure*}[!h]
  \centering
  \includegraphics[width=\textwidth]{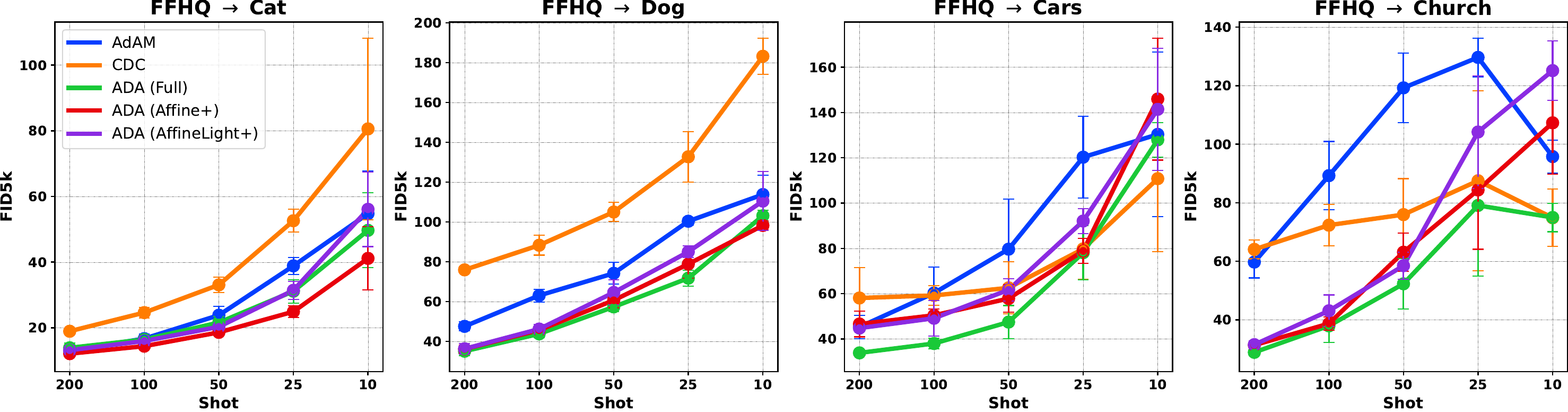}
  \caption{Few-shot training results for different number of shots.}
  \label{fig:fig_few_shot_with_err}
\end{figure*}

\begin{figure*}[!h]
  \centering
  \includegraphics[width=\textwidth]{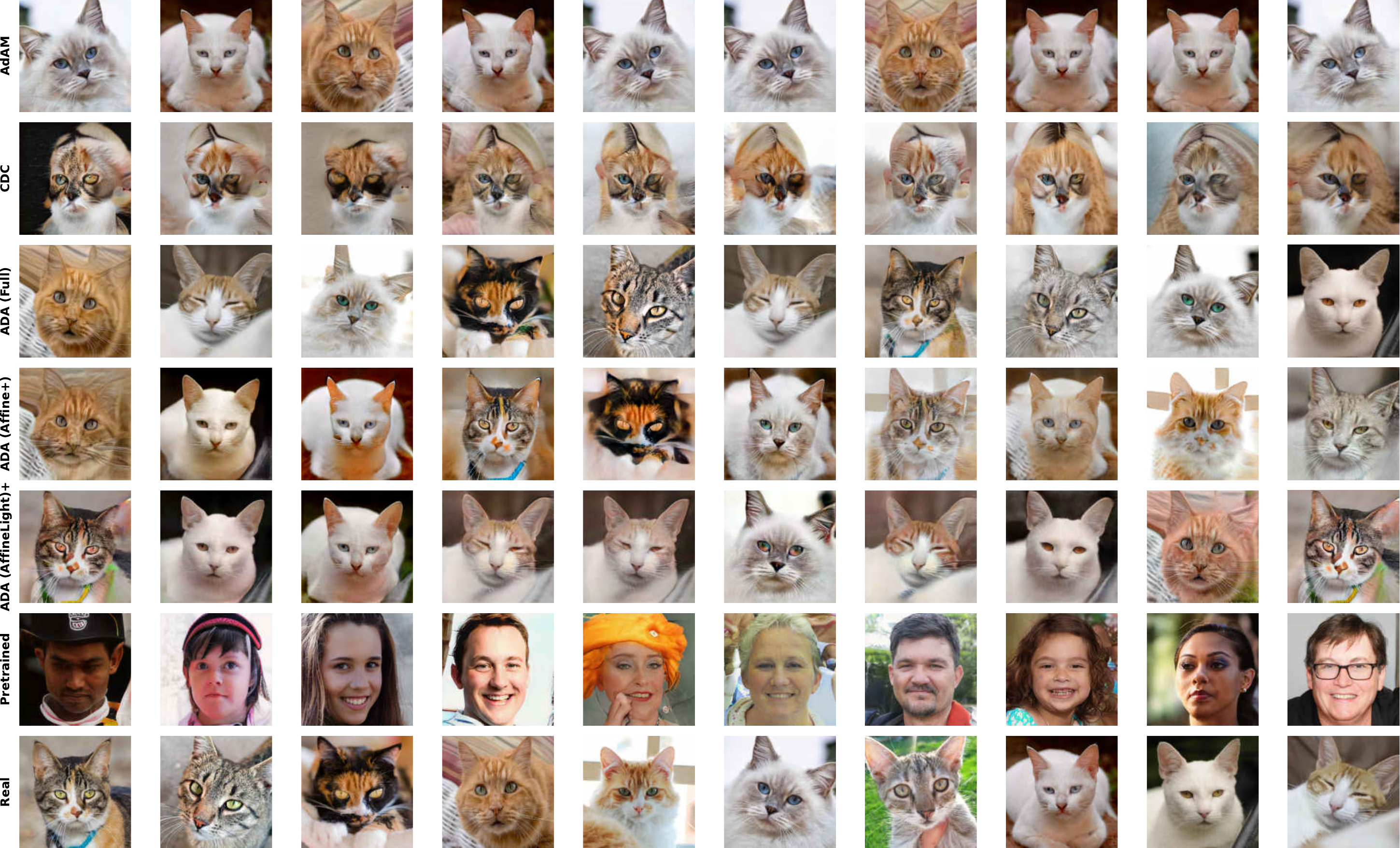}
  \caption{Few-shot training results for Cats with 10 shots.}
  \label{fig:10_shot_cat}
\end{figure*}

\begin{figure*}[!h]
  \centering
  \includegraphics[width=\textwidth]{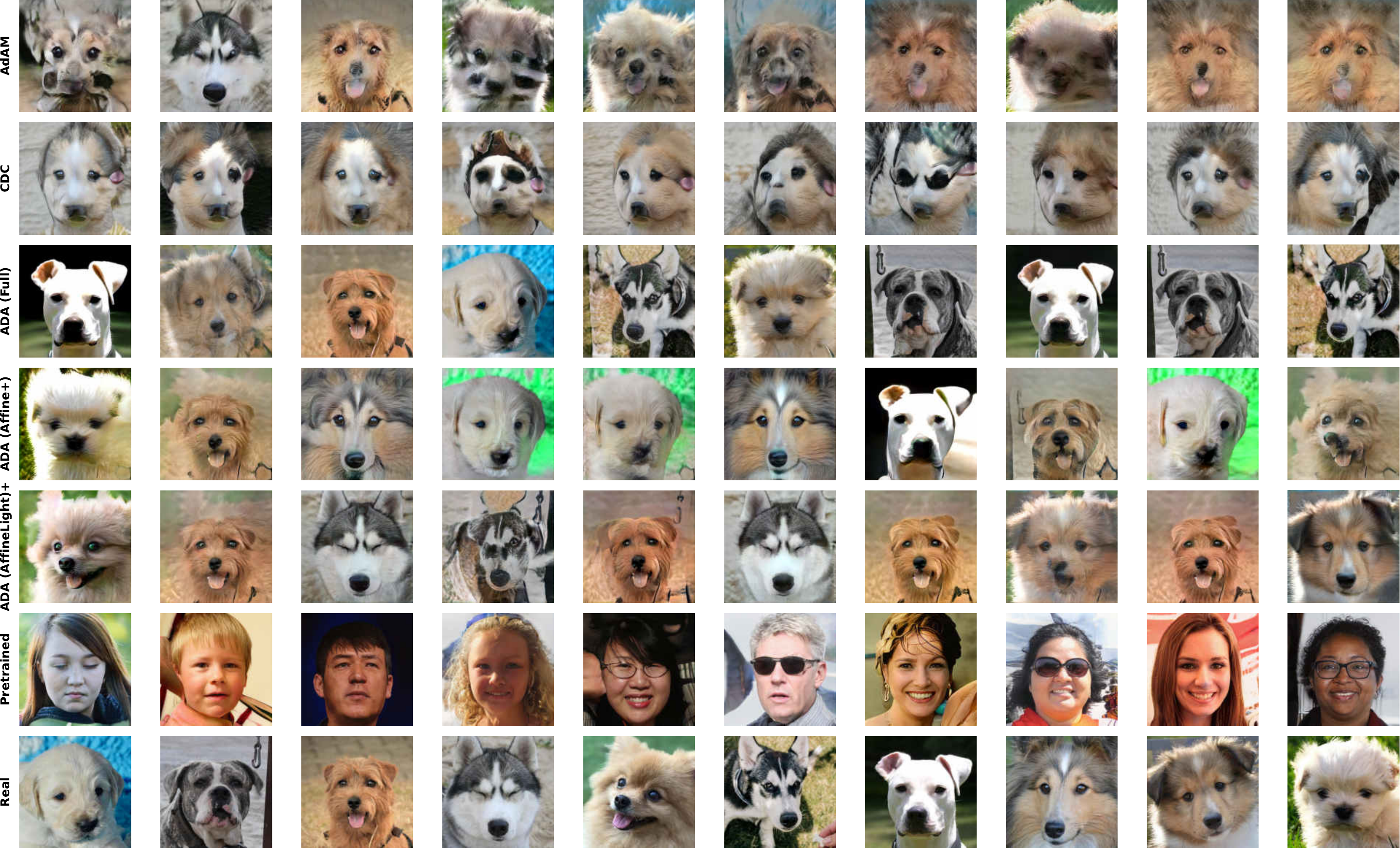}
  \caption{Few-shot training results for Dogs with 10 shots.}
  \label{fig:10_shot_dog}
\end{figure*}

\begin{figure*}[!h]
  \centering
  \includegraphics[width=\textwidth]{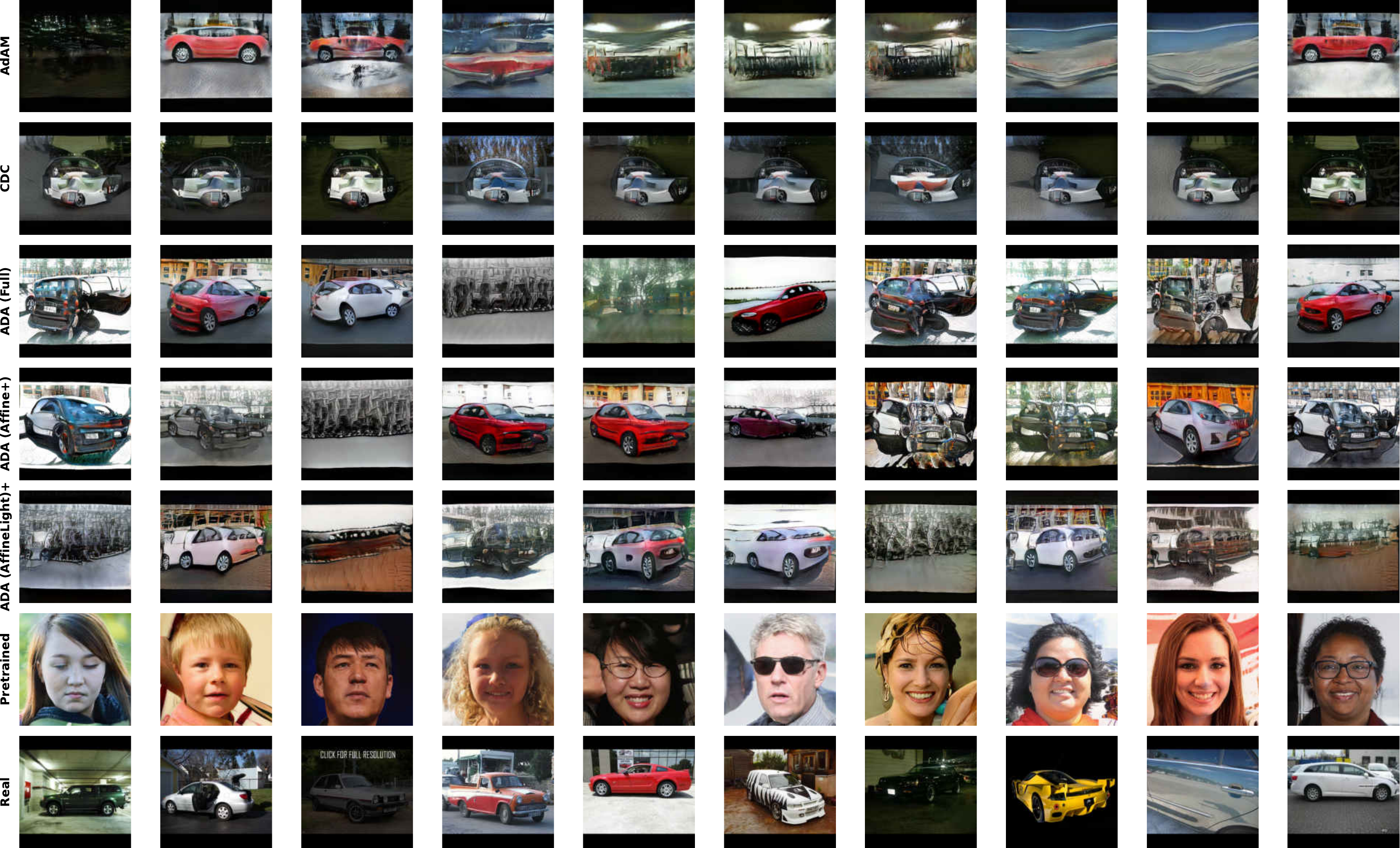}
  \caption{Few-shot training results for Cars with 10 shots.}
  \label{fig:10_shot_car}
\end{figure*}

\begin{figure*}[!h]
  \centering
  \includegraphics[width=\textwidth]{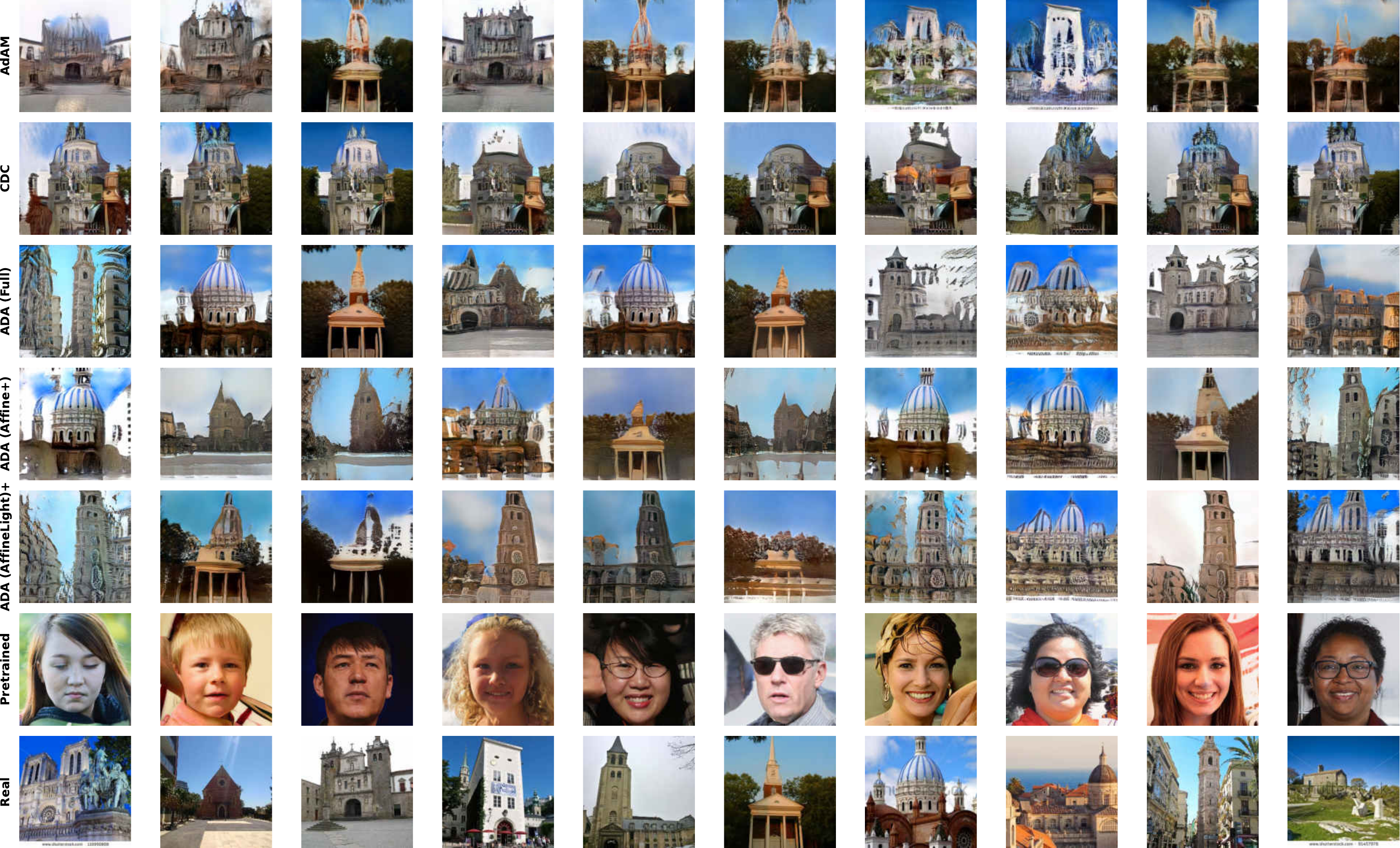}
  \caption{Few-shot training results for Churches with 10 shots.}
  \label{fig:10_shot_church}
\end{figure*}

We provide extensive quantitative results in \Cref{table:afhqcat_256,table:afhqdog_256,table:car_256,table:church_256} and in \Cref{fig:fig_few_shot_with_err} and sample examples for 10-shot setting in \Cref{fig:10_shot_cat,fig:10_shot_dog,fig:10_shot_car,fig:10_shot_church} for all four datasets with different number of shots. We observe that Affine+ and AffineLight+ parameterizations achieve comparable results on most few-shot setting with baselines while having much less trainable parameters.

\begin{table*}[!h]
\centering
\caption{Results for few-shot training for Cats with different shots.}
	\label{table:afhqcat_256}
	
    \begin{tabular}{lllllll}
    \toprule
    Shots &  Size & 10 & 25 &     50 &   100 &      200\\
    \midrule\midrule
    AdAM     & 19M    & $54.8 \pm 12.9$ &  $38.9 \pm 2.6$ & $23.8 \pm 2.6$  & $16.7 \pm 1.3$  &  $13.0 \pm 1.4$  \\
    CDC     & 30M     & $80.6 \pm 27.7$ &  $52.6 \pm 3.5$ & $33.1 \pm 2.4$  & $24.6 \pm 1.6$  &  $19.0 \pm 0.9$  \\
    ADA (Full)   & 30M      & $49.7 \pm 11.4$ &  $31.1 \pm 3.6$ & $21.6 \pm 1.9$  & $16.4 \pm 0.7$  &  $13.9 \pm 1.4$  \\
    \midrule
    ADA (Affine+)   & 5.1M    & $41.1 \pm 9.6$  &  \textbf{24.9 $\pm$ 1.7} & \textbf{18.6 $\pm$ 0.9}  & \textbf{14.4 $\pm$ 0.1}  &  \textbf{12.1 $\pm$ 0.6}  \\
    ADA (AffineLight+) & \textbf{0.6M} & $56.1 \pm 11.4$ &  $31.4 \pm 2.7$ & $20.3 \pm 0.9$  & $16.0 \pm 1.5$  &  $13.2 \pm 1.3$  \\
    \midrule\bottomrule
    \end{tabular}
  \vspace{-0.5cm}
\end{table*}

\begin{table*}[!h]
\centering
\caption{Results for few-shot training for Dogs with different shots.}
	\label{table:afhqdog_256}
	
    \begin{tabular}{lllllll}
    \toprule
    Shots &  Size & 10 & 25 &     50 &   100 &      200\\
    \midrule\midrule
    AdAM     & 19M         & $113.7 \pm 9.6$   &  $100.3 \pm 1.8$   &   $74.1 \pm 5.5$    &   $63.0 \pm 3.2$ &  $47.5 \pm 2.3$   \\
    CDC     & 30M          & $183.2 \pm 9.2$   &  $132.7 \pm 12.7$  &   $105.0 \pm 4.8$    &  $88.3 \pm 5.0$ &  $75.9 \pm 1.6$   \\
    ADA (Full)   & 30M          & $103.1 \pm 2.7$   &  $71.7 \pm 4.0$    &   $57.3 \pm 2.3$    &   $43.7 \pm 1.9$ &  $35.0 \pm 2.3$   \\
    \midrule
    ADA (Affine+)   & 5.1M     & $98.2 \pm 2.3$    &  $78.8 \pm 7.0$    &   $60.6 \pm 3.7$    &   $45.6 \pm 1.7$ &  $36.1 \pm 1.9$   \\
    ADA (AffineLight+) & \textbf{0.6M} & $110.4 \pm 14.8$  &  $85.0 \pm 2.9$    &   $64.5 \pm 6.3$    &   $46.2 \pm 2.4$ &  $36.2 \pm 2.7$   \\
    \midrule\bottomrule
    \end{tabular}
  \vspace{-0.5cm}
\end{table*}
\begin{table*}[!h]
\centering
\caption{Results for few-shot training for Cars with different shots.}
	\label{table:car_256}
	
    \begin{tabular}{lllllll}
    \toprule
    Shots &  Size & 10 & 25 &     50 &   100 &      200\\
    \midrule\midrule
    AdAM     & 19M    & $130.3 \pm 36.3$  &  $120.3 \pm 18.0$  &   $79.6 \pm 22.0$    &  $60.3 \pm 11.5$  &  $45.3 \pm 5.2$    \\
    CDC     & 30M     & $110.8 \pm 32.3$  &  $79.8 \pm 13.5$   &   $62.5 \pm 11.5$    &  $59.2 \pm 4.2$  &  $58.1 \pm 13.4$    \\
    ADA (Full)   & 30M      & $127.9 \pm 7.6$   &  $78.1 \pm 11.9$   &   $47.4 \pm 7.3$    &   \textbf{$38.0 \pm 2.3$}  &  \textbf{$33.8 \pm 1.3$}     \\
    \midrule
    ADA (Affine+)   & 5.1M   & $145.9 \pm 26.9$  &  $78.9 \pm 5.5$    &   $57.8 \pm 6.1$    &   $50.4 \pm 2.9$  &  $46.6 \pm 5.7$     \\
    ADA (AffineLight+) & \textbf{0.6M} & $141.4 \pm 27.0$  &  $92.1 \pm 5.5$    &   $61.7 \pm 4.8$    &   $49.1 \pm 7.9$  &  $44.8 \pm 1.4$     \\
    \midrule\bottomrule
    \end{tabular}
  \vspace{-0.5cm}
\end{table*}
\begin{table*}[!h]
\centering
\caption{Results for few-shot training for Churches with different shots.}
	\label{table:church_256}
	
    \begin{tabular}{lllllll}
    \toprule
    Shots &  Size & 10 & 25 &     50 &   100 &      200\\
    \midrule\midrule
    AdAM     & 19M         & $95.8 \pm 5.7$    & $129.7 \pm 6.6$      &  $119.3 \pm 11.9$    &  $89.3 \pm 11.7$ &  $59.7 \pm 5.4$   \\
    CDC     & 30M          & $74.9 \pm 9.8$    & $87.5 \pm 30.7$      &  $75.9 \pm 12.3$    &   $72.3 \pm 7.1$  &  $64.1 \pm 3.2$   \\
    ADA (Full)   & 30M          & $75.0 \pm 4.9$    & $79.1 \pm 24.2$      &  $52.3 \pm 8.6$    &    $37.9 \pm 5.6$  &  $28.9 \pm 1.2$   \\
    \midrule
    ADA (Affine+)   & 5.1M     & $107.3 \pm 17.6$  & $84.4 \pm 20.3$      &  $63.1 \pm 6.6$    &    $38.7 \pm 2.3$  &  $31.3 \pm 1.3$   \\
    ADA (AffineLight+) & \textbf{0.6M} & $125.2 \pm 10.1$  & $104.2 \pm 19.2$     &  $58.5 \pm 4.6$    &    $43.1 \pm 5.4$  &  $31.5 \pm 1.3$   \\
    \midrule\bottomrule
    \end{tabular}
\end{table*}

\FloatBarrier
\subsubsection{Additional comparison of StyleSpace with other parameterizations in the case of few-shot domain adaptation}

To further extend analysis of parameterizations we conducted a set of experiments that compares StyleSpace with Full, Affine+, and AffineLight+ parameterizations in a few-shot learning regime. To analyze a performance gap between StyleSpace and other parameterizations we consider Dog $\rightarrow$ Cat adaptation problem. 

\begin{figure}[!h]
\minipage{0.49\textwidth}
  \includegraphics[width=0.99\textwidth]{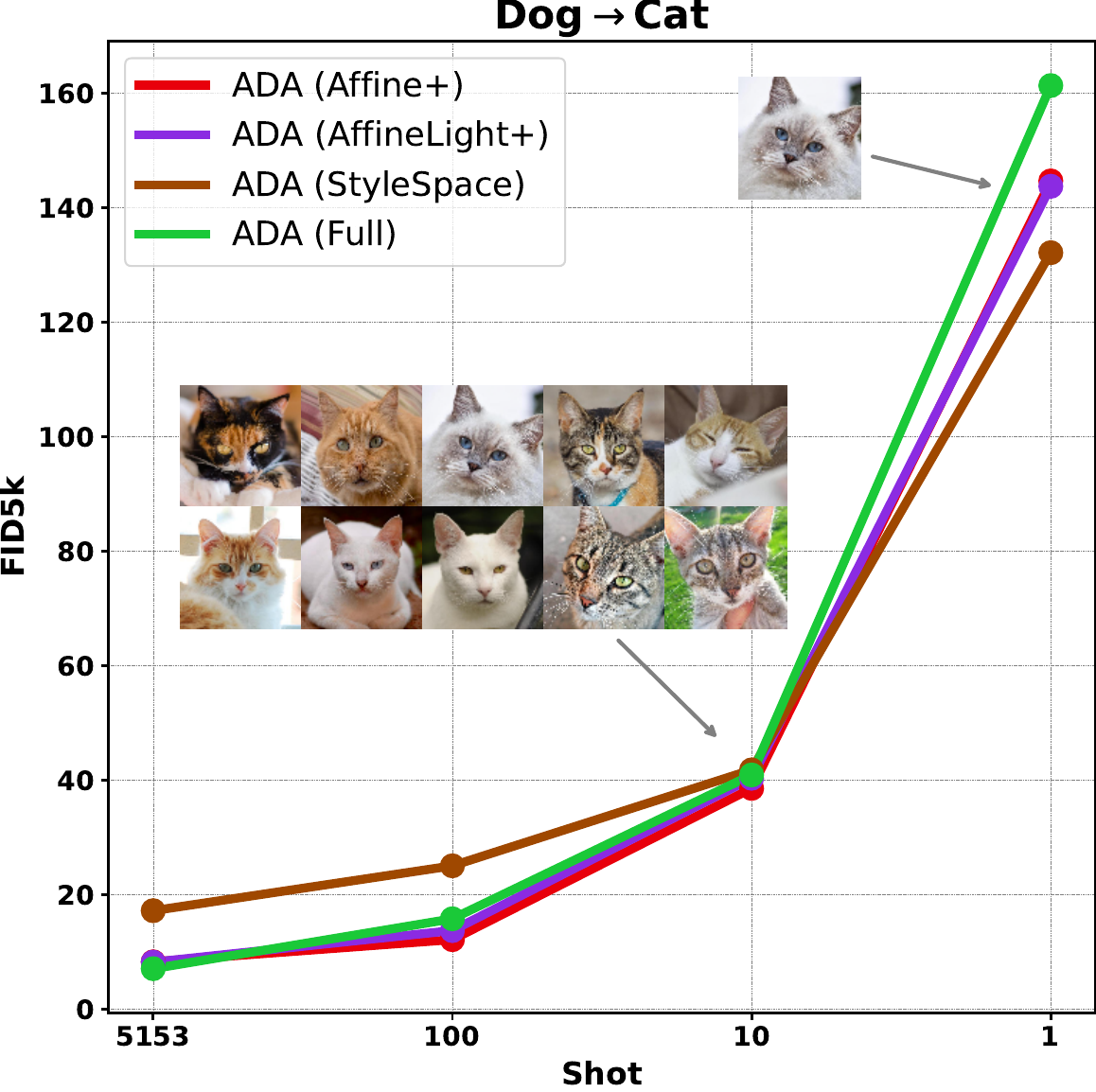}
\endminipage\hfill
\minipage{0.49\textwidth}
  \includegraphics[width=0.99\textwidth]{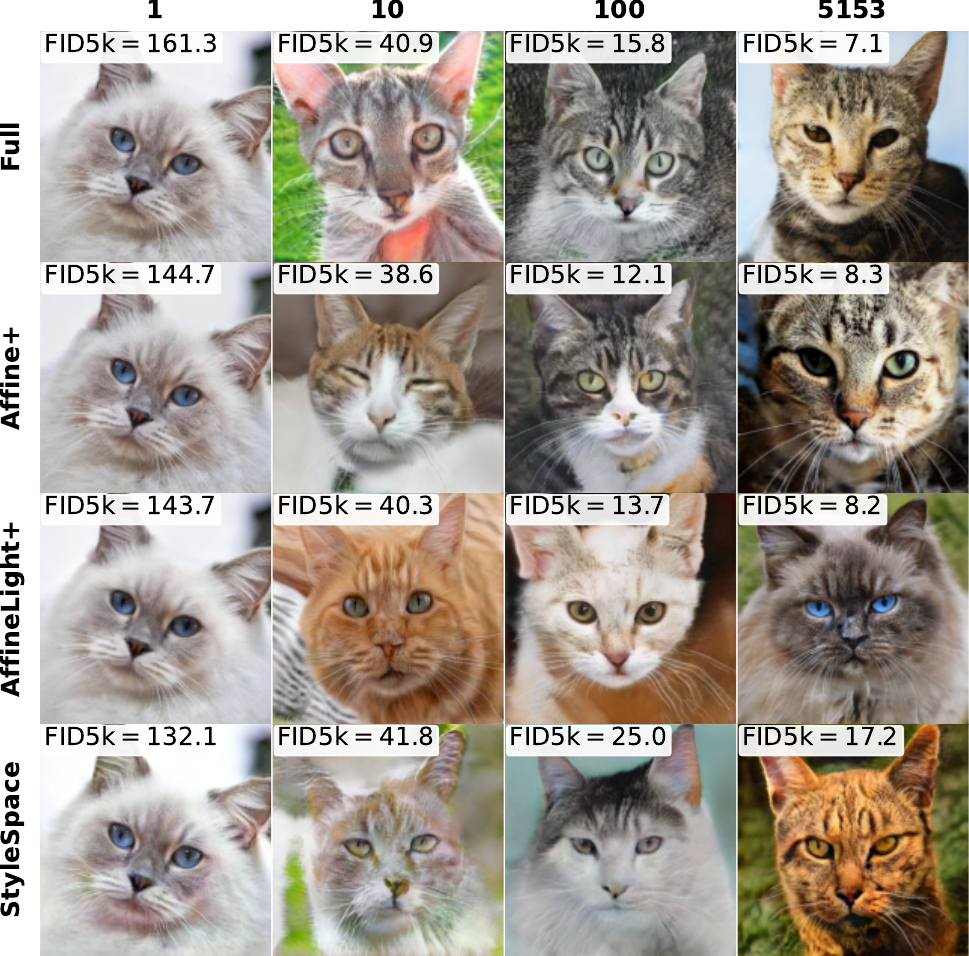}
\endminipage
\caption{Few-shot adaptation for Dog $\rightarrow$ Cat setup. We observe increasing quality gap between StyleSpace and parameterizations with more parameters when increasing number of shots. It can be seen that StyleSpace fails to overfit to generate training samples (shown on the left figure) in 10-shot scenario  due to the low number of parameters.
\label{fig:fig_few_shot_stylespace}}
\end{figure}

We provide the results in Figure~\ref{fig:fig_few_shot_stylespace}. We see that in the case of one-shot adaptation, the performance gap between all parameterizations is negligible and we observe that they successfully overfit to the one available shot. Conversely, when we increase the number of shots, the gap between StyleSpace and other parameterizations tends to grow, particularly when the number of shots surpasses $10$. It confirms that StyleSpace parameterization is not suitable for the case when we adapt to a dissimilar target domain with various modes.

\FloatBarrier

\subsection{Training Time}
\label{app:training_time}
\begin{table}[!h]
\vspace{-0.8em}
\caption{Sec/Kimg shows the average time across different domains to process thousand of images during training. Speedup shows the average increase in training time relative to Full space.}
\label{tab:training_time}
\begin{adjustbox}{width=1.0\textwidth,center}
\begin{tabular}{lccccccc}
\toprule
Space    & Full            & Mapping         & SyntConv         & Affine          & Affine+         & StyleSpace      & StyleSpace+     \\\midrule\midrule
Sec/Kimg & $97 \pm 3$      & $91 \pm 4$      & $100 \pm 1$      & $90 \pm 1$      & $93 \pm 1$      & $\mathbf{ 89 \pm 1 }$      & $93 \pm 2$      \\
Speedup  & $0.0 \pm 2.6\%$ & $6.6 \pm 4.4\%$ & $-3.2 \pm 0.8\%$ & $7.7 \pm 1.3\%$ & $4.1 \pm 1.3\%$ & $\mathbf{ 8.8 \pm 1.5\% }$ & $4.2 \pm 1.7\%$ \\
\midrule\bottomrule
\end{tabular}
\end{adjustbox}
\end{table}

To examine performance of proposed parameterizations we provide average training time for few-shot domains in Table~\ref{tab:training_time}. We observe a consistent $5-10\%$ speedup for the StyleSpace, Affine+ and AffineLight+ compared to the Full parameterization. Such difference is explained with the lack of a backward pass through Mapping Network which can only give a marginal speedup proportional to the number of its parameters.

\FloatBarrier

\subsection{Cross-domain image translation}
\label{app:i2i}
% Вставить, что не влезло + описать сетап. + wild домен

% We provide additional results for the image-to-image translation task for other datasets pairs complimentary to Figure~\ref{fig:i2i} and describe the exact experimental setup.
We provide results for the image-to-image translation task for different datasets pairs and describe the exact experimental setup.

We use the following approach to solve the I2I task (the same as in \cite{wu2021stylealign}): first, we invert a source image to the $Z_{opt}$ latent space using a source domain generator. Then, we feed this representation to a generator in the target domain. In a reference-based setup, we additionally use the target generator to obtain the latent code for a reference image and then combine the first $6$ style codes from the source image with the latest codes from the reference image. We perform $1000$ optimization steps with truncation $\psi = 0.7$, and we use truncation $\psi = 0.8$ at inference time. 

We use validation splits of the AFHQ datasets ($500$ images) for the source and reference images. For reference-based I2I, we randomly select $10$ reference images for each source image.

\begin{figure}[!h]
{
    \begin{tabular}{ccc}
        \includegraphics[width=0.30\textwidth]{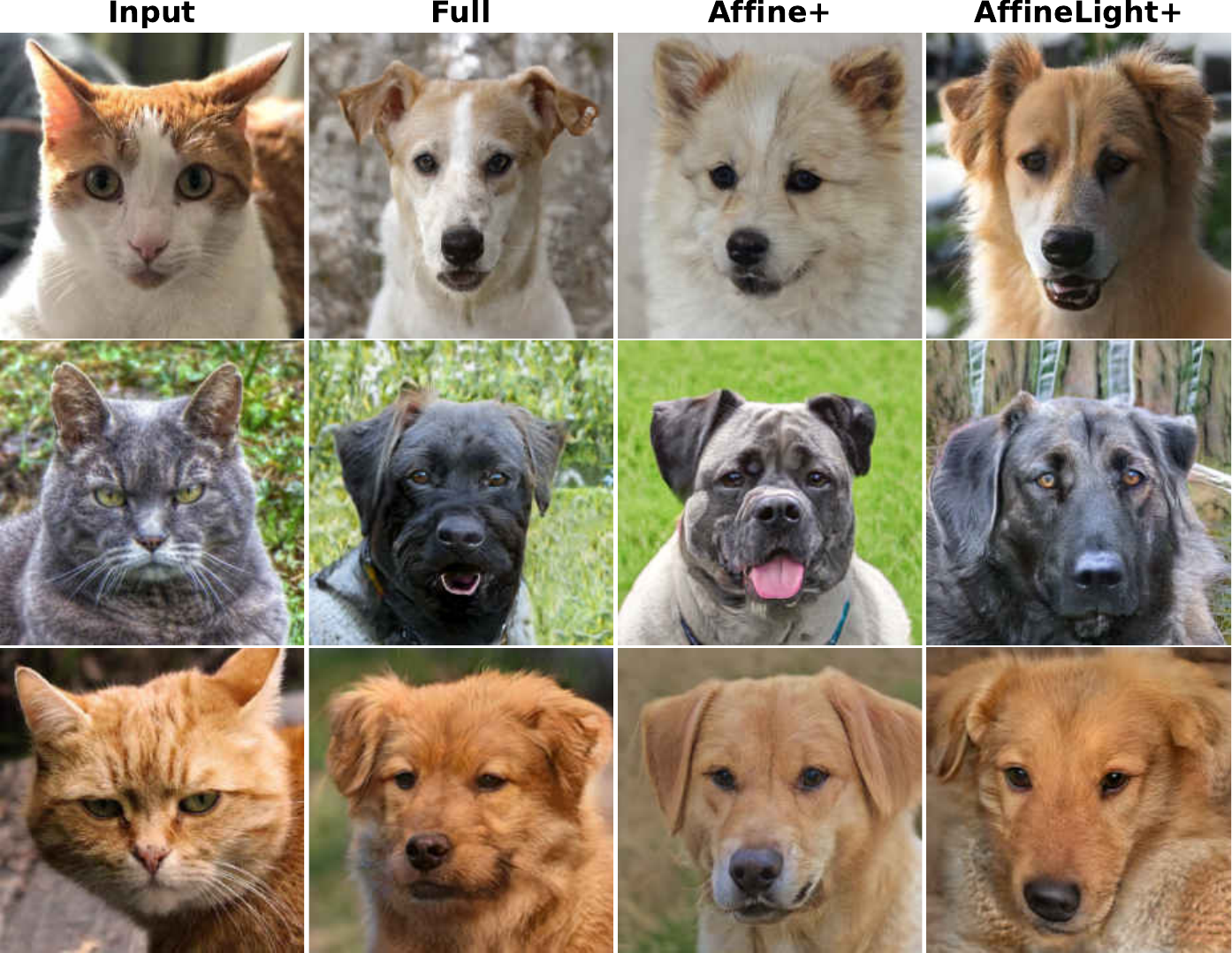} &
        \includegraphics[width=0.30\textwidth]{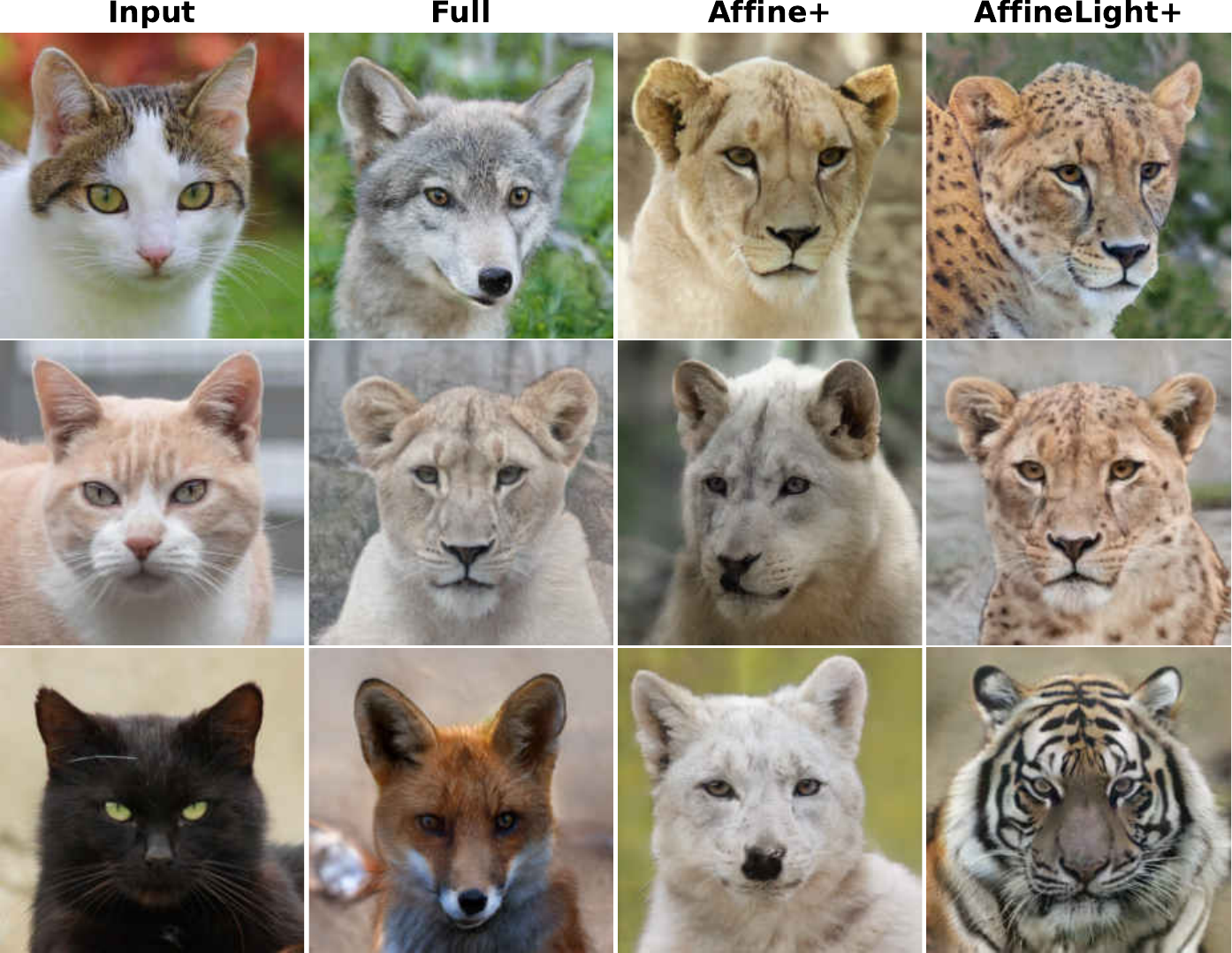} &
        \includegraphics[width=0.30\textwidth]{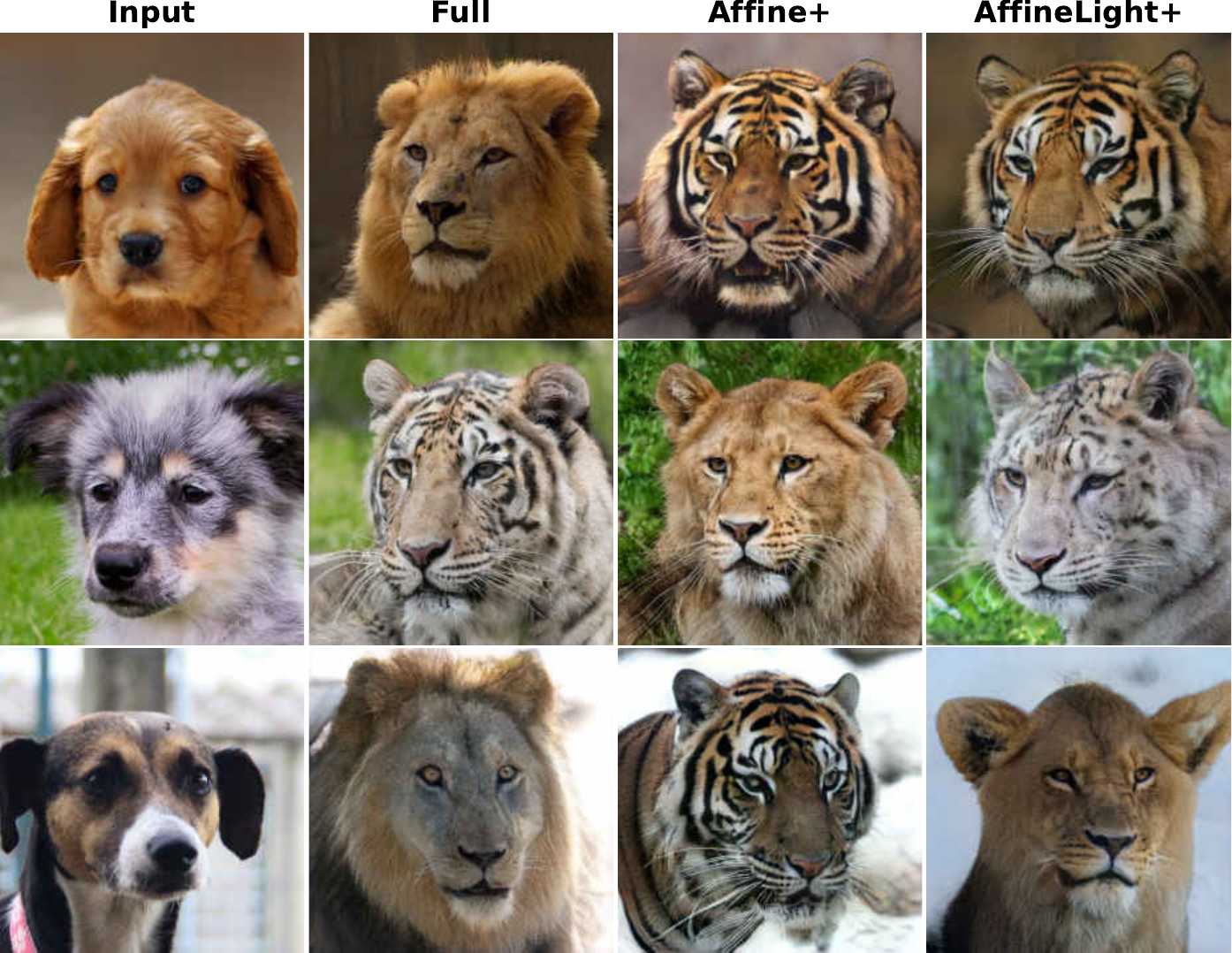} \\
        \includegraphics[width=0.30\textwidth]{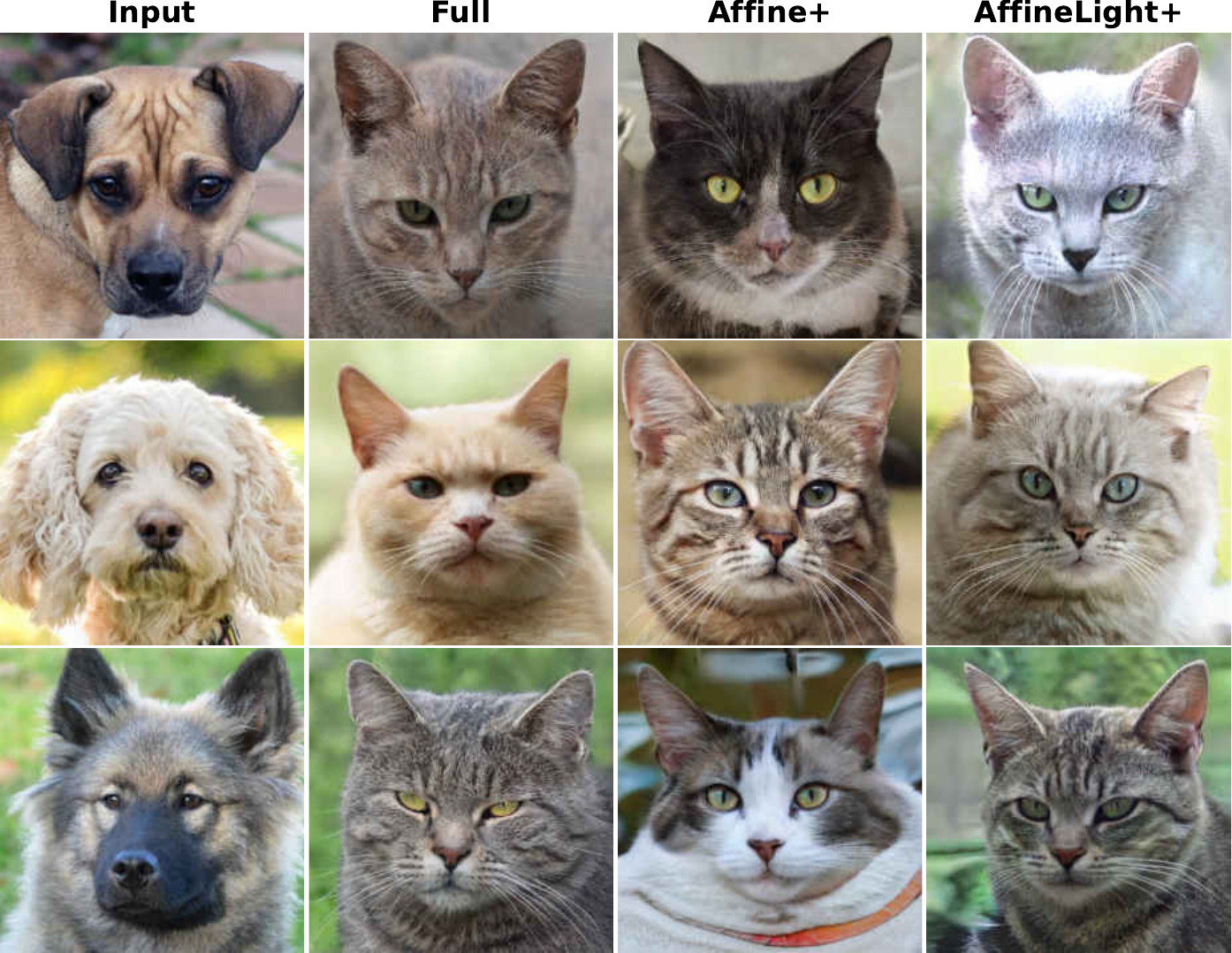} &
        \includegraphics[width=0.30\textwidth]{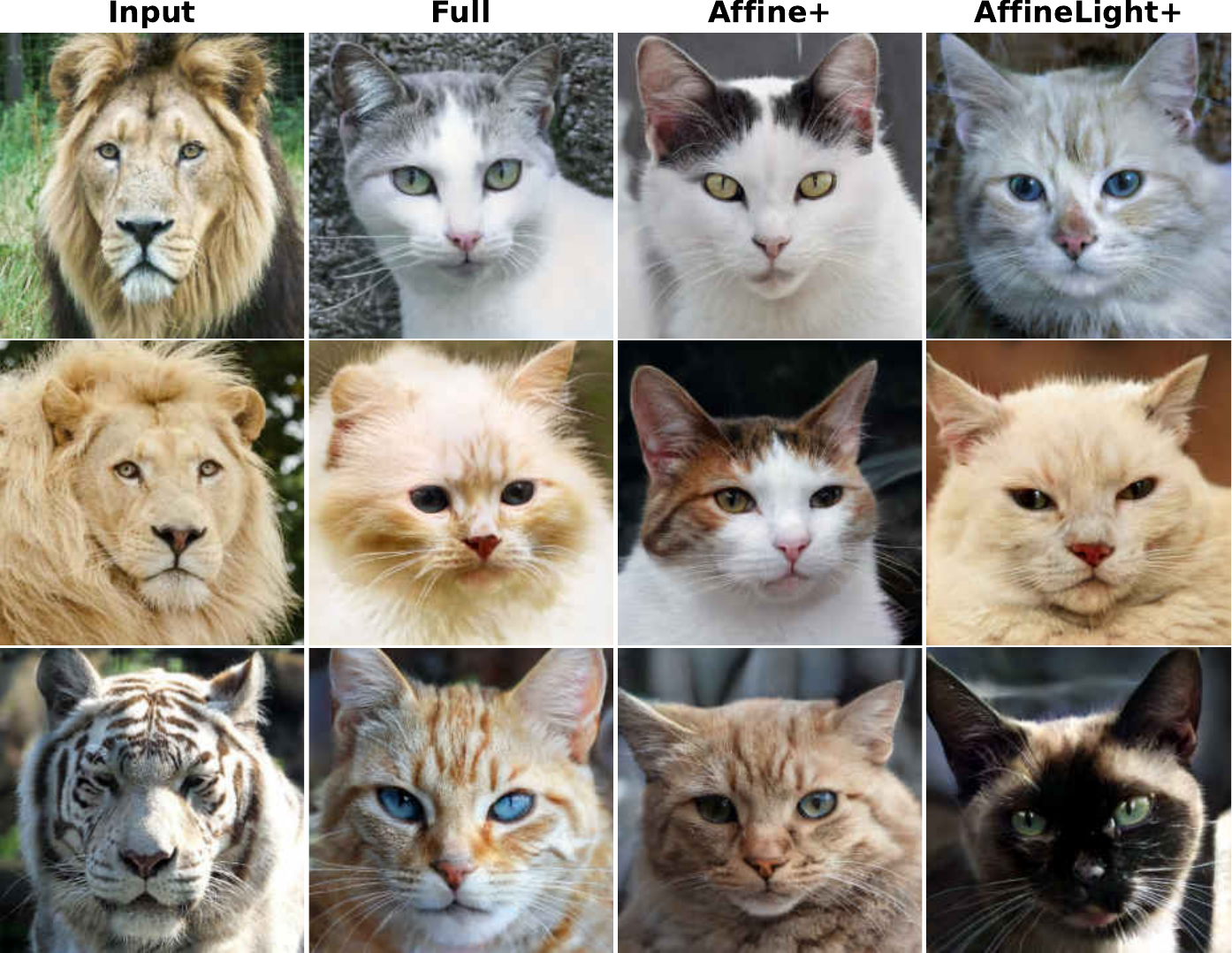} &
        \includegraphics[width=0.30\textwidth]{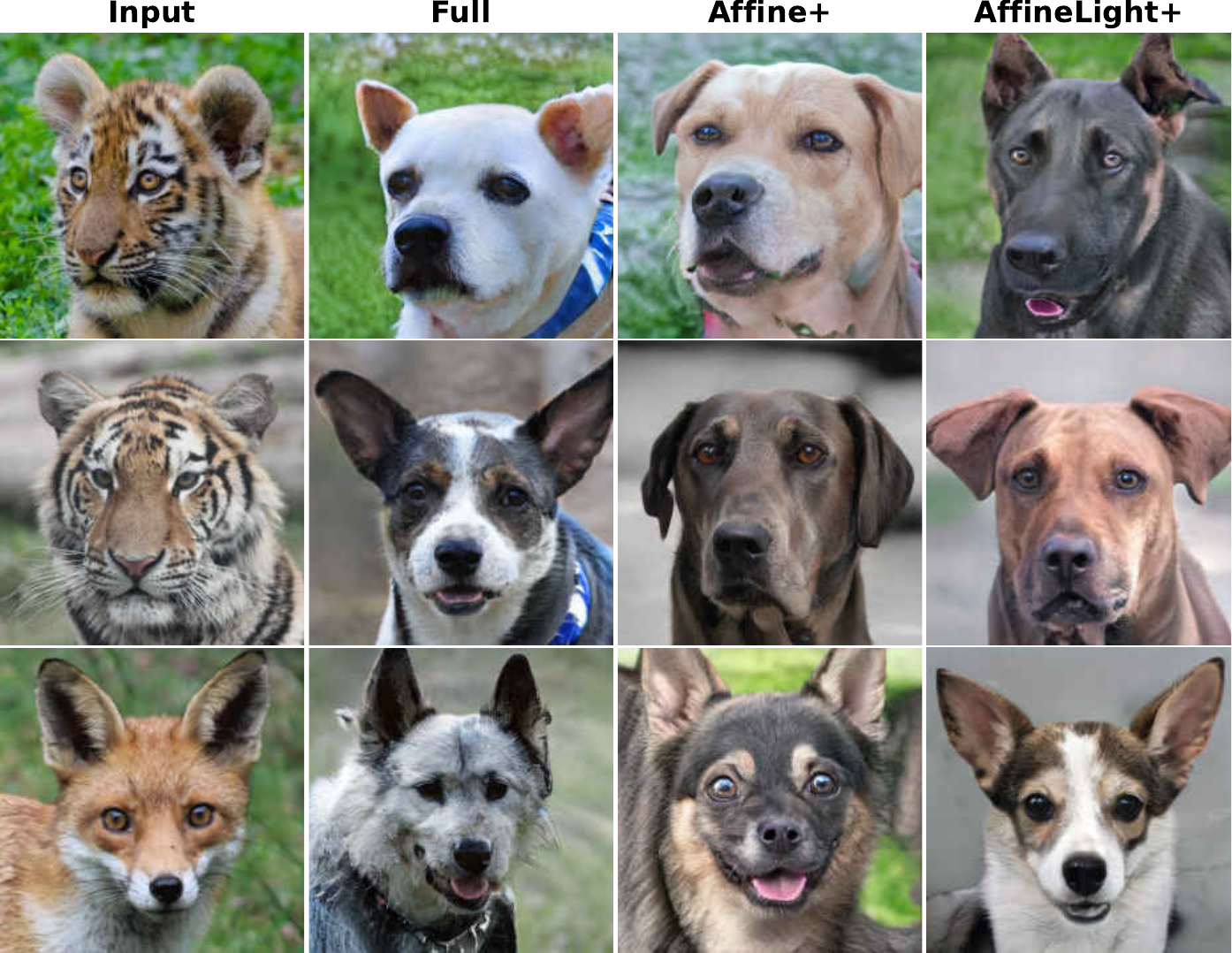} \\
    \end{tabular}
}
\centering
\begin{adjustbox}{width=0.95\textwidth,center}
% \begin{tabular}{cccccccccccccc}
% \toprule
%  &       & \multicolumn{2}{c}{cat2dog} & \multicolumn{2}{c}{cat2wild} & \multicolumn{2}{c}{dog2cat} & \multicolumn{2}{c}{dog2wild} & \multicolumn{2}{c}{wild2cat} & \multicolumn{2}{c}{wild2dog} \\
%  &   Size &     FID & KID$\times 10^{3}$ &      FID & KID$\times 10^{3}$ &     FID & KID$\times 10^{3}$ &      FID & KID$\times 10^{3}$ &      FID & KID$\times 10^{3}$ &      FID & KID$\times 10^{3}$ \\
% \midrule\midrule
% Full        &  60.6M &  $40.5$ &             $11.6$ &   $12.8$ &             $2.44$ &  $18.6$ &             $2.78$ &   $17.5$ &             $4.64$ &   $18.6$ &             $3.28$ &   $42.6$ &             $12.7$ \\
% Affine+     &  10.2M &  $42.3$ &             $12.9$ &   $16.3$ &             $4.92$ &  $18.4$ &             $2.37$ &   $13.2$ &             $3.14$ &   $21.5$ &             $3.18$ &   $41.4$ &             $10.6$ \\
% StyleSpace+ &   1.1M &  $54.8$ &             $24.9$ &   $25.4$ &             $11.2$ &  $21.5$ &             $5.00$ &   $19.9$ &             $6.09$ &   $21.5$ &             $3.49$ &   $53.8$ &             $22.0$ \\
% \midrule\bottomrule
% \end{tabular}

\begin{tabular}{cccccccccccccc}
\toprule
 &       & \multicolumn{2}{c}{cat2dog} & \multicolumn{2}{c}{cat2wild} & \multicolumn{2}{c}{dog2cat} & \multicolumn{2}{c}{dog2wild} & \multicolumn{2}{c}{wild2cat} & \multicolumn{2}{c}{wild2dog} \\
 &   Size &     FID & KID$\times 10^{3}$ &      FID & KID$\times 10^{3}$ &     FID & KID$\times 10^{3}$ &      FID & KID$\times 10^{3}$ &      FID & KID$\times 10^{3}$ &      FID & KID$\times 10^{3}$ \\
\midrule\midrule
Full        &  60.6M &  $40.5$ &             $11.6$ &   $12.8$ &             $2.44$ &  $18.6$ &             $2.78$ &   $17.5$ &             $4.64$ &   $18.6$ &             $3.28$ &   $42.6$ &             $12.7$ \\
Affine+     &  10.2M &  $42.3$ &             $12.9$ &   $16.3$ &             $4.92$ &  $18.4$ &             $2.37$ &   $13.2$ &             $3.14$ &   $21.5$ &             $3.18$ &   $41.4$ &             $10.6$ \\
AffineLight+ & 1.2M  & $45.5$ & $14.3$   & $17.1$ & $5.01$   & $19.5$ & $3.91$   & $13.8$ & $3.48$   & $21.3$ & $3.68$   & $48.2$ & $16.1$ \\
\midrule\bottomrule
\end{tabular}
\end{adjustbox}
\caption{Comparison of I2I translation for different parameterizations. I2I with the generator in $S+$ demonstrates the ability to capture the pose and the style from the source image while producing natural-looking results despite the presence of a gap in the metrics. At the same time, Affine+ parameterization can significantly reduce this gap.}
\label{fig:app_i2i_unconditional}
\end{figure}

\begin{figure}[!h]
{
\centering
    \begin{tabular}{ccc}
        \includegraphics[width=0.30\textwidth]{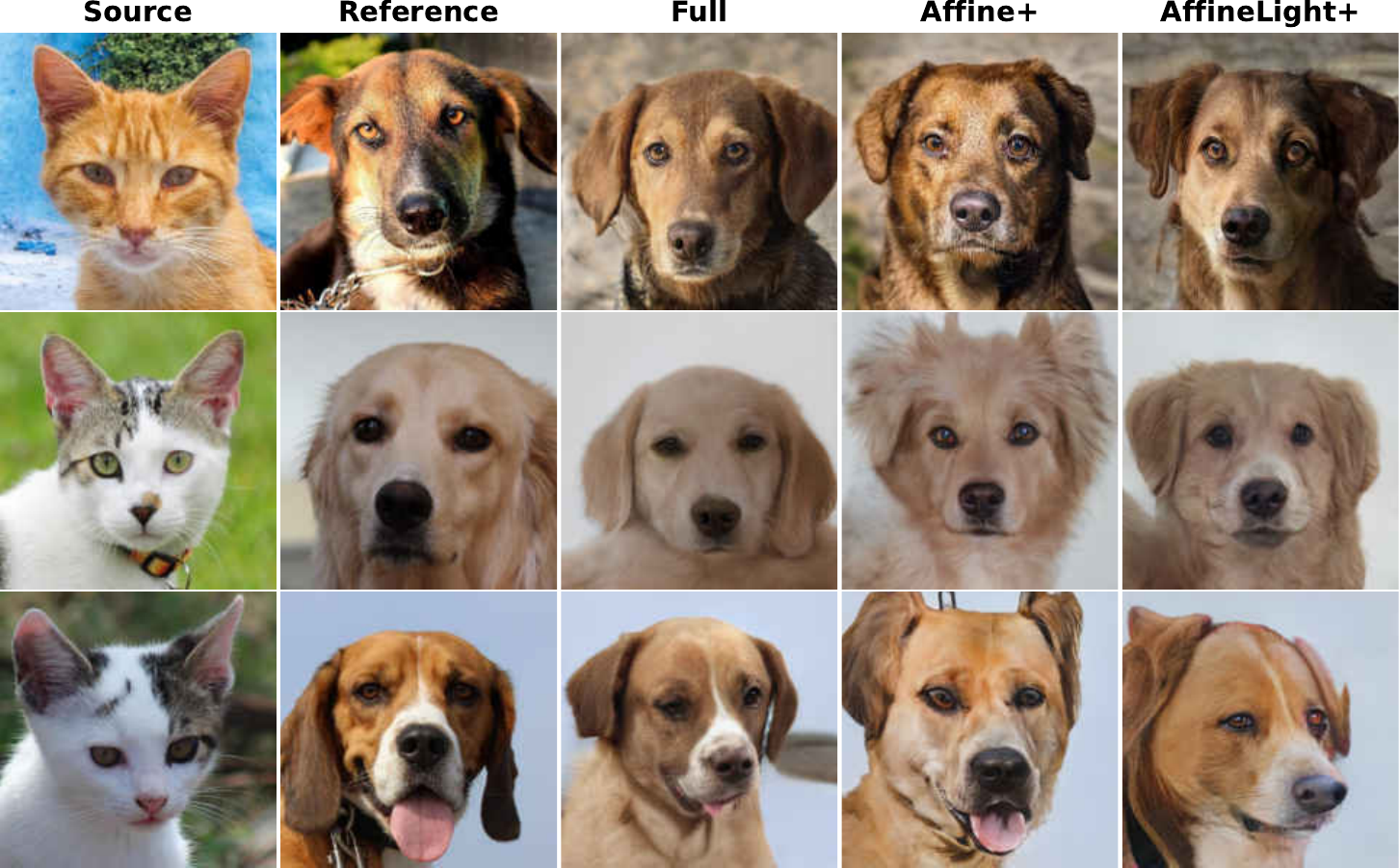} &
        \includegraphics[width=0.30\textwidth]{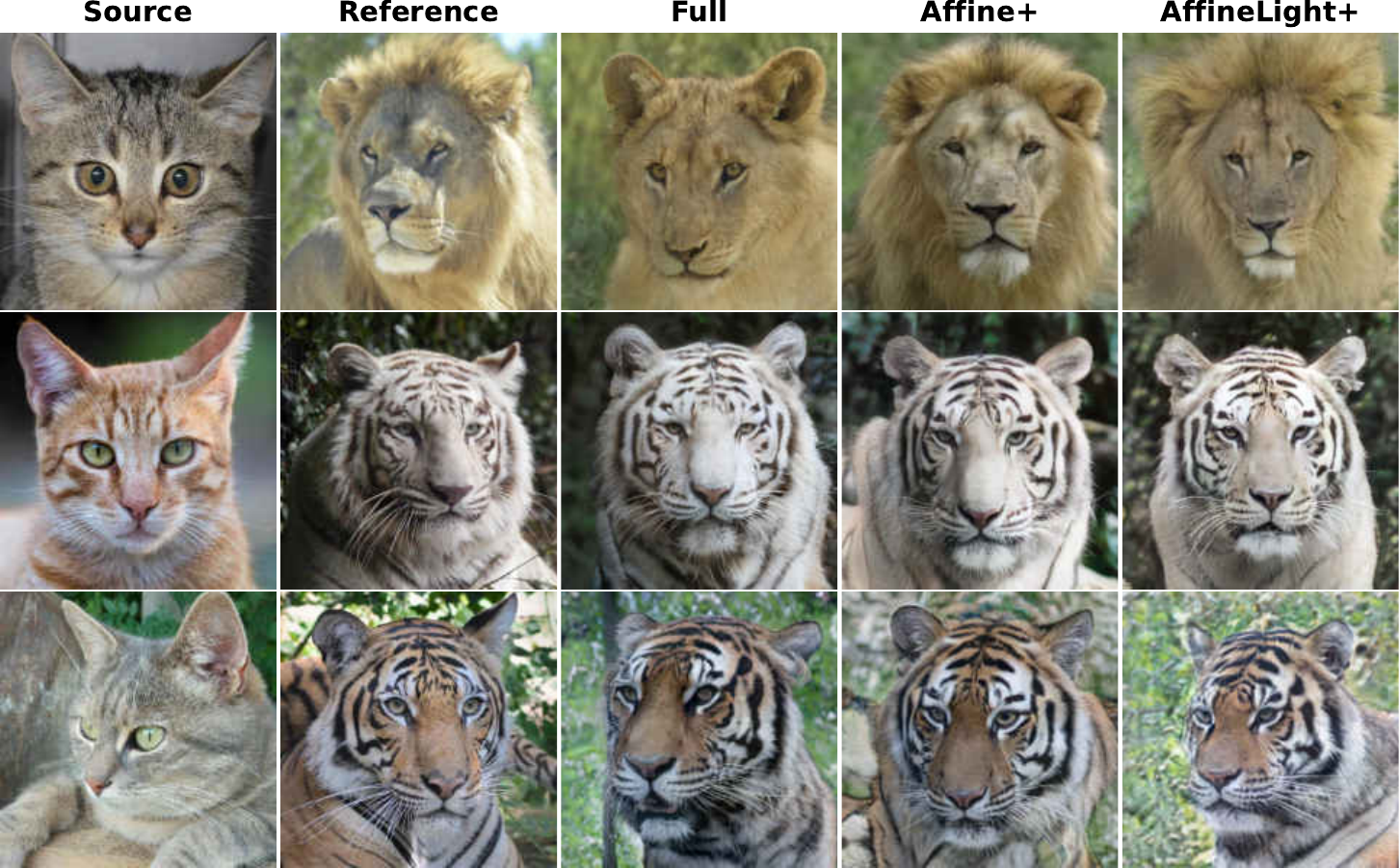} &
        \includegraphics[width=0.30\textwidth]{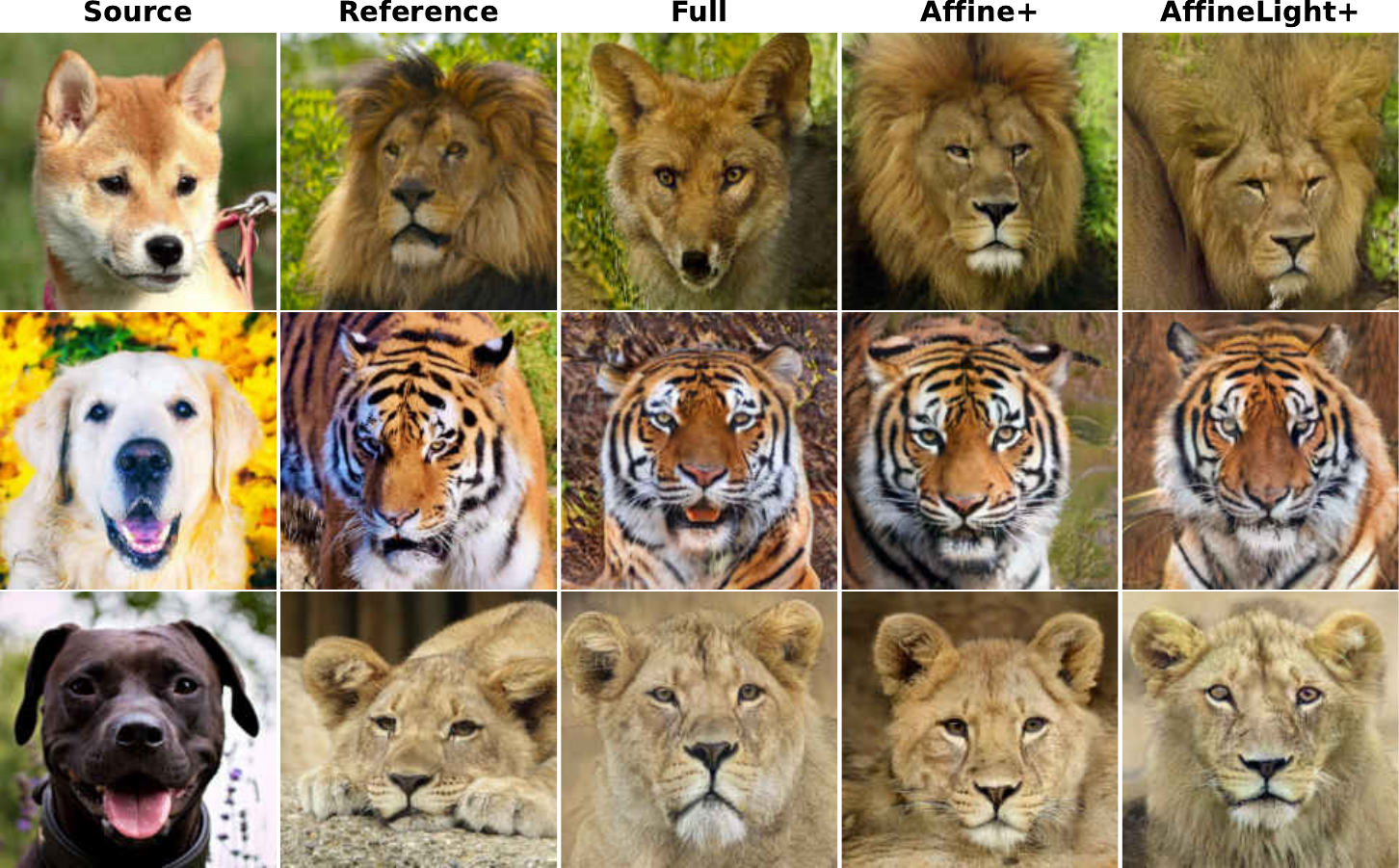} \\
        \includegraphics[width=0.30\textwidth]{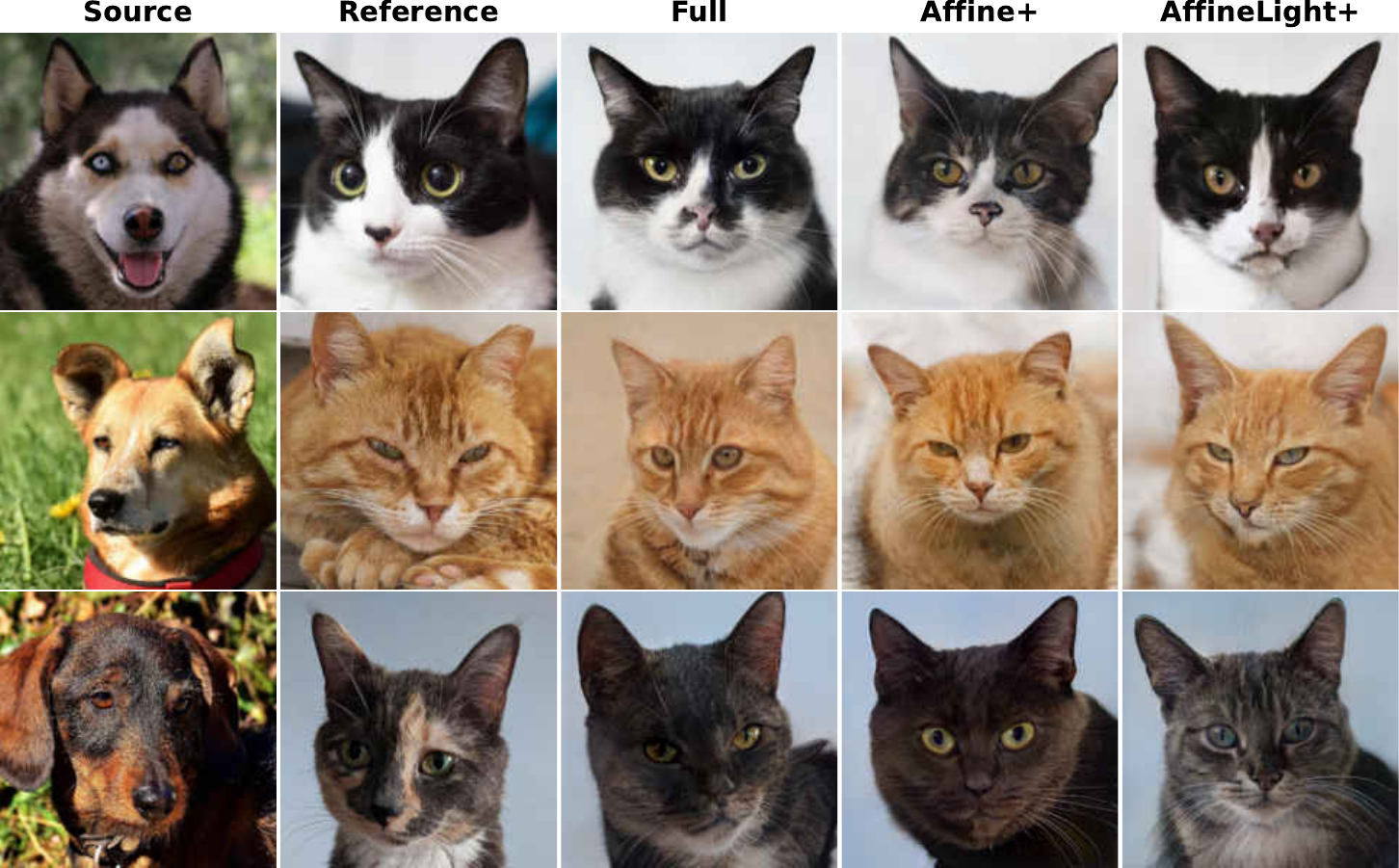} &
        \includegraphics[width=0.30\textwidth]{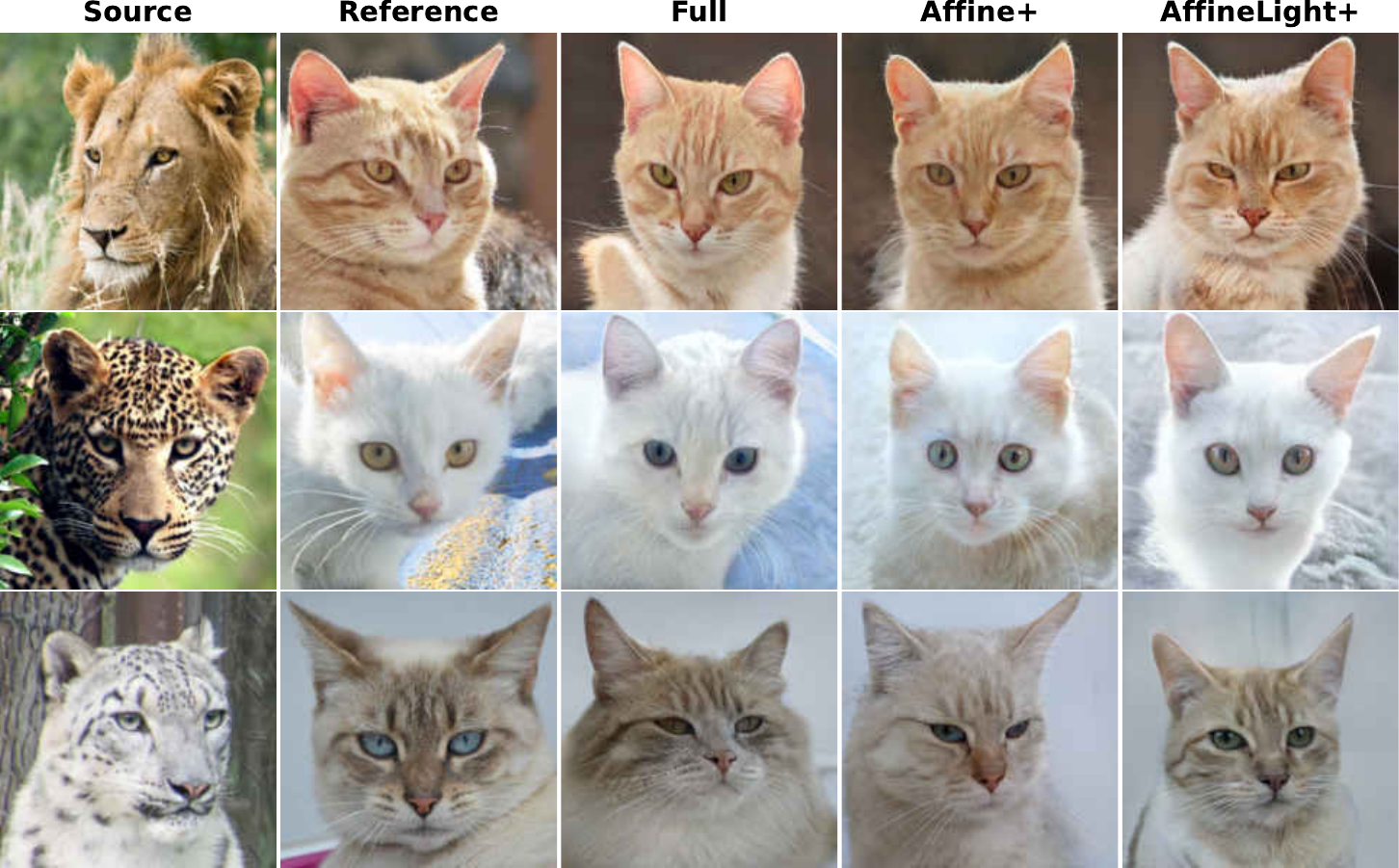} &
        \includegraphics[width=0.30\textwidth]{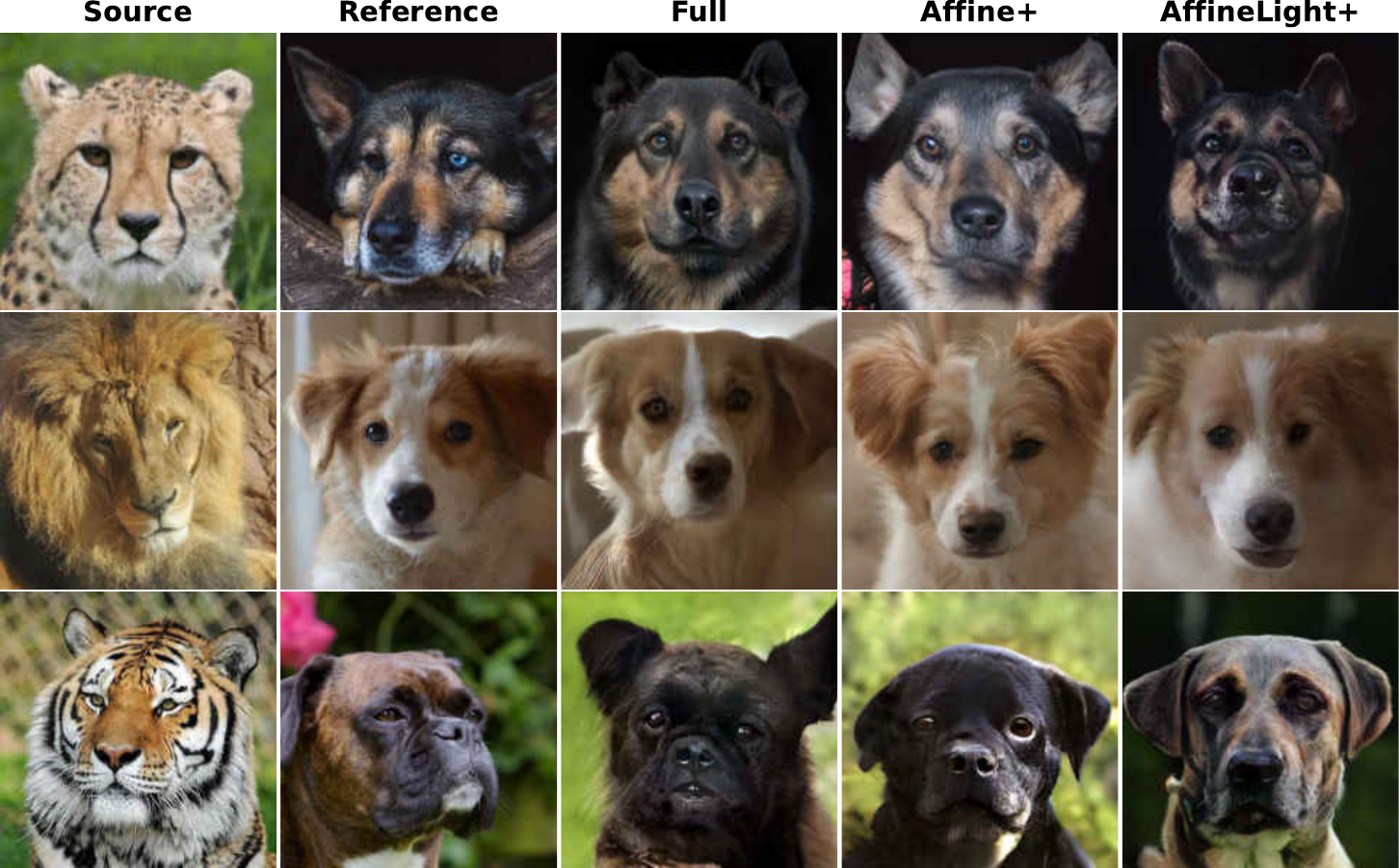} \\
    \end{tabular}
}
\centering
\begin{adjustbox}{width=0.95\textwidth,center}
\centering
% \begin{tabular}{cccccccccccccc}
% \toprule
%  &       & \multicolumn{2}{c}{cat2dog} & \multicolumn{2}{c}{cat2wild} & \multicolumn{2}{c}{dog2cat} & \multicolumn{2}{c}{dog2wild} & \multicolumn{2}{c}{wild2cat} & \multicolumn{2}{c}{wild2dog} \\
%  &   Size &     FID & KID$\times 10^{3}$ &      FID & KID$\times 10^{3}$ &     FID & KID$\times 10^{3}$ &      FID & KID$\times 10^{3}$ &      FID & KID$\times 10^{3}$ &      FID & KID$\times 10^{3}$ \\
% \midrule\midrule
% Full        &  60.6M &  $25.1$ &             $12.5$ &   $9.03$ &             $2.59$ &  $7.64$ &             $2.22$ &   $11.9$ &             $3.60$ &   $7.71$ &             $2.52$ &   $24.0$ &             $11.5$ \\
% Affine+     &  10.2M &  $21.8$ &             $7.33$ &   $5.79$ &             $0.80$ &  $8.44$ &             $2.53$ &   $5.59$ &             $0.78$ &   $8.44$ &             $2.32$ &   $20.9$ &             $7.00$ \\
% StyleSpace+ &   1.1M &  $28.3$ &             $13.6$ &   $7.92$ &             $1.65$ &  $10.6$ &             $3.43$ &   $8.13$ &             $1.71$ &   $10.2$ &             $3.32$ &   $28.6$ &             $14.1$ \\
% \midrule\bottomrule
% \end{tabular}

\begin{tabular}{cccccccccccccc}
\toprule
 &       & \multicolumn{2}{c}{cat2dog} & \multicolumn{2}{c}{cat2wild} & \multicolumn{2}{c}{dog2cat} & \multicolumn{2}{c}{dog2wild} & \multicolumn{2}{c}{wild2cat} & \multicolumn{2}{c}{wild2dog} \\
 &   Size &     FID & KID$\times 10^{3}$ &      FID & KID$\times 10^{3}$ &     FID & KID$\times 10^{3}$ &      FID & KID$\times 10^{3}$ &      FID & KID$\times 10^{3}$ &      FID & KID$\times 10^{3}$ \\
\midrule\midrule
Full        &  60.6M &  $25.1$ &   $12.5$ &          $9.03$ &   $2.59$ &         $7.64$ &   $2.22$ &        $11.9$ &   $3.60$ &       $7.71$ &   $2.52$ &              $24.0$ &  $11.5$ \\
Affine+     &  10.2M &  $21.8$ &   $7.33$ &          $5.79$ &   $0.80$ &         $8.44$ &   $2.53$ &        $5.59$ &   $0.78$ &       $8.44$ &   $2.32$ &              $20.9$ &  $7.00$ \\
AffineLight+ & 1.2M  & $25.1$ & $10.8$   & $6.66$ & $1.39$   & $9.36$ & $3.03$   & $6.67$ & $1.42$   & $9.26$ & $2.77$   & $25.4$ & $11.5$ \\
\midrule\bottomrule
\end{tabular}
\end{adjustbox}
\caption{Comparison of reference-based image translation for different parameterizations. All three variants produce realistic images that combine pose and structure from the source image with texture and color from the reference. A quantitative comparison in the table above indicates parity between different parameterizations.}
\label{fig:app_i2i_ref_based}
\end{figure}

Figure~\ref{fig:app_i2i_unconditional} confirms that AffineLight+ and Affine+ can transfer pose and style from the source image to the image in the target domain. Interestingly, Affine+ seems to preserve style a bit worse than the Full or AffineLight+ parameterizations (most notably for the Cat target domain). For reference-based I2I, Figure~\ref{fig:app_i2i_ref_based} shows that all parameterizations are comparable both quantitatively and qualitatively.

\FloatBarrier

\subsection{Cross-domain image morphing}
\label{app:morphing}
\begin{figure*}[!h]
  \centering
  \includegraphics[width=0.9\textwidth]{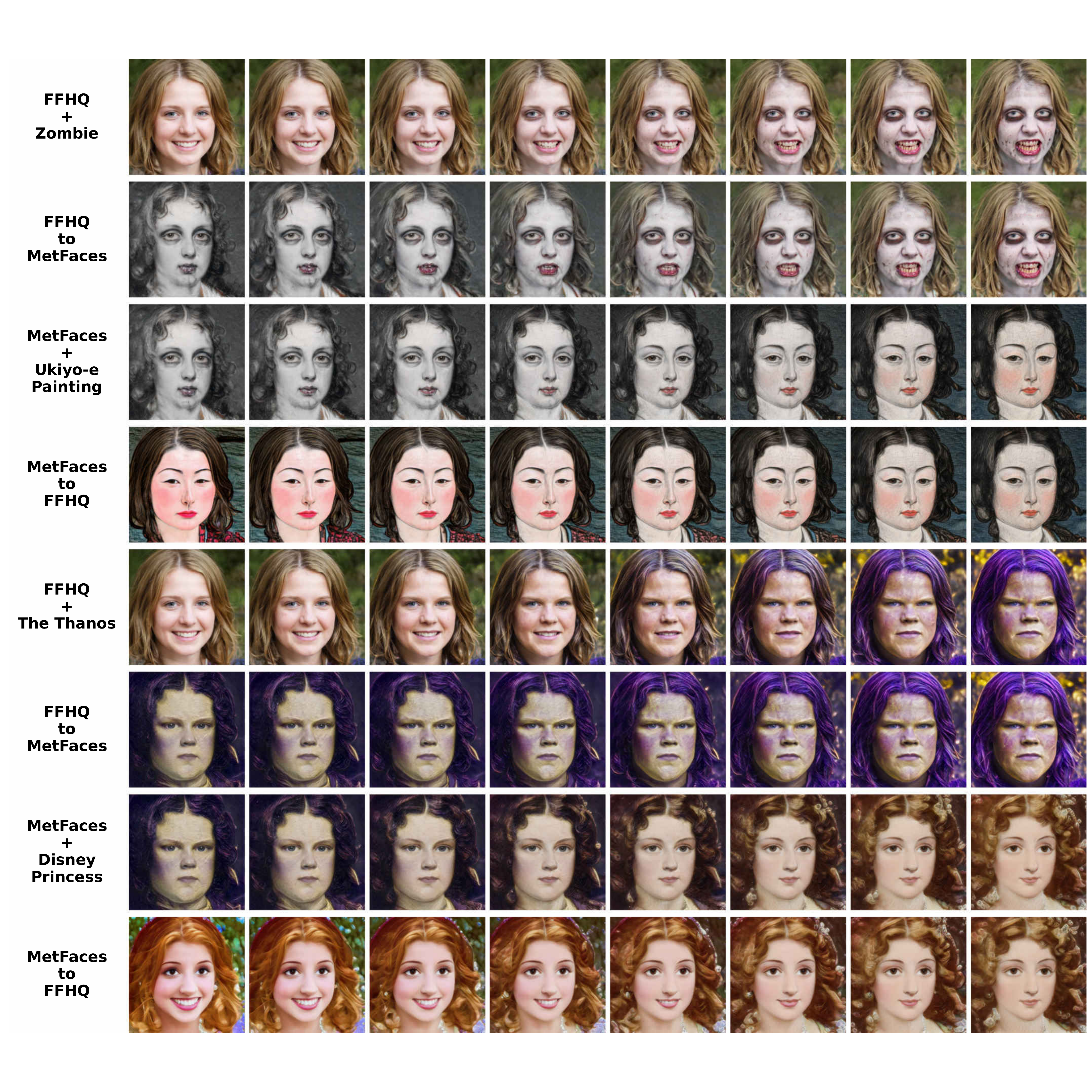}
  \caption{Example of complex cross-domain image morphing. \\
  The first four lines represent "$\textit{FFHQ} \xrightarrow{} \textit{FFHQ}~+~ \textit{Zombie} \xrightarrow{} \textit{MetFaces}~+~ \textit{Zombie} \xrightarrow{} \textit{MetFaces}~+~ \textit{Ukiyo-e Painting} \xrightarrow{} \textit{FFHQ}~+~\textit{Ukiyo-e Painting}$". \\
  The second four lines represent "$\textit{FFHQ} \xrightarrow{} \textit{FFHQ}~+~\textit{The Thanos} \xrightarrow{} \textit{MetFaces}~+~\textit{The Thanos} \xrightarrow{} \textit{MetFaces}~+~\textit{Disney Princess} \xrightarrow{} \textit{FFHQ}~+~\textit{Disney Princess}$".}
  \label{fig:app_morph_0}
  \vspace{-0.5cm}
\end{figure*}

\begin{figure*}[!h]
  \centering
  \includegraphics[width=\textwidth]{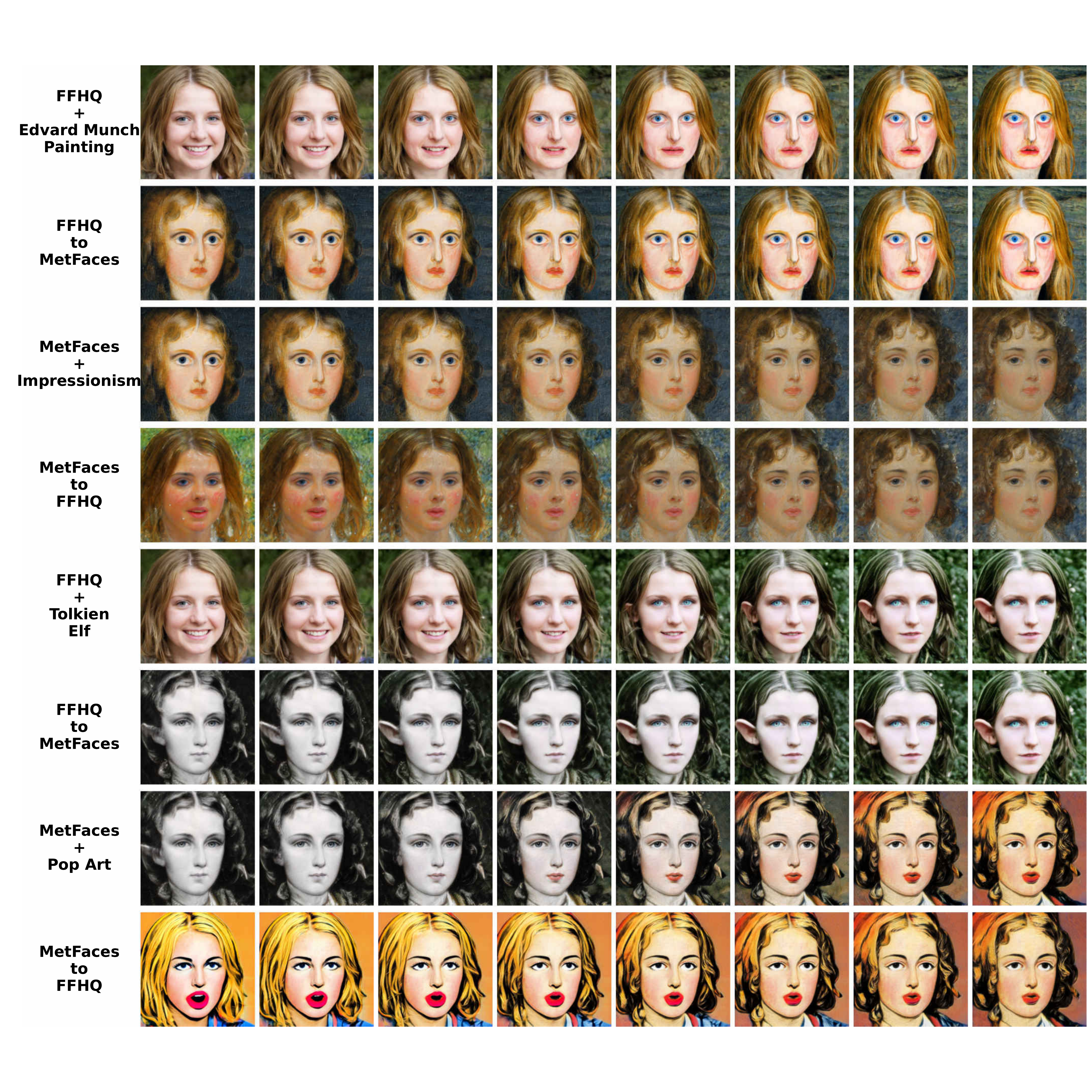}
  \caption{Example of complex cross-domain image morphing. \\
  The first four lines represent "$\textit{FFHQ} \xrightarrow{} \textit{FFHQ}~+~ \textit{Edvard Munch Painting} \xrightarrow{} \textit{MetFaces}~+~ \textit{Edvard Munch Painting} \xrightarrow{} \textit{MetFaces}~+~ \textit{Impressionism} \xrightarrow{} \textit{FFHQ}~+~\textit{Impressionism}$". \\
  The second four lines represent "$\textit{FFHQ} \xrightarrow{} \textit{FFHQ}~+~\textit{Tolkien Elf} \xrightarrow{} \textit{MetFaces}~+~\textit{Tolkien Elf} \xrightarrow{} \textit{MetFaces}~+~\textit{Pop Art} \xrightarrow{} \textit{FFHQ}~+~\textit{Pop Art}$".}
  \label{fig:app_morph_1}
  \vspace{-0.5cm}
\end{figure*}

\begin{figure*}[!h]
  \centering
  \includegraphics[width=\textwidth]{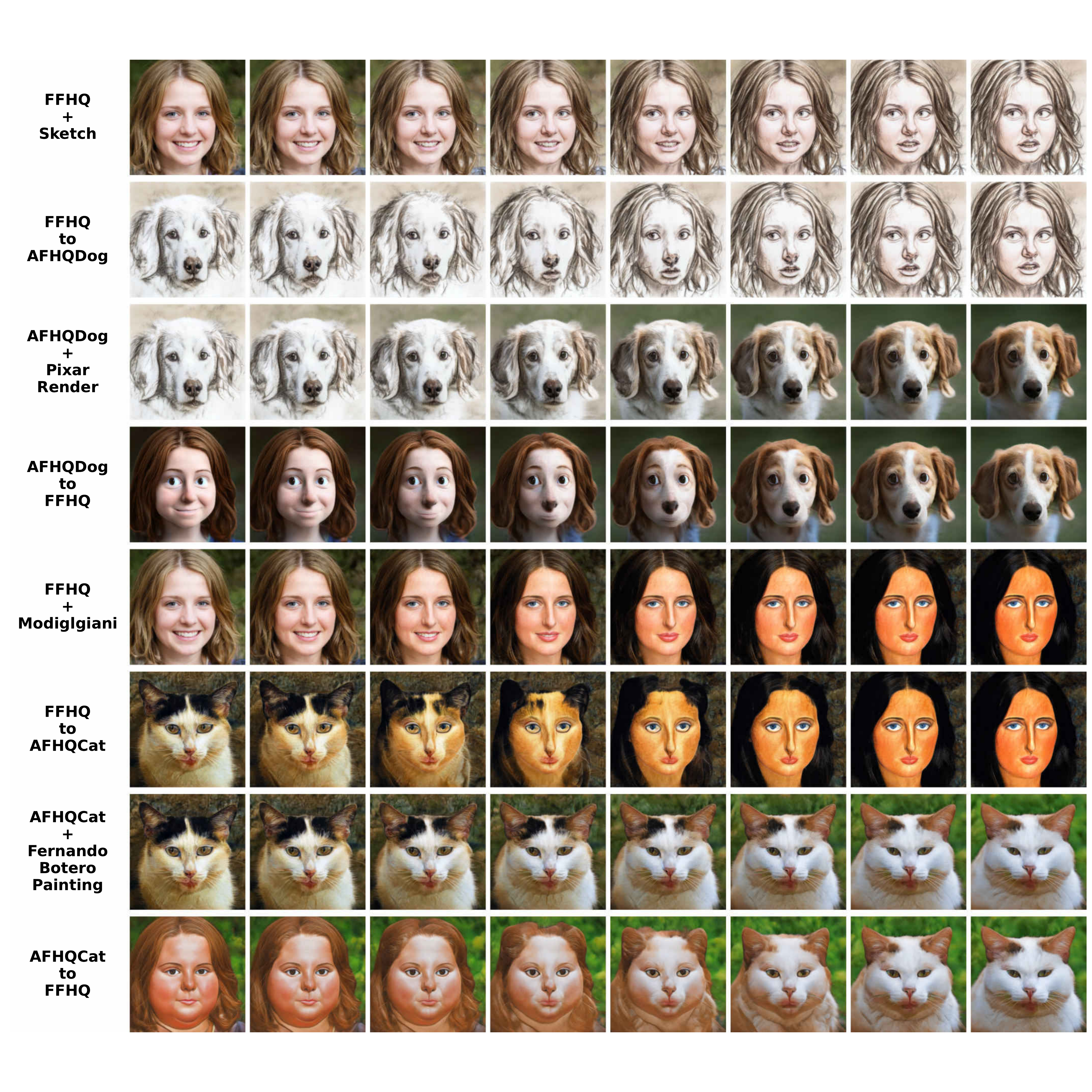}
  \caption{Example of complex cross-domain image morphing. \\
  The first four lines represent "$\textit{FFHQ} \xrightarrow{} \textit{FFHQ}~+~ \textit{Sketch} \xrightarrow{} \textit{AFHQDog}~+~ \textit{Sketch} \xrightarrow{} \textit{AFHQDog}~+~ \textit{Pixar Rendering} \xrightarrow{} \textit{FFHQ}~+~\textit{Pixar Rendering}$". \\
  The second four lines represent "$\textit{FFHQ} \xrightarrow{} \textit{FFHQ}~+~\textit{Modigliani} \xrightarrow{} \textit{AFHQCat}~+~\textit{Modigliani} \xrightarrow{} \textit{AFHQCat}~+~\textit{Fernando Botero Painting} \xrightarrow{} \textit{FFHQ}~+~\textit{Fernando Botero Painting}$".}
  \label{fig:app_morph_2}
  \vspace{-0.5cm}
\end{figure*}

\begin{figure*}[!h]
  \centering
  \includegraphics[width=\textwidth]{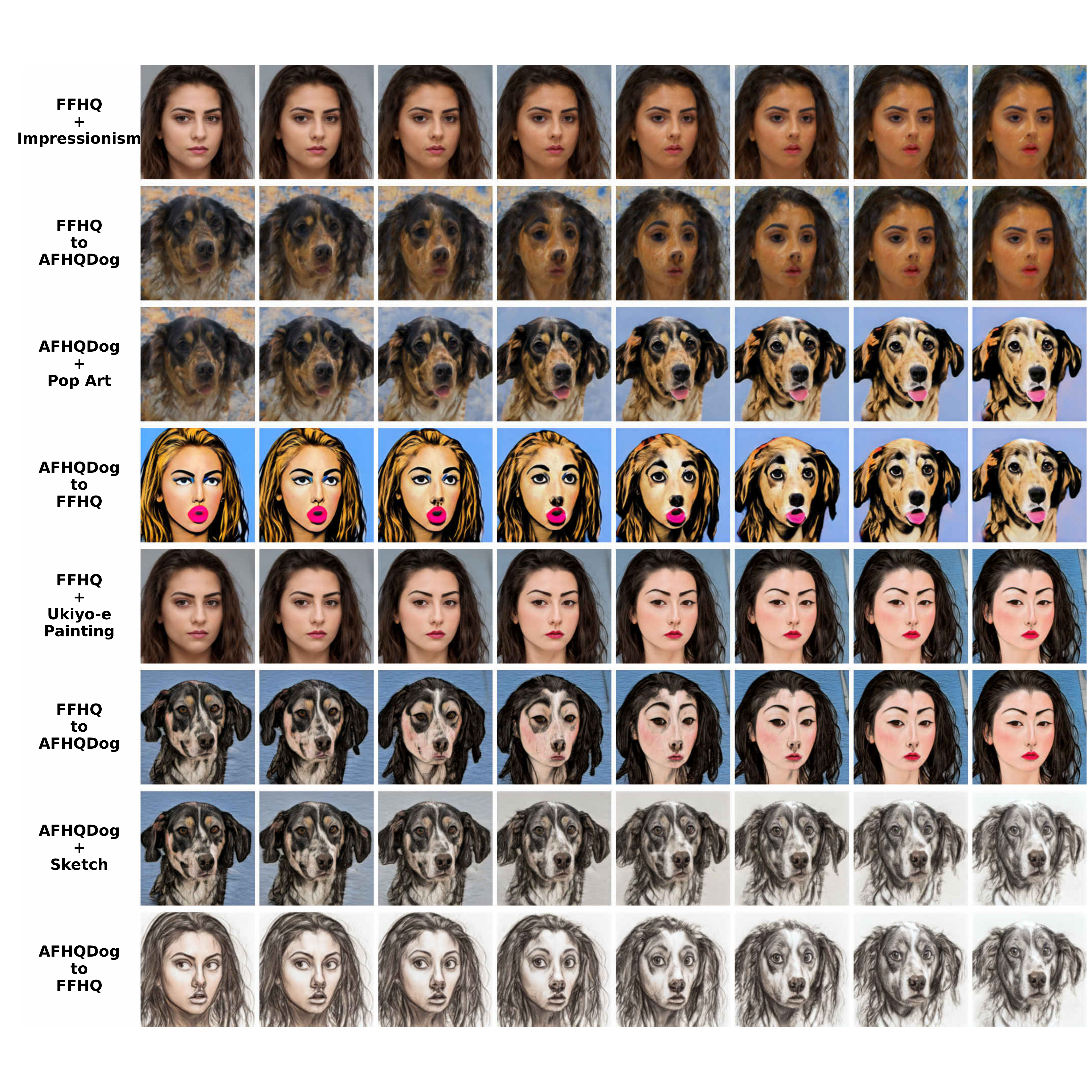}
  \caption{Example of complex cross-domain image morphing. \\
  The first four lines represent "$\textit{FFHQ} \xrightarrow{} \textit{FFHQ}~+~ \textit{Impressionism} \xrightarrow{} \textit{AFHQDog}~+~ \textit{Impressionism} \xrightarrow{} \textit{AFHQDog}~+~ \textit{Pop Art} \xrightarrow{} \textit{FFHQ}~+~\textit{Pop Art}$". \\
  The second four lines represent "$\textit{FFHQ} \xrightarrow{} \textit{FFHQ}~+~\textit{Ukiyo-e Painting} \xrightarrow{} \textit{AFHQDog}~+~\textit{Ukiyo-e Painting} \xrightarrow{} \textit{AFHQDog}~+~\textit{Sketch} \xrightarrow{} \textit{FFHQ}~+~\textit{Sketch}$".}
  \label{fig:app_morph_3}
  \vspace{-0.5cm}
\end{figure*}

We provide more examples of complex cross-domain image morphing in \Cref{fig:app_morph_0,fig:app_morph_1,fig:app_morph_2,fig:app_morph_3}. Each complex image morphing is represented by 4 rows. First row represents cross-domain image morphing via smoothly adding StyleDomain direction in FFHQ domain. Second row shows cross-domain image morphing via weights interpolation between aligned generators with fixed StyleDomain direction.
The third row demonstrates gradual interpolation between two StyleDomain directions that correspond to two different domains. Fourth row represents backward StyleGAN2 weight interpolation to FFHQ domain with fixed StyleDomain direction. There are two examples of cross-domain image morphing in each Figure. 

We observe that transitions are very smooth and by leveraging properties of StyleDomain directions we can obtain complex cross-domain image morphing.

%Examples [\ref{fig:app_morph_0}, \ref{fig:app_morph_1}, \ref{fig:app_morph_2}, \ref{fig:app_morph_3}] of combined morphing are provided in this section. Each combined morphing is represented with 4 rows. First row represents cross-domain image morphing via StyleSpace offsets model in FFHQ domain. Second row represents cross-domain image morphing via weights interpolation with fixed StyleSpace offsets. Third row represents gradually interpolation between two StyleSpace offsets models. Fourth row represents backward StyleGAN weight interpolation to FFHQ domain with fixed StyleSpace offsets. There are two combined cross-domain image morphings are presented for each Figure.

\FloatBarrier

\end{document}